\documentclass[10pt,twoside]{article}
\usepackage{fullpage}

\usepackage[utf8]{inputenc} % allow utf-8 input
\usepackage[T1]{fontenc}    % use 8-bit T1 fonts
\usepackage{hyperref}       % hyperlinks
\usepackage{url}            % simple URL typesetting
\usepackage{booktabs}       % professional-quality tables
\usepackage{amsfonts}       % blackboard math symbols
\usepackage{nicefrac}       % compact symbols for 1/2, etc.
\usepackage{microtype}      % microtypography

\hypersetup{breaklinks=true}
\newenvironment{itemizex}%
{\begin{itemize}[]}%
{\end{itemize}}

\newenvironment{enumeratex}[1][]%
{\begin{enumerate}[#1, wide=\parindent]}%
{\end{enumerate}}

\usepackage{floatrow, makecell, subfig}
\newfloatcommand{capbtabbox}{table}[][\FBwidth]

\usepackage{amssymb, graphicx, amsmath, listings, mathtools, amsthm, tabularx, xspace, dsfont}
\usepackage[shortlabels]{enumitem}

\newtheorem{theorem}{Theorem}[section]
\newtheorem{lemma}[theorem]{Lemma}
\newtheorem{proposition}[theorem]{Proposition}

\newcommand{\figheightintuition}{1.75in}
\newcommand{\figheightexp}{1.8in}
\newcommand{\figheightlegend}{0.5in}

\newcommand{\indicator}{\mathds{1}}
\newcommand{\reals}{\mathbb{R}}
\newcommand{\Prob}{\mathbb{P}}
\newcommand{\Expect}{{\mathbb E}}
\newcommand{\sample}{\sim}

\DeclareMathOperator*{\argmin}{argmin}
\DeclareMathOperator*{\argmax}{argmax}

\newcommand{\del}{\nabla}
\newcommand{\binomial}{\text{Binom}}

\newcommand{\given}{\mid}

\newcommand{\union}{\cup}
\newcommand{\defn}{\vcentcolon=}

\DeclarePairedDelimiter\ceil{\lceil}{\rceil}
\DeclarePairedDelimiter\floor{\lfloor}{\rfloor}
\DeclarePairedDelimiter{\abs}{\lvert}{\rvert}
\DeclarePairedDelimiter{\norm}{\lVert}{\rVert}
\newcommand{\norminf}[1]{\norm{#1}_\infty}

\newcommand{\greaterorder}{\gtrsim}
\newcommand{\lessorder}{\lesssim}

\newcommand{\stepone}{\text{(i)}\xspace}
\newcommand{\steptwo}{\text{(ii)}\xspace}
\newcommand{\stepthree}{\text{(iii)}\xspace}

\newcommand{\mle}{MLE\xspace}
\newcommand{\expandedmle}{stretched-MLE\xspace}
\newcommand{\Expandedmle}{Stretched-MLE\xspace}
\newcommand{\standardmle}{standard MLE\xspace} % when we need to emphasize it's the standard MLE

\newcommand{\stretched}{stretched\xspace}
\newcommand{\stretching}{stretching\xspace}

\newcommand{\app}{appendix\xspace}
\newcommand{\App}{Appendix\xspace}

\def\Erdos{Erd\H{o}s}
\def\Renyi{R\'enyi}

\newcommand{\weightgt}{\weight^*}
\newcommand{\weightestmle}{\widehat{\weight}_\text{\normalfont MLE}}

\newcommand{\param}{\theta}
\newcommand{\paramgt}{\param^*}
\newcommand{\numitems}{d}
\newcommand{\numcomparisons}{k}
\newcommand{\numobservations}{k}

\newcommand{\boundgt}{B}
\newcommand{\sigbound}{\tau}
\newcommand{\term}{R}
\newcommand{\termmax}{\term^+}
\newcommand{\termmin}{\term^-}
\newcommand{\idxitem}{i}
\newcommand{\idxitemalt}{j}
\newcommand{\idxitemthree}{m}
\newcommand{\idxcomparison}{r}
\newcommand{\idxitemthreemax}{{\idxitemthree^+}}
\newcommand{\idxitemthreemin}{{\idxitemthree^-}}
\newcommand{\negloglikelihood}{\mathcal{\ell}}
\newcommand{\negloglikelihoodreparam}{\mathcal{\ell}_\weight}

\newcommand{\paramest}{\widehat{\param}}
\newcommand{\paramestbound}[1]{\widehat{\param}^{(#1)}}
\newcommand{\paramestboundgt}{\paramestbound{\boundgt}}
\newcommand{\paramestboundused}{\paramestbound{\boundused}}
\newcommand{\paramestunconstrain}{\paramestbound{\infty}}

\newcommand{\paramoracle}{\widetilde{\param}}
\newcommand{\paramoracleunconstrain}{\paramoracle^{(\infty)}}
\newcommand{\paramoracleevent}{\paramoracle^\meancondition}

\newcommand{\boundused}{A}
\newcommand{\bias}{\beta}
\newcommand{\accuracy}{\alpha}
\newcommand{\numwins}{W}
\newcommand{\mean}{\mu}
\newcommand{\varmean}{t} % free argument in the definition of functions
\newcommand{\meangt}{\mean^*}
\newcommand{\meanemp}{\mean}
\newcommand{\meanoracle}{\widetilde{\mean}}
\newcommand{\meanoracleevent}{\widetilde{\mean}^\meancondition}
\newcommand{\meancondition}{v}

\newcommand{\bigO}{{\cal O}}
\newcommand{\bigOmega}{\Omega}
\newcommand{\bigOlog}{\widetilde{\bigO}}
\newcommand{\bigTheta}{\Theta}
\newcommand{\smallo}{o}

\newcommand{\func}{f}
\newcommand{\funcalt}{g} % for order notations
\newcommand{\numwinsobs}{w}
\newcommand{\graphundirected}{G'}
\newcommand{\graphcomparison}{G}

\newcommand{\const}{c}

\newcommand{\funclog}{h}
\newcommand{\funclogplus}{{\funclog^+}}
\newcommand{\funclogplusd}{\funclog^+_\numitems}
\newcommand{\funcloglinear}{{\funclog_0}}
\newcommand{\meanmax}{\mean_+}
\newcommand{\meanmin}{\mean_-}
\newcommand{\meanmaxd}{\mean_{\numitems,+}}
\newcommand{\meanmind}{\mean_{\numitems,-}}

\newcommand{\numsamples}{n}
\newcommand{\diforacle}{\delta}
\newcommand{\difcondition}{\widetilde{\delta}}
\newcommand{\difexpect}{\Delta}

\newcommand{\funcsigmoid}{f}
\newcommand{\sigvar}{x} % argument for sigmoid function
\newcommand{\funcsigmoiddiff}{g}
\newcommand{\siganchor}{x}
\newcommand{\sigdif}{t}
\newcommand{\meanvalue}{\lambda}

\newcommand{\meanvaluealt}{\widetilde{\meanvalue}}
\newcommand{\event}{E}
\newcommand{\eventintro}{\mathcal{E}}
\newcommand{\eventcomplement}{\overline{\event}}
\newcommand{\eventlemmaub}{\event_0}
\newcommand{\eventlemmalb}{\event_0}
\newcommand{\eventlemmaubcomp}{\overline{\eventlemmalb}}
\newcommand{\eventlemmalbcomp}{\overline{\eventlemmalb}}
\newcommand{\whp}{w.h.p.\xspace}
\newcommand{\whpone}{w.h.p.($\frac{1}{\numitems\numcomparisons}$)}
\newcommand{\whponecondition}[1]{w.h.p.($\frac{1}{\numitems\numcomparisons}\given #1$)}
\newcommand{\whponeconditionmean}{w.h.p.($\frac{1}{\numitems\numcomparisons}\given \eventmean$)}

\newcommand{\paramoracleclip}{\widetilde{\param}^{(\boundgt)}}

\newcommand{\eventmean}{\event_\meancondition}

\newcommand{\rangeparam}{\Theta}
\newcommand{\rangeboxused}{\rangeparam_\boundused}
\newcommand{\rangeboxgt}{\rangeparam_{\boundgt}}
\newcommand{\rangeboxinf}{\rangeparam_\infty}
\newcommand{\rangeboxoracle}{\rangeparam_\text{\normalfont{oracle}}}

\newcommand{\constitems}{\numitems_0}
\newcommand{\constcomparisons}{\numcomparisons_0}
\newcommand{\probedge}{p}
\newcommand{\probedgecomp}{q}
\newcommand{\perm}{\pi}
\newcommand{\permshift}{s}

\newcommand{\meanempboundary}{\frac{1}{2}}
\newcommand{\weight}{w}

\newcommand\lfrac[2]{\frac{#1\hfill}{#2\hfill}} % left-align for derivatives

\newcommand{\di}{\,\mathrm{d}}
\newcommand{\onevector}{\mathbf{1}}
\newcommand{\resultcomparison}{X}
\newcommand{\Bernoulli}{\text{Bernoulli}}
\newcommand{\paramcomb}{p}
\newcommand{\probcomb}{\mathcal{P}}
\newcommand{\probcomble}{\mathcal{\probcomb_{\text{le}}}}
\newcommand{\probcombeq}{\mathcal{\probcomb_{\text{eq}}}}

\newcommand{\numcomb}{n}

\newcommand{\selectcomb}{s}

\newcommand{\pobs}{p_{\normalfont obs}}

\newcommand{\constubsketchupper}{\const_{U}}
\newcommand{\constubsketchlower}{\const_{L}}

\newcommand{\laplacian}{L}
\newcommand{\laplacianinv}{\laplacian^\dagger}
\DeclareMathOperator*{\trace}{tr}
\newcommand{\deltwo}{\del^2}

\newcommand{\eigenvalue}{\lambda}

\title{\bf \Large Stretching the Effectiveness of MLE\\ from Accuracy to Bias for Pairwise Comparisons}

\author{\\%
    Jingyan Wang$^\dagger$, Nihar B. Shah$^\dagger$ and R. Ravi$^\ast$\\~\\
    School of Computer Science$^\dagger$ and Tepper School of Business$^\ast$\\
    Carnegie Mellon University\\
    \texttt{\{jingyanw, nihars\}@cs.cmu.edu, ravi@cmu.edu}
}

\date{} 

\begin{document}

\maketitle

\begin{abstract}
A number of applications (e.g., AI bot tournaments, sports, peer grading, crowdsourcing) use pairwise comparison data and the Bradley-Terry-Luce (BTL) model to evaluate a given collection of items (e.g., bots, teams, students, search results). Past work has shown that under the BTL model, the widely-used maximum-likelihood estimator (MLE) is minimax-optimal in estimating the item parameters, in terms of the mean squared error. However, another important desideratum for designing estimators is fairness. In this work, we consider fairness modeled by the notion of bias in statistics. We show that the MLE incurs a suboptimal rate in terms of bias. We then propose a simple modification to the MLE, which ``stretches'' the bounding box of the maximum-likelihood optimizer by a small constant factor from the underlying ground truth domain. We show that this simple modification leads to an improved rate in bias, while maintaining minimax-optimality in the mean squared error. In this manner, our proposed class of estimators provably improves fairness represented by bias without loss in accuracy.
\end{abstract}

%%%%%%%%%%%%
\section{Introduction}

A number of applications involve data in the form of pairwise comparisons among a collection of items, and entail an evaluation of the individual items from this data. An application gaining increasing popularity is competition between pairs of AI bots (e.g.,~\cite{ontanon2013starcraft}). Here a number of AI bots compete with each other in pairwise matchups for a certain task, where each bot plays every other bot a certain number of times in a \emph{round robin} fashion, with the goal of evaluating the quality of each bot. A second example is the evaluation of self-play of AI algorithms in their training phase~\cite{silver2017go}, where again, different copies of an AI bot play against each other a number of times. Applications involving humans include sports and online games such as the English Premier League of football~\cite{kiraly2017sports,elo2019epl} (unofficial ratings) and official world rankings for chess (e.g., FIDE~\cite{fide2017chess} and USCF~\cite{glickman2017chess} ratings). The influence of scientific journals has also been analyzed in this manner, where citations from one journal to another are modeled by pairwise comparisons~\cite{stigler1994citation}.

A common method of evaluating the items based on pairwise comparisons is to assume that the probability of an item beating another equals the logistic function of the difference in the true quality of the two items, and then infer the true quality from the observed outcomes of the comparisons (e.g., the Elo rating system). Various applications employ such an approach to rating from pairwise comparisons, with some modifications tailored to that specific application. Our goal is not to study the application-specific versions, but the foundational underpinnings of such rating systems.

In this paper, we study the pairwise-comparison model that underlies~\cite{glickman1999rating,aldous2017elo} these rating systems, namely the Bradley-Terry-Luce (BTL) model~\cite{bradley1952btl, luce1959btl}. The BTL model assumes that each item is associated to an unknown real-valued parameter representing the quality of that item, and assumes that the probability of an item beating another is the logistic function applied to the difference of the parameters of these two items. The BTL model is also   employed in the applications of peer grading~\cite{shah2013case,lamon2016whooppee} (where the grades of the students are set as the BTL parameters to be estimated), crowdsourcing~\cite{chen2016overcoming,ponce2016chalearn},  and understanding consumer choice in marketing~\cite{green1981consumer}. 

%%%%%%%%%
\subsection{BTL model and maximum likelihood estimation}
Now we present a formal definition of the BTL model. Let $\numitems \ge 2$ denote the number of items. The $\numitems$ items are associated to an unknown parameter vector $\paramgt\in \reals^\numitems$ whose  $\idxitem^\text{th}$ entry represents the underlying quality of item $\idxitem \in [\numitems]$. When any item $i \in [\numitems]$ is compared with any item $\idxitemalt \in [\numitems]$ in the BTL model, the item $\idxitem$ beats item $\idxitemalt$ with probability
\begin{align}
    \label{eq:BTLprob}
    \frac{1}{1 + e^{-(\paramgt_\idxitem - \paramgt_\idxitemalt)}},
\end{align}
independent of all other comparisons. The probability of item $j$ beating $i$ is one minus the expression~\eqref{eq:BTLprob} above. We consider the ``league format''~\cite{aldous2017elo} of comparisons where every pair of items is compared $\numcomparisons$ times.

We follow the usual assumption~\cite{hajek2014likelihood,shah2016topology} under the BTL model that the true parameter vector $\paramgt$ lies in the set $\rangeboxgt$ parameterized by a \emph{constant} $\boundgt > 0$ and satisfy:
\begin{align}\label{eq:domain_true}
    \rangeboxgt = \{\param \in \reals^\numitems \given  \norminf{\param}\le \boundgt\, \text{ and }\sum_{\idxitem=1}^\numitems\param_\idxitem=0\}.
\end{align}
The first constraint requires that the magnitude of the parameters is bounded by some constant $\boundgt$. We call this constraint the ``box constraint''. A box constraint is necessary, because otherwise the estimation error can diverge to infinity~\cite[Appendix G]{shah2016topology}. The second constraint requires the parameters to sum to $0$. This is without loss of generality due to the shift-invariance property of the BTL model.

A large amount of both theoretical~\cite{hunter2004mm,hajek2014likelihood,szorenyi2015online,negahban2016centrality,shah2016topology} and applied~\cite{stigler1994citation,sham1995genetics,chen2016overcoming,ponce2016chalearn} literature focuses on the goal of estimating the parameter vector $\paramgt$ of the BTL model. A standard and widely-studied estimator is the maximum-likelihood estimator (\mle):
\begin{align}\label{eq:intro_mle}
    \paramestboundgt = \argmin_{\param\in \rangeboxgt} \negloglikelihood(\param),
\end{align}
where $\negloglikelihood$ is the negative log-likelihood function. Letting $\numwins_{\idxitem\idxitemalt}$ denote a random variable representing the number of times that item $\idxitem \in [\numitems]$ beats item $\idxitemalt\in [\numitems]$, the log-likelihood function $\negloglikelihood$ is given by:
     \begin{align*}
        \negloglikelihood(\param) \defn \negloglikelihood(\{\numwins_{\idxitem\idxitemalt}\}; \param) & = - \sum_{1 \leq \idxitem < \idxitemalt \leq \numitems} \left[\numwins_{\idxitem\idxitemalt} \log\left(\frac{1}{1 + e^{-(\param_\idxitem-\param_\idxitemalt)}}\right) +  \numwins_{\idxitemalt\idxitem}\log\left(\frac{1}{1 + e^{-(\param_\idxitemalt-\param_\idxitem)}}\right)\right].
    \end{align*}

%%%%%%%%%%%%
\subsection{Metrics}
\paragraph{Accuracy.}
A common metric used in the literature on estimating the BTL model is the \emph{accuracy} of the estimate, measured in terms of the mean squared error.
Formally, the accuracy of any estimator $\paramest$ is defined as:
\begin{align*} 
\accuracy(\paramest) \defn \sup_{\paramgt\in \rangeboxgt}\Expect[ \norm{ \paramest - \paramgt }_2^2 ] .
\end{align*}
Importantly, past work~\cite{hajek2014likelihood, shah2016topology} has shown that the \mle~\eqref{eq:intro_mle} has the appealing property of being minimax-optimal in terms of the accuracy.

\paragraph{Bias. }  Another important desideratum for designing and evaluating estimators is fairness. For example, in sports or online games, we do not want to assign scores in such a way that it systematically gives certain players higher scores than their true quality, but at the same time gives certain other players lower scores than their true quality. In this paper, we use the standard definition of \emph{bias} in statistics as the notion of fairness. For any estimator, the bias incurred by this estimator on a parameter is defined as the difference between the expected value of the estimator and the true value of the parameter. Since our parameters are a vector, we consider the worst-case bias, that is, the maximum magnitude of the bias across all items. Formally, the bias of any estimator $\paramest$ is defined as:
\begin{align*}
    \bias(\paramest) \defn \sup_{\paramgt\in \rangeboxgt} \norm{\Expect[\paramest] - \paramgt}_\infty.
\end{align*}

With this background, we now provide an overview of the contributions of this paper.

\begin{figure}
\begin{floatrow}
\ffigbox{%
  \includegraphics[width=\linewidth]{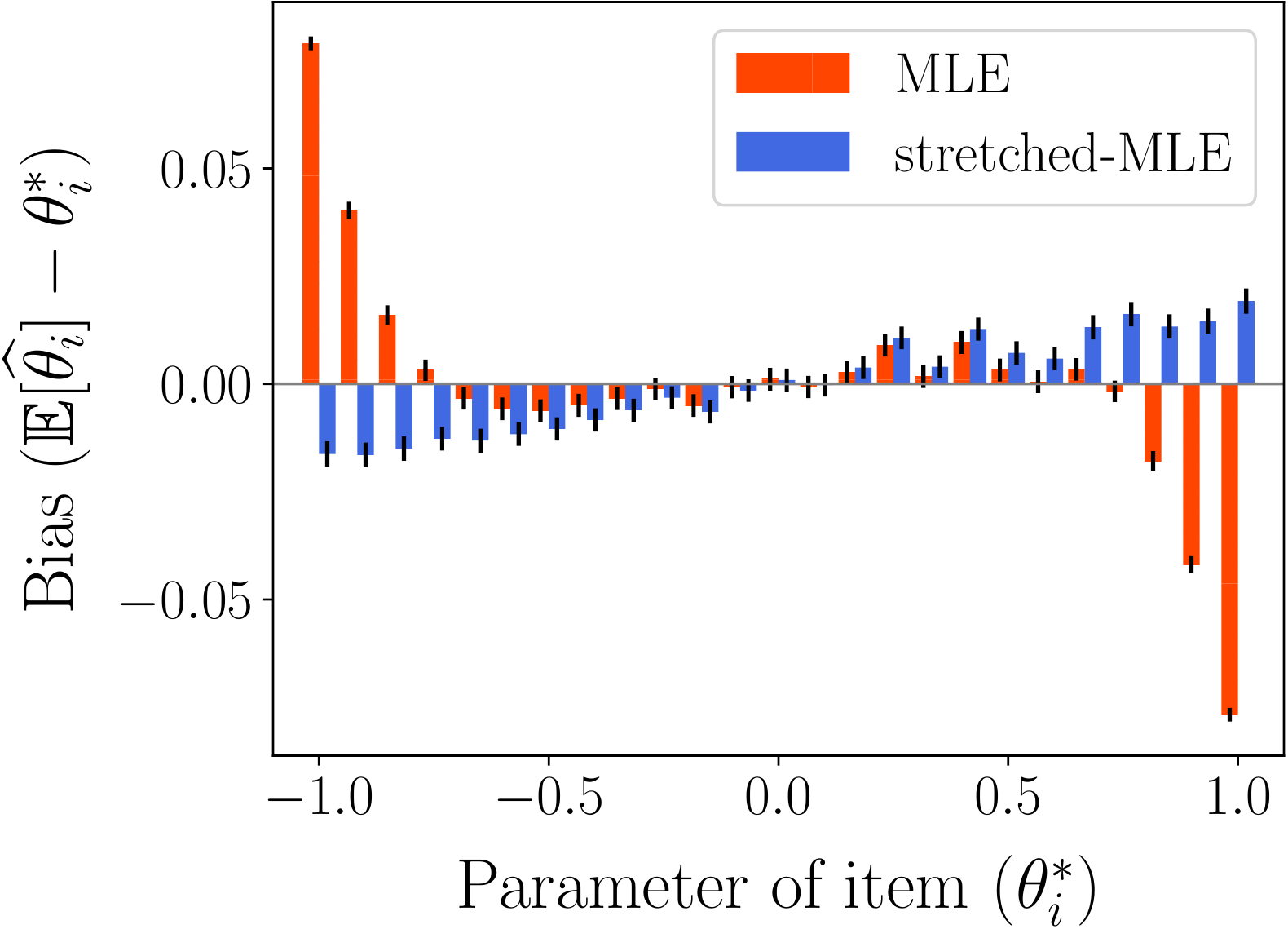}%
}{%
  \caption{Biases on items of different parameters, induced by the \mle and our \expandedmle (with $\boundused = 2$). Our estimator significantly reduces the maximum magnitude of the bias across the items. Note that this figure plots the bias including its sign: A positive bias means over-estimation of the parameter, and a negative bias means under-estimation of the parameter. Each bar is a mean over $5000$ iterations. }\label{fig:example_bias}%
}
\capbtabbox{%
  \begin{tabular}{c|c|c} \hline
  Estimator & Bias & \thead{Mean squared\\ error} \\ \hline
  \thead{Standard\\\mle~~~$\paramestboundgt$} &  \thead{$\bigOmega(\frac{1}{\sqrt{\numitems\numcomparisons}})$\\ (Thm. \ref{thm:mle_ub_lb}\ref{part:lb})} & \thead{$\bigO(\frac{1}{\numcomparisons}) $\\minimax-optimal\\\cite{hajek2014likelihood,shah2016topology}} \\\hline
  \thead{Unconstrained\\ \mle~~~$\paramestunconstrain$} &  \thead{Undefined} & \thead{$\infty$} \\\hline
  \thead{Stretched\\ \mle~~~$\paramestbound{\boundused}$} & \thead{$\bigOlog(\frac{1}{\numitems\numcomparisons})$ \\ (Thm. \ref{thm:mle_ub_lb}\ref{part:ub})} & \thead{$\bigO(\frac{1}{\numcomparisons})$\\ minimax-optimal \\ (Thm. \ref{thm:l2_ub_lb}\ref{part:l2_ub})}\\ \hline
  \end{tabular}
  \vspace{35pt}
}{%
  \caption{Theoretical guarantees for the MLE $\paramestboundgt$, unconstrained MLE $\paramestunconstrain$ and the proposed \expandedmle $\paramestbound{\boundused}$ (with a constant $\boundused$ such that $ \boundused > \boundgt$). The proposed \expandedmle achieves a better rate on bias, while retaining minimax optimality in terms of accuracy. Recall that $\numitems$ denotes the  number of items and $\numcomparisons$ denotes the number of comparisons per pair.
  }\label{tab:summary}%
}
\end{floatrow}
\end{figure}

%%%%%%%%%%%%
\subsection{Contribution I: Performance of \mle}

Our first contribution is to analyze the widely-used \mle~\eqref{eq:intro_mle} in terms of its bias. Let us begin with a visual illustration through simulation. Consider $\numitems=25$ items with parameter values equally spaced in the interval $[-1,1]$, where $\numcomparisons=5$ pairwise comparisons are observed between each pair of items under the BTL model. We estimate the parameters using the \mle, and plot the bias on each item across $5000$ iterations of the simulation in Figure~\ref{fig:example_bias} (striped red). 
The \mle shows a systematic bias: it induces a negative bias (under-estimation) on the large positive parameters, and a positive bias (over-estimation) on the large negative parameters. In the applications of interest, the \mle thus systematically underestimates the abilities of the top players/students/items and overestimates the abilities of those at the bottom.

In this paper, we theoretically quantify the bias incurred by the \mle.

\begin{theorem}[MLE bias lower bound; Informal]
The \mle~\eqref{eq:intro_mle} incurs a bias $\beta(\paramestboundgt)$ lower bounded as $\Omega(\frac{1}{\sqrt{\numitems\numcomparisons}})$. 
\end{theorem}

As shown by our results to follow, this bias is suboptimal. Our proof for this result indicates that the bias is incurred because the \mle operates under the accurately specified model with the box constraint at $\boundgt$. That is, the \mle ``clips'' the estimate to lie within the set $\rangeboxgt$. This issue is visible in the simulation of Figure~\ref{fig:example_bias} where the bias is the largest when the true values of the parameters are near the boundaries $\pm \boundgt$. For example, consider a true parameter whose value equals $\boundgt$. The estimate of this parameter sometimes equals the largest allowed value $\boundgt$ (due to the box constraint), and sometimes is smaller than $\boundgt$ (due to the randomness of the data). Therefore, in expectation, the estimate of this parameter incurs a negative bias. An analogous argument explains the positive bias when the true parameter equals or is close to $-\boundgt$.

\subsection{Contribution II: Proposed stretched estimator and its theoretical guarantees}

Our goal is to design an estimator with a lower bias while maintaining  high accuracy. Since the \mle~\eqref{eq:intro_mle} is already widely studied and used, it is also desirable from a practical and computational standpoint that the new estimator is a simple modification of the \mle~\eqref{eq:intro_mle}. With this motivation in mind, an intuitive approach is to consider the \mle but without the box constraint ``$\norminf{\param}\le \boundgt$''. We call the estimator without the box constraint as the ``unconstrained \mle'', and denote it by $\paramestunconstrain$, because removing the box constraint is equivalent to setting the box constraint to $\infty$:
\begin{align}
    \paramestunconstrain = \argmin_{\param\in \rangeboxinf} \negloglikelihood(\param),\label{eq:intro_mle_unconstrain}
\end{align}
where $\rangeboxinf\defn \{\param\in \reals^\numitems \given \sum_{\idxitem=1}^\numitems \param_\idxitem = 0\}$. The unconstrained \mle $\paramestunconstrain$ incurs an unbounded error in terms of accuracy. This is because with non-zero probability an item beats all others, in which case the unconstrained \mle estimates the parameter of this item as $\infty$, thereby inducing an unbounded mean squared error.

Consequently, in this work, we propose the following simple modification to the \mle which is a middle ground between the standard \mle~\eqref{eq:intro_mle} and the unconstrained \mle. Specifically, we consider a ``\expandedmle'', which is associated to a parameter $\boundused$ such that $\boundused > \boundgt$. Given the parameter $\boundused$, the \expandedmle is identical to~\eqref{eq:intro_mle} but ``stretches'' the box constraint to $\boundused$:
\begin{align}
    \paramestbound{\boundused} = \argmin_{\param\in \rangeboxused} \negloglikelihood(\param),
    \label{eq:intro_expmle}
\end{align}
where $\rangeboxused \defn \{\param \in \reals^\numitems \given  \norminf{\param}\le \boundused \, \text{ and }\sum_{\idxitem=1}^\numitems\param_\idxitem=0\}$. That is, $\rangeboxused$ simply replaces the box constraint $\norminf{\param} \le \boundgt$ in~\eqref{eq:domain_true} by the ``\stretched'' box constraint $\norminf{\param} \le \boundused$. 

The bias induced by the \expandedmle (with $\boundused = 2$) in the previous experiment is also shown in Figure~\ref{fig:example_bias} (solid blue). Observe that the maximum bias (incurred at the leftmost item with the largest negative parameter, or the rightmost item with the largest positive parameter) is significantly reduced compared to the \mle. Moreover, the bias induced by the \expandedmle looks qualitatively more evened out across the items. 

Our second main theoretical result proves that the \expandedmle indeed incurs a significantly lower bias.
\begin{theorem}[\Expandedmle bias upper bound; Informal]
The \expandedmle~\eqref{eq:intro_expmle} with $\boundused = 2$ incurs a bias $\bias(\paramestboundused)$ upper bounded as $\bigOlog(\frac{1}{\numitems \numcomparisons})$. 
\end{theorem}

Given the significant bias reduction by our estimator, a natural question is about the accuracy of the \expandedmle, particularly given the unbounded error incurred by the unconstrained \mle. We prove that our \expandedmle is able to maintain the same minimax-optimal rate on the mean squared error as the \standardmle. 
\begin{theorem}[\Expandedmle accuracy upper bound; Informal]
The \expandedmle~\eqref{eq:intro_expmle} with $\boundused = 2$ incurs a mean squared error $\accuracy(\paramestboundused)$ upper bounded as $\bigO(\frac{1}{\numcomparisons})$, which is minimax-optimal. 
\end{theorem}
This result shows {\bf a win-win} by our \expandedmle: \emph{reducing the bias while retaining the accuracy guarantee}. The comparison of the \mle and the \expandedmle in terms of accuracy and bias is summarized in Table~\ref{tab:summary}. Another attractive feature of our result is that the proposed \expandedmle is a simple modification of the standard \mle, which can easily be incorporated in any existing implementation. It is important to note that while our modification to the estimator is simple to implement, our theoretical analyses and the proofs are non-trivial.

%%%%%%
\subsection{Related work}
The logistic nature~\eqref{eq:BTLprob} of the BTL model relates our work to studies of logistic regression (e.g.,~\cite{portnoy1988exponential,he2000increasing,sur2018logistic, fan2019nonuniformity}), among which the paper~\cite{sur2018logistic} is the most closely related to ours. The paper~\cite{sur2018logistic} considers an unconstrained \mle in logistic regression, and shows its bias in the opposite direction as compared to our results on the standard \mle (constrained) in the BTL model. Specifically, the paper~\cite{sur2018logistic} shows that the large positive coefficients are overestimated, and the large negative coefficients are underestimated. There are several additional key differences between the results in~\cite{sur2018logistic} as compared to the present paper. The paper~\cite{sur2018logistic} studies the asymptotic bias of the unconstrained \mle, showing that the unconstrained \mle is not consistent. On the other hand, we operate in a regime where the MLE is still consistent, and study finite-sample bounds. Moreover, the paper~\cite{sur2018logistic} assumes that the predictor variables are i.i.d. Gaussian. On the other hand, in the BTL model the probability that item $\idxitem$ beats item $\idxitemalt$ can be written as $\frac{1}{1 + e^{-x_{\idxitem\idxitemalt}^T\paramgt}}$, where each predictor variable $x_{\idxitem\idxitemalt}\in \reals^\numitems$ has entry $\idxitem$ equal to $1$, entry $\idxitemalt$ equal to $-1$, and the remaining entries equal to $0$.

A common way to achieve bias reduction is to employ finite-sample correction, such as Jackknife~\cite{quenouille1949jackknife} and other methods~\cite{cox1968residual,anderson1979correction,firth1993bias} to the MLE (or other estimators). These methods operate in a low-dimensional regime (small $\numitems$) where the MLE is asymptotically unbiased. Informally, these methods use a Taylor expansion and write the expression for the bias as an infinite sum $\frac{f_1(\paramgt)}{n} + \frac{f_2(\paramgt)}{n^2} + \ldots$, where $n$ is the number samples, for some functions $f_1, f_2, \ldots$. These works then modify the estimator in a variety of ways to eliminate the lower-order terms in this bias expression. However, since the expression is an  infinite sum, eliminating the first term does not guarantee a low rate of the bias. Moreover, since the functions $f_i$ are implicit functions of $\paramgt$, eliminating lower-order terms does not directly translate to explicit worst-case guarantees.

Returning to the pairwise-comparison setting, in addition to the mean squared error, some past work has also considered accuracy in terms of the $\ell_1$ norm error~\cite{agarwal2018totalvariation} and the $\ell_\infty$ norm error~\cite{suh2015spectral,chen2019topk,jang2017topk}. The $\ell_\infty$ bound for a regularized MLE is analyzed in~\cite{chen2019topk}. Our proof for bounding the bias of the standard \mle (unregularized) relies on a high-probability $\ell_\infty$ bound for the unconstrained \mle (unregularized). It is important to note that the bound for regularized \mle from~\cite{chen2019topk} does not carry to unregularized MLE, because the proof from~\cite{chen2019topk} relies on the strong convexity of the regularizer. On the other hand, \emph{our intermediate result provides a partial answer to the open question in~\cite{chen2019topk}} about the $\ell_\infty$ norm for the unregularized MLE (Lemma~\ref{lem:bound_diff_each} in Appendix~\ref{sec:proof_bias_ub_lb}): We establish an $\ell_\infty$ bound for unregularized MLE when $\pobs= 1$, which has the same rate as that of the regularized MLE in~\cite{chen2019topk}.

Another common occurrence of bias is the phenomenon of regression towards the mean~\cite{stigler1997regression}. Regression towards the mean refers to the phenomenon that random variables taking large (or small) values in one measurement are likely to take more moderate (closer to average) values in subsequent measurements. On the contrary, we consider items whose indices are fixed (and are not order statistics). For fixed indices, our results suggest that under the BTL model, the bias (under-estimation of large true values) is in the opposite direction as that in regression towards the mean (over-estimation of large observed values). 

Finally, the paper~\cite{kiraly2017sports} models the notion of fairness in Elo ratings in terms of the ``variance'', where an estimator is considered fair if the estimator is not much affected by the underlying randomness of the pairwise-comparison outcomes. The paper~\cite{kiraly2017sports} empirically evaluates this notion of fairness on the English Premier League data, but presents no theoretical results.

%%%%%%%%%%%%
\section{Main results}\label{sec:main_results}
In this section, we formally provide our main theoretical results on bias and on the mean squared error.

%%%%%%%%%%%%
\subsection{Bias}\label{sec:result_bias}
Recall that $\numitems$ denotes the number of items and $\numcomparisons$ denotes the number of comparisons per pair of items. The true parameter vector is $\paramgt\in \rangeboxgt$ for some pre-specified constant $\boundgt>0$. The following theorem provides bounds on  the bias of the standard \mle $\paramestboundgt$ and that of our \expandedmle $\paramestbound{\boundused}$ with parameter $\boundused$.  In particular, it shows that if $\boundused$ is a finite constant strictly greater than $\boundgt$, then our \expandedmle has a much smaller bias than the \mle when $\numitems$ and $\numcomparisons$ are sufficiently large.

\begin{theorem}\label{thm:mle_ub_lb}
\begin{subequations}
\begin{enumerate}[(a)]
    \item \label{part:lb}
    There exists a constant $\const>0$ that depends only on the constant $\boundgt$, such that
    \begin{align}\label{eq:theorem_lb}
        \bias(\paramestboundgt)  
         \ge \frac{\const}{\sqrt{\numitems\numcomparisons}},
    \end{align}
    for all $\numitems \ge \constitems$ and all $\numcomparisons \ge \constcomparisons$, where $\constitems$ and $\constcomparisons$ are constants that depend only on the constant $\boundgt$.

    %%%
    \item \label{part:ub}
    Let $\boundused$ be any finite constant such that $\boundused > \boundgt$. There exists a constant $\const>0$ that depends only on the constants $\boundused$ and $\boundgt$, such that
    \begin{align}\label{eq:theorem_ub}
        \bias(\paramestbound{\boundused})  
         \le  \const\frac{\log\numitems+\log\numcomparisons}{\numitems\numcomparisons},
    \end{align}
    for all $\numitems \ge \constitems$ and all $\numcomparisons \ge \constcomparisons$, where $\constitems$ and $\constcomparisons$ are constants that depend only on the constants $\boundused$ and $\boundgt$.
\end{enumerate}
\end{subequations}
\end{theorem}

We note that in Theorem~\ref{thm:mle_ub_lb}\ref{part:ub}, we allow $\boundused$ to be any positive constant as long as $\boundused > \boundgt$. Therefore, the difference between $\boundused$ and $\boundgt$ can be any arbitrarily small constant. It is perhaps surprising that \stretching the box constraint only by a small constant yields such a significant improvement in the bias. We provide intuition behind this result in Section~\ref{sec:intuition_tradeoff}.

We devote the remainder of this section to providing a sketch of the proof of Theorem~\ref{thm:mle_ub_lb}. We first prove Theorem~\ref{thm:mle_ub_lb}\ref{part:ub} and then Theorem~\ref{thm:mle_ub_lb}\ref{part:lb}, because the proof of Theorem ~\ref{thm:mle_ub_lb}\ref{part:lb} depends on the proof of Theorem~\ref{thm:mle_ub_lb}\ref{part:ub}. The complete proof is provided in Appendix~\ref{sec:proof_bias_ub_lb}.

For Theorem~\ref{thm:mle_ub_lb}\ref{part:ub}, we first analyze the unconstrained \mle $\paramestunconstrain$. By plugging $\paramestunconstrain$ into the first-order optimality condition of the negative log-likelihood function and using concentration on the comparison outcomes, we prove an $\ell_\infty$ bound of the form $\norm{\paramestunconstrain  - \paramgt}_\infty = \bigOlog(\frac{1}{\sqrt{\numitems\numcomparisons}})$ with sufficiently high probability (which partially resolves the open problem from~\cite{chen2019topk}, in the regime where $\pobs = 1$). Next, using a second-order mean value theorem on the first-order optimality condition and taking an expectation, we show a result of the form $\norm{\Expect[\paramestunconstrain\given \eventintro] - \paramgt}_\infty \approx \norm{\paramestunconstrain - \paramgt}_\infty^2 = \bigOlog(\frac{1}{\numitems\numcomparisons})$, where $\eventintro$ is some high-probability event (recall from Table~\ref{tab:summary} that for unconstrained \mle, the bias $\norm{\Expect[\paramestunconstrain] - \paramgt}_\infty$ without conditioning on $\eventintro$ is undefined). Finally, we show that the unconstrained \mle $\paramestunconstrain$ and the \expandedmle $\paramestbound{\boundused}$ are identical with high probability for sufficiently large $\numitems$ and $\numcomparisons$, and perform some algebraic manipulations to finally arrive at the claim~\eqref{eq:theorem_ub}. 

For Theorem~\ref{thm:mle_ub_lb}\ref{part:lb}, we first prove a bound on the order of $\frac{1}{\sqrt{\numitems}}$ when there are $\numitems=2$ items. Then for general $\numitems$, we consider the bias on item $1$ under the true parameter vector $\paramgt = [\boundgt, -\frac{\boundgt}{\numitems-1}, \ldots, -\frac{\boundgt}{\numitems-1}]$. We construct an ``oracle'' \mle, such that analyzing the bias of the ``oracle'' \mle can be reduced to the proof of the $2$-item case, and thereby prove a bias on the order of $\frac{1}{\sqrt{\numitems\numcomparisons}}$ for the oracle \mle. Finally, we show that the difference between the oracle \mle and the \standardmle is small, by repeating arguments from the proof of Theorem~\ref{thm:mle_ub_lb}\ref{part:ub}.

\subsubsection{Intuition for Theorem~\ref{thm:mle_ub_lb}}\label{sec:intuition_tradeoff}
\begin{figure}
    \centering
    \subfloat[]{\includegraphics[height=\figheightintuition]{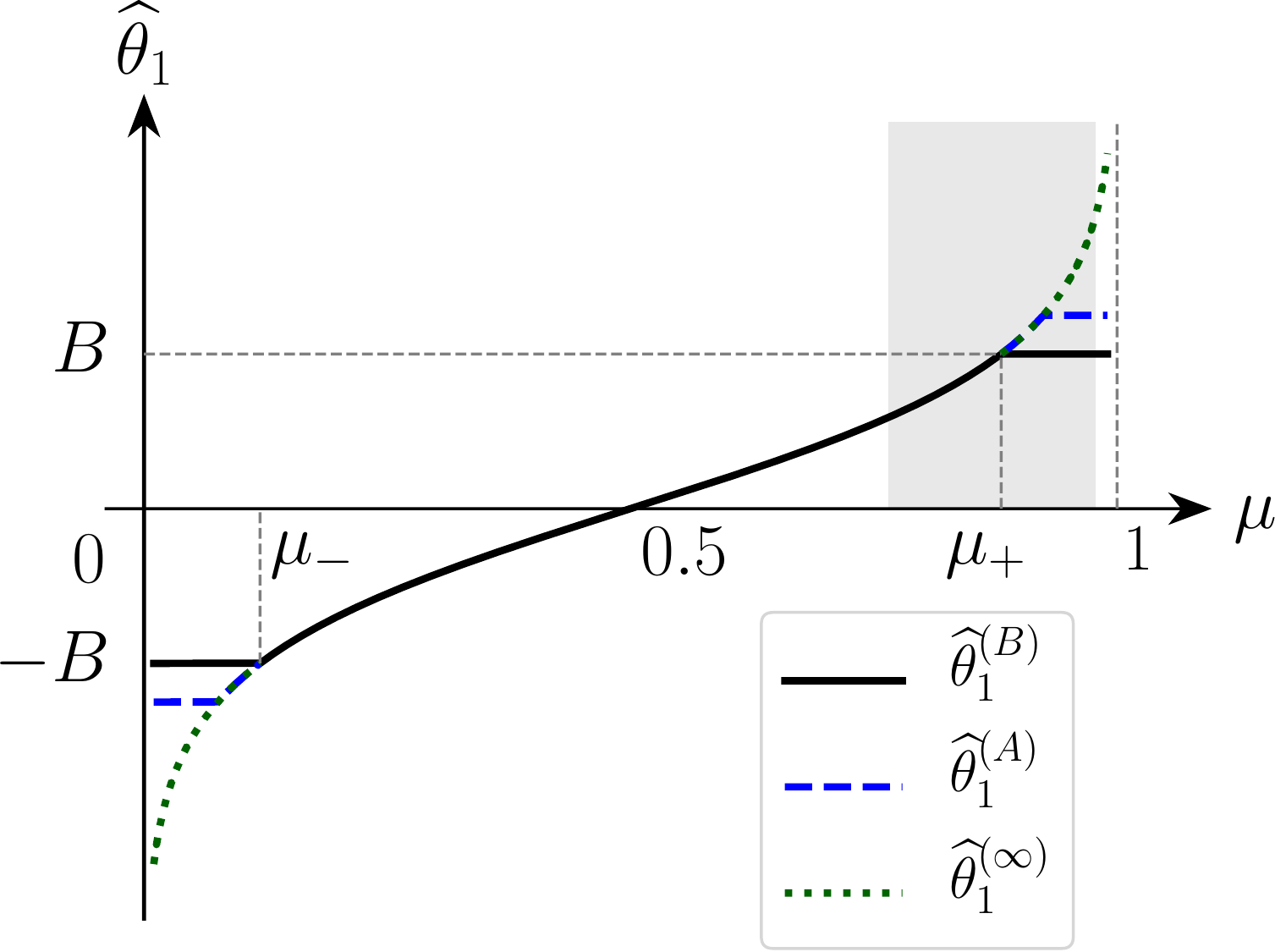}\label{fig:tradeoff_mle_plot}}~~~
    \subfloat[]{\includegraphics[height=\figheightintuition]{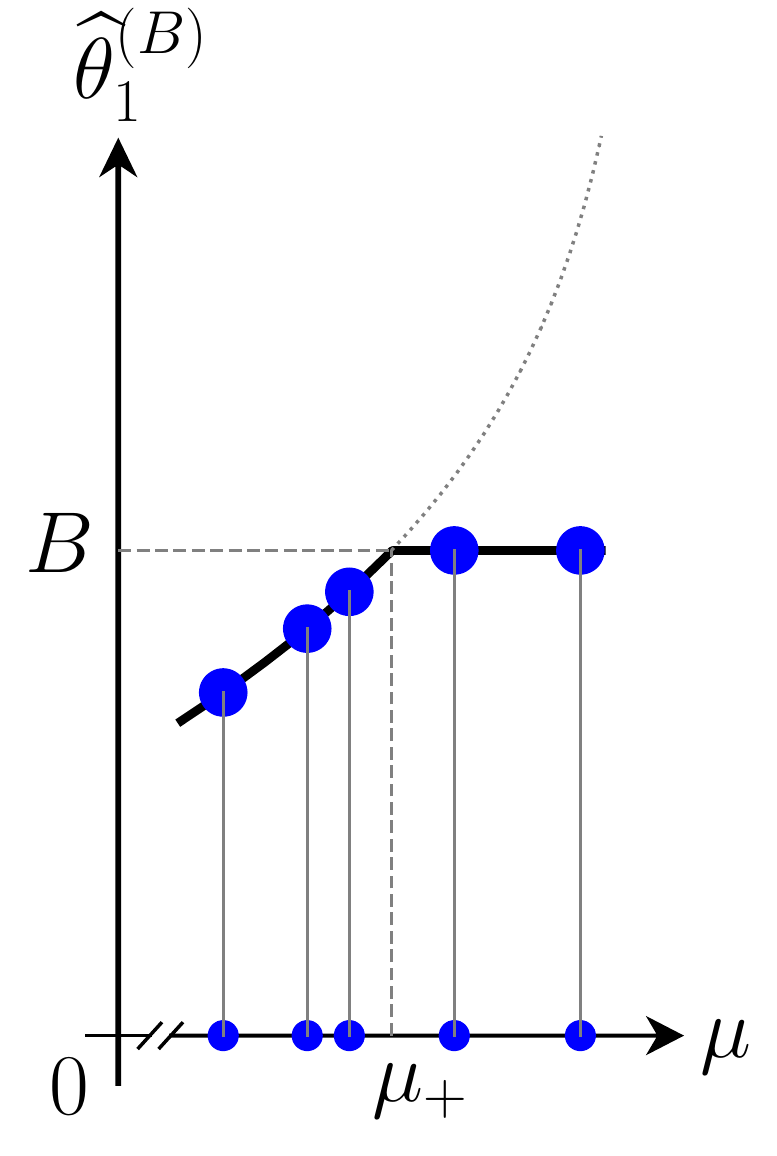}\label{fig:tradeoff_standard}}~~~
    \subfloat[]{\includegraphics[height=\figheightintuition]{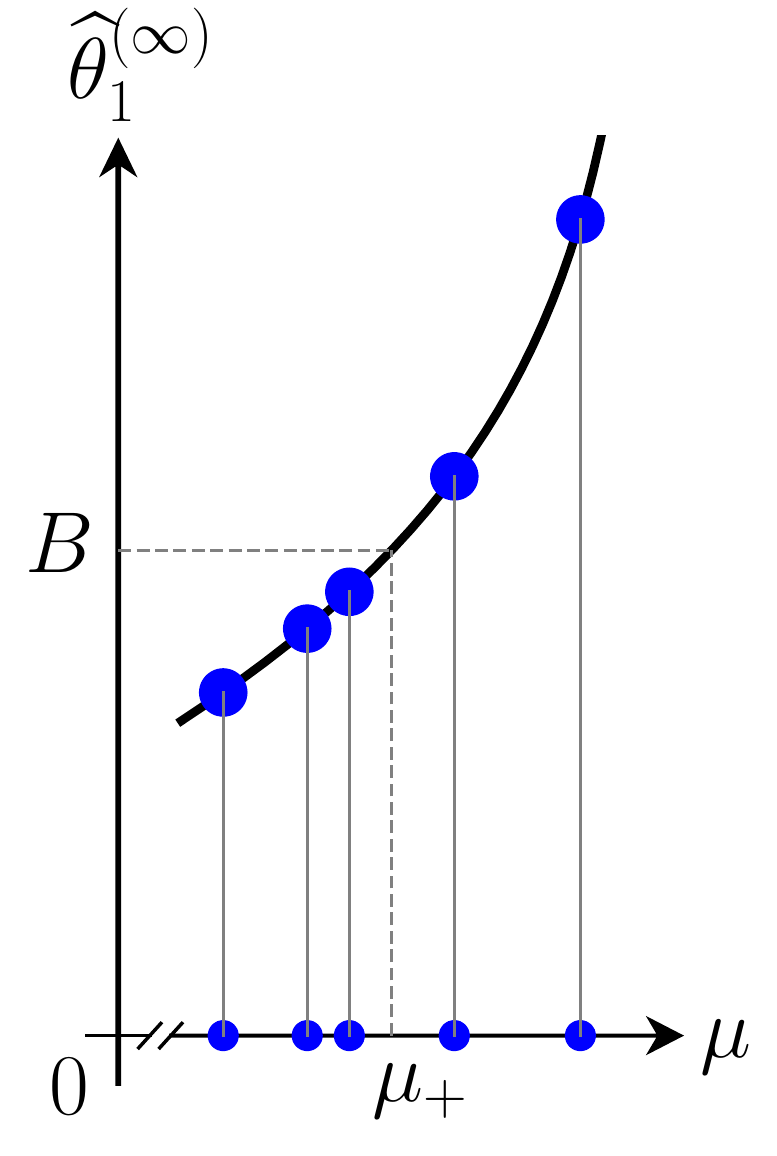}\label{fig:tradeoff_unconstrained}}~~~
    \subfloat[]{\includegraphics[height=\figheightintuition]{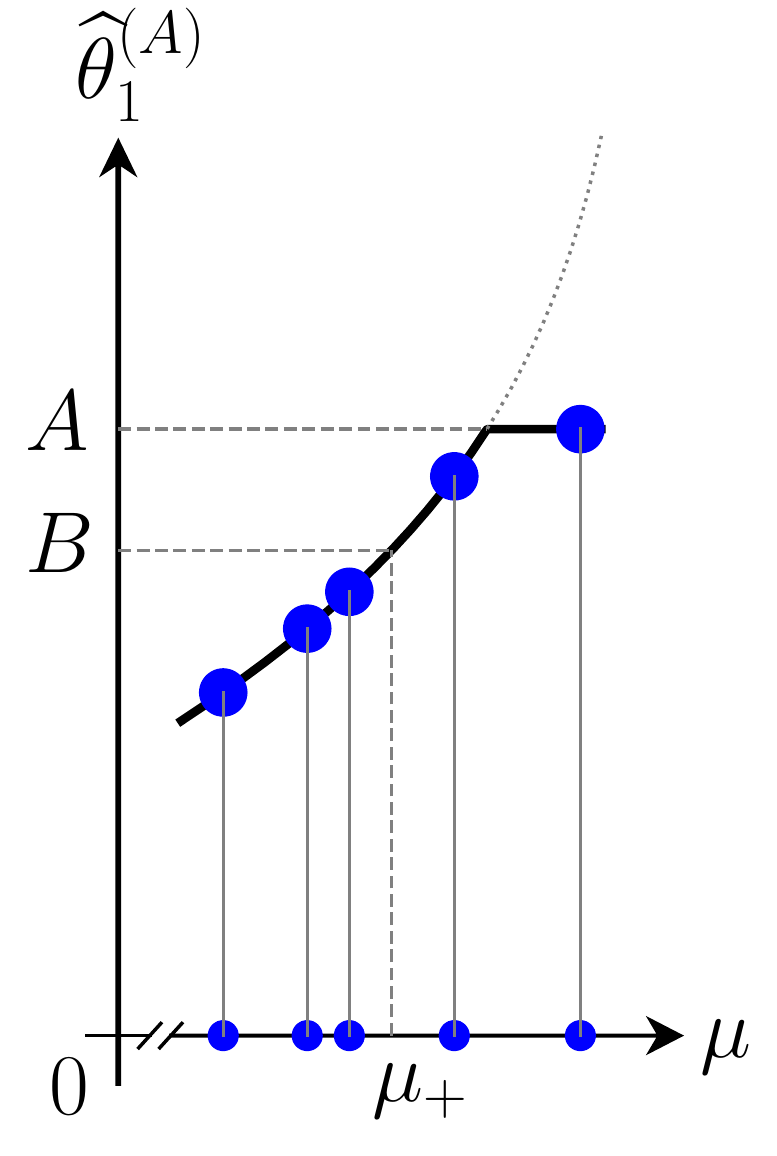}\label{fig:tradeoff_ours}}
    \caption{\label{fig:tradeoff_intuition}
        Intuition on the sources of bias. (a) The estimators \standardmle $\paramestboundgt$, \expandedmle $\paramestboundused$ and unconstrained \mle $\paramestunconstrain$ (on item $1$), as a function of $\meanemp$ when there are $\numitems=2$ items. We consider $\paramgt  = [\boundgt, -\boundgt]$, under which the true probability that item $1$ beats item $2$ is $\meanmax$. We zoom in to the region around $\mean = \meanmax$ indicated by the grey box. (b) The \standardmle $\paramestboundgt$ incurs a negative bias, because the estimate is required to be at most $\boundgt$. (c) The unconstrained \mle $\paramestunconstrain$ incurs a positive bias by Jensen's inequality, because the estimator function is convex on $\mean\in (0.5, 1)$. (d) Our estimator balances out the negative bias and the positive bias.
    }
\end{figure}

In this section, we provide intuition why \stretching the box constraint from $\boundgt$ to $\boundused$ significantly reduces the bias. Specifically, we consider a simplified setting with $\numitems=2$ items. Due to the centering constraint, we have $\paramgt_2 = -\paramgt_1$ for the true parameters, and we have $\paramest_2 = -\paramest_1$ for any estimator $\paramest$ that satisfies the centering constraint. Therefore, it suffices to focus only on item $1$. Denote $\meanemp$ as the random variable representing the fraction of times that item $1$ beats item $2$, and denote the true probability that item $1$ beats item $2$ as $\meangt \defn \frac{1}{1 + e^{-(\paramgt_1 - \paramgt_2)}}$. We consider the true parameter of item $1$ as $\paramgt_1\in [-\boundgt, \boundgt]$. Then we have $\meangt\in [\meanmin, \meanmax]$, where $\meanmin = \frac{1}{1 + e^{2\boundgt}}$ and $\meanmax = \frac{1}{1 + e^{-2\boundgt}}$. The \standardmle $\paramestboundgt$, the \expandedmle $\paramestboundused$ and the unconstrained \mle $\paramestunconstrain$ can be solved in closed form:
\begin{align*}
    \paramestboundgt_1(\mean) & = 
    \begin{cases}
        -\boundgt & \text{if } \mean\in [0, \meanmin]\\
        -\frac{1}{2}\log\left(\frac{1}{\mean} - 1\right) & \text{if } \mean\in (\meanmin, \meanmax)\\
        \boundgt & \text{if } \mean \in [\meanmax, 1].
    \end{cases}\\
    \paramestboundused_1(\mean) & = 
    \begin{cases}
        -\boundused & \text{if } \mean\in \left[0, \frac{1}{1 + e^{2\boundused}}\right]\\
        -\frac{1}{2}\log\left(\frac{1}{\mean} - 1\right) & \text{if } \mean\in \left(\frac{1}{1 + e^{2\boundused}}, \frac{1}{1 + e^{-2\boundused}}\right)\\
        \boundused & \text{if } \mean \in \left[\frac{1}{1 + e^{-2\boundused}}, 1\right].
    \end{cases}\\
    \paramestunconstrain_1(\mean) & = 
        -\frac{1}{2}\log\left(\frac{1}{\mean} - 1\right).
\end{align*}
See Fig.~\ref{fig:tradeoff_mle_plot} for a comparison of these three estimators.

Now we consider the bias incurred by these three estimators. For intuition, let us consider the case $\paramgt_1=\boundgt$, which incurs the largest bias in our simulation of Fig.~\ref{fig:example_bias}. If the observation $\meanemp$ were noiseless (and thus equals the true probability $\meanmax$), then all three estimators would output the true parameter $\boundgt$. However, the observation $\meanemp$ is noisy, and only concentrates around $\meanmax$. To investigate how these three estimators behave differently under this noise, we zoom in to the region around $\meanemp = \meanmax$ indicated by the grey box in Fig.~\ref{fig:tradeoff_mle_plot}. (Note that the observation $\meanemp$ can lie outside the grey box, but for intuition we ignore this low-probability event due to concentration.)

The behaviors of the three estimators in the grey box are shown in Fig.~\ref{fig:tradeoff_standard}, Fig.~\ref{fig:tradeoff_unconstrained} and Fig.~\ref{fig:tradeoff_ours}, respectively. For each of these estimators, the blue dots on the x-axis denotes the noisy observation of $\meanemp$ across different iterations, and the blue dots on the estimator function denotes the corresponding noisy estimates. The expected value of the estimator is a mean over the blue dots on the estimator function. For the \standardmle $\paramestboundgt$ (Fig.~\ref{fig:tradeoff_standard}), the box constraint requires that the estimate shall never exceed $\boundgt$. We call this phenomenon the ``clipping'' effect, which introduces a negative bias. For the unconstrained \mle $\paramestunconstrain$ (Fig.~\ref{fig:tradeoff_unconstrained}), since the estimator function is convex, by Jensen's inequality, the unconstrained \mle $\paramestunconstrain$ introduces a positive bias. Our proposed \expandedmle $\paramestboundused$ (Fig.~\ref{fig:tradeoff_ours}) lies in the middle between the \standardmle and the unconstrained \mle. Therefore, the \expandedmle balances out the negative bias from the ``clipping'' effect and the positive bias from the convexity of the estimator function, thereby yielding a smaller bias on the item parameter. In practice, one can numerically tune the parameter $\boundused$ to minimize the bias across all possible parameter vector $\paramgt\in \rangeboxgt$. Simulation results on different values of $\boundused$ are included in Section~\ref{sec:simulation}.

%%%%%%
\subsection{Accuracy}\label{sec:result_l2}
Given the result of Theorem~\ref{thm:mle_ub_lb} on the bias reduction of the estimator $\paramestbound{\boundused}$, we revisit the mean squared error. Past work~\cite{hajek2014likelihood, shah2016topology} has shown that the standard \mle $\paramestboundgt$ is minimax-optimal in terms of the mean squared error. The following theorem shows that this minimax-optimality also holds for our proposed \expandedmle $\paramestbound{\boundused}$, where $\boundused$ is any constant such that $\boundused > \boundgt$. The theorem statement and its proof follows Theorem 2 from~\cite{shah2016topology}, after some modification to accommodate the bounding box parameter $\boundused$.

\begin{theorem}\label{thm:l2_ub_lb}
\begin{subequations}
\begin{enumerate}[(a)]
    \item \label{part:l2_lb}
    [Theorem 2(a) from~\cite{shah2016topology}]
    There exists a constant $\const>0$ that depends only on the constant $\boundgt$, such that any estimator $\paramest$ has a mean squared error lower bounded as
        \begin{align}\label{eq:thm_l2_lb}
            \accuracy(\paramest) \ge \frac{\const}{\numcomparisons},
        \end{align}
        for all $\numcomparisons\ge \constcomparisons$, where $\constcomparisons$ is a constant that depends only on the constant $\boundgt$.
        
    %%%%%%
    \item \label{part:l2_ub}
    Let $\boundused$ be any finite constant such that $\boundused > \boundgt$.
    There exists a constant $\const >0$ that depends only on the constants $\boundused$ and $\boundgt$, such that
        \begin{align}\label{eq:thm_l2_ub}
            \accuracy(\paramestbound{\boundused}) \le \frac{\const}{\numcomparisons}.
        \end{align}
\end{enumerate}
\end{subequations}
\end{theorem}

Theorem~\ref{thm:l2_ub_lb} shows that using the estimator $\paramestbound{\boundused}$ retains the minimax-optimality achieved by $\paramestboundgt$ in terms of the mean squared error. Combining Theorem~\ref{thm:mle_ub_lb} and Theorem~\ref{thm:l2_ub_lb} shows the Pareto improvement of our estimator $\paramestbound{\boundused}$: the estimator $\paramestbound{\boundused}$ decreases the rate of the bias, while still performing optimally on the mean squared error.

The proof of Theorem~\ref{thm:l2_ub_lb} closely mimics the proof of Theorem 2(b) from~\cite{shah2016topology}, replacing the steps involving the domain $\rangeboxgt$ by the stretched domain $\rangeboxused$. The details are provided in \App~\ref{sec:proof_l2_ub_lb}. 

%%%%%%
\section{Simulations}\label{sec:simulation}
In this section, we explore our problem space and compare the standard \mle and our proposed \expandedmle by simulations. In what follows, we set $\boundgt = 1$, and unless specified otherwise we set $\boundused=2$ and $\paramgt = [1, -\frac{1}{\numitems-1}, -\frac{1}{\numitems-1}, \ldots, -\frac{1}{\numitems-1}]$. We also evaluate the performance of other values of $\paramgt$ subsequently. Error bars in all the plots represent the standard error of the mean.

\begin{enumeratex}[(i)]
\item \label{item:simulation_vary_d}
\textbf{Dependence on $\numitems$:}
We vary the number of items $\numitems$, while fixing $\numcomparisons=5$. The results are shown in Fig.~\ref{fig:rate_d}. Observe that the \expandedmle has a significantly smaller bias, and performs on par with the \mle in terms of the mean squared error when $\numitems$ is large. Moreover, the simulations also suggest the rate of bias as of order $\frac{1}{\sqrt{\numitems}}$ for the \mle and $\frac{1}{\numitems}$ for the \expandedmle, as predicted by our theoretical results.

\begin{figure}
    \centering
    \subfloat[Bias]{\includegraphics[height=\figheightexp]{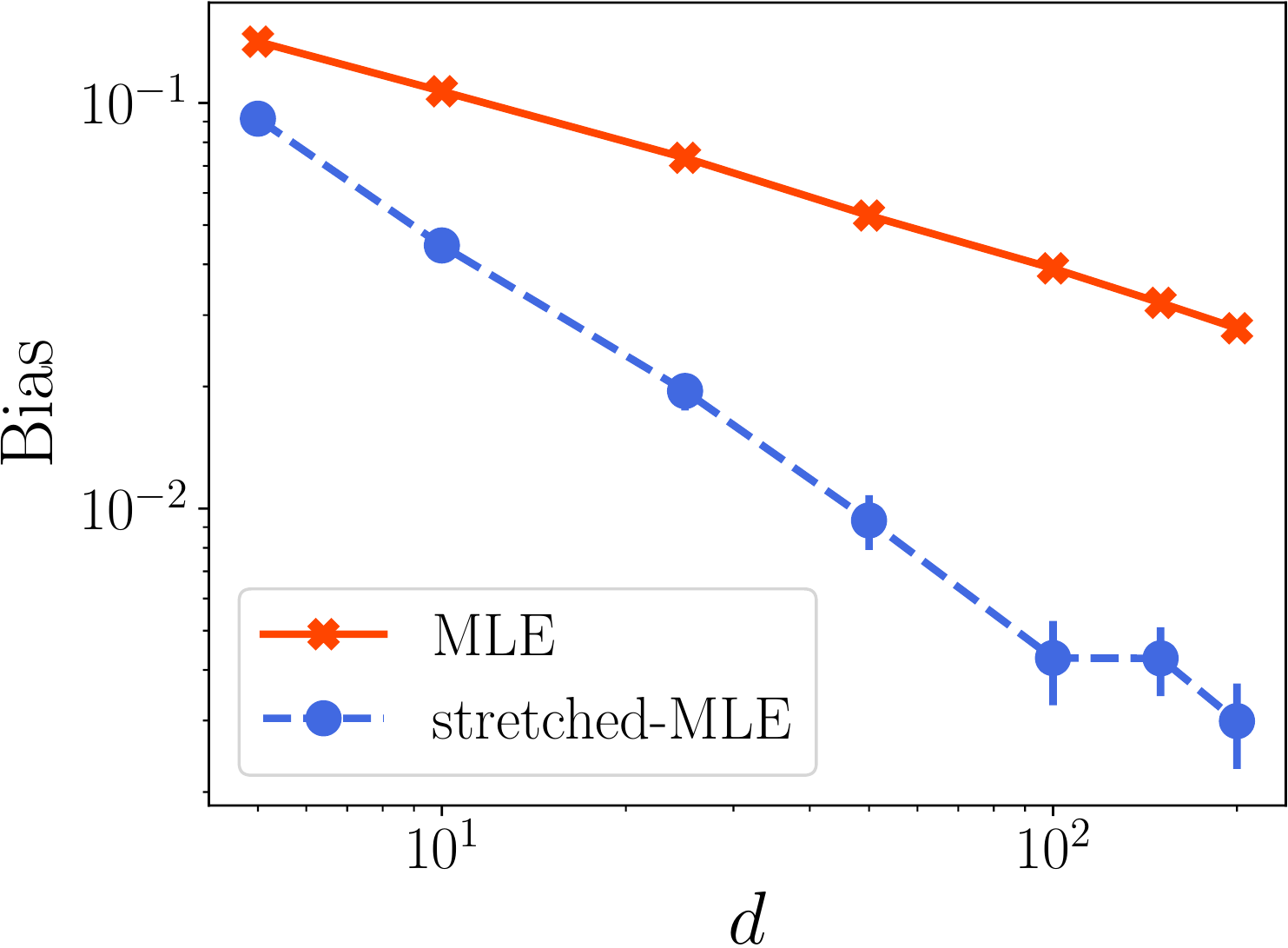}\label{fig:vary_d_bias}}~~~~~~~
    \subfloat[Mean squared error]{\includegraphics[height=\figheightexp]{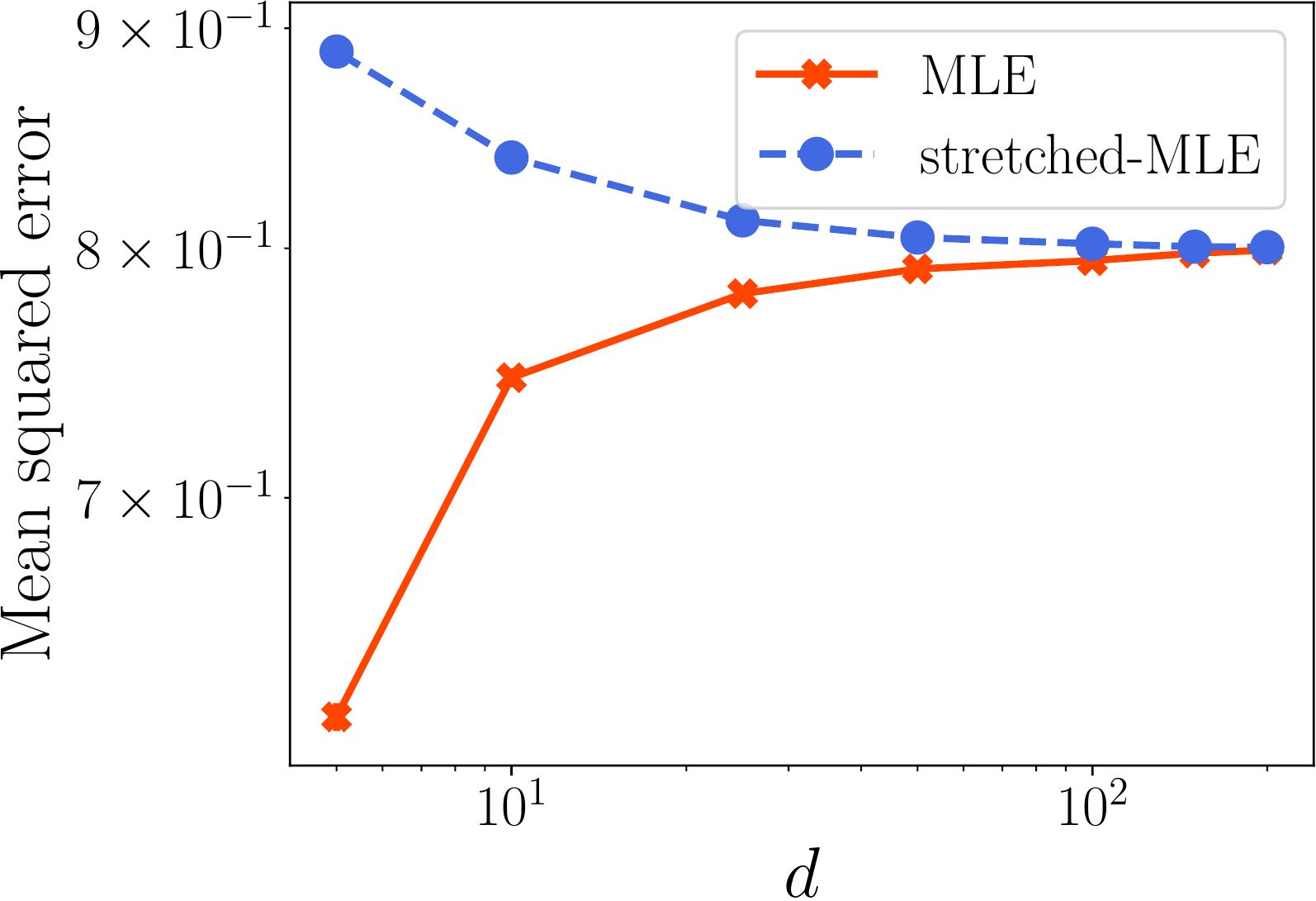}\label{fig:vary_d_mse}}
    \caption{\label{fig:rate_d}
        Performance of estimators for various values of $\numitems$, with $\numcomparisons=5$ and $\boundused = 2$. Each point is a mean over $10000$ iterations.
    }
\end{figure}

\begin{figure}
    \centering
    \subfloat[Bias]{\includegraphics[height=\figheightexp]{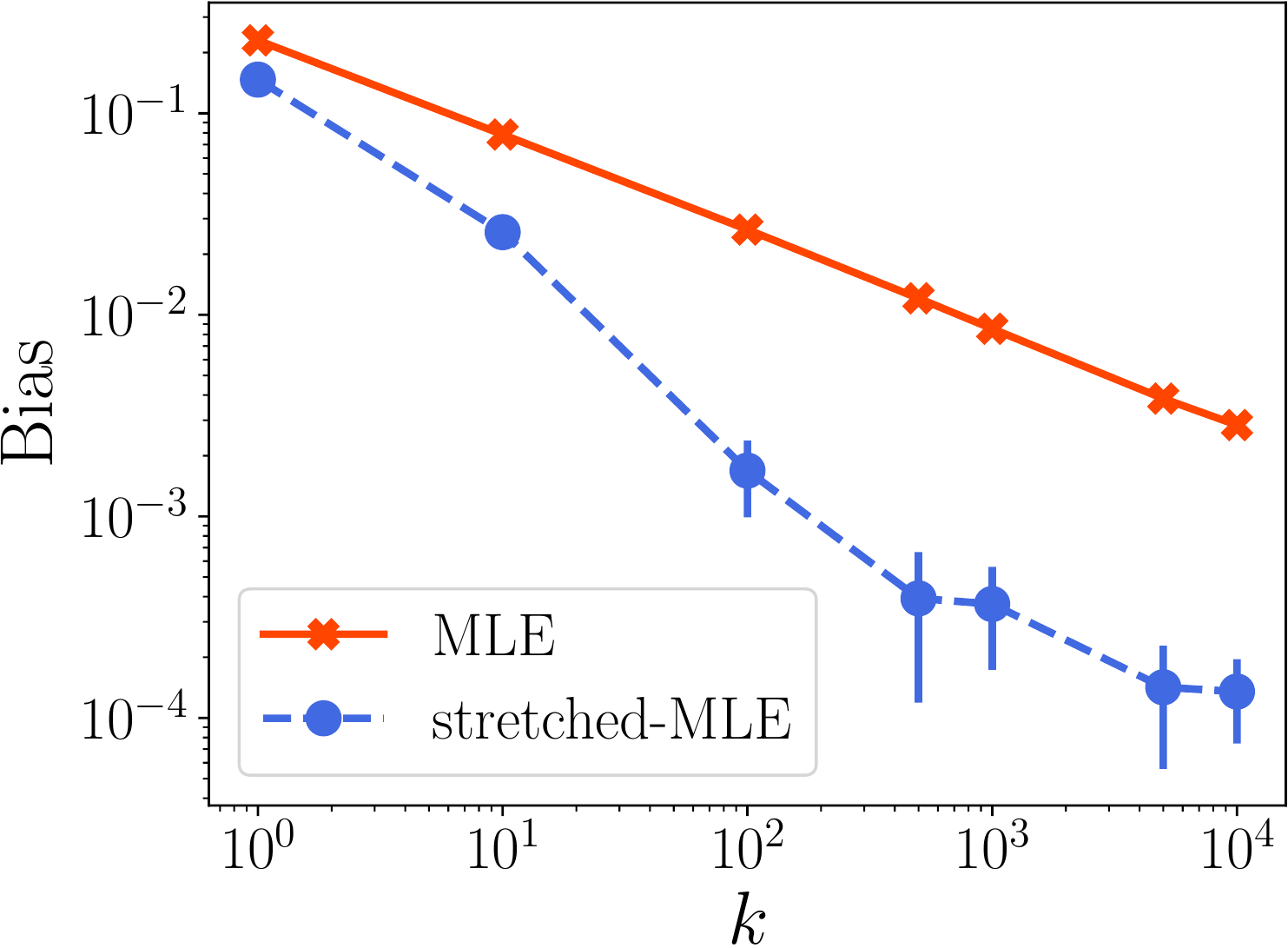}\label{fig:vary_k_bias}}~~~~~~~
    \subfloat[Mean squared error]{\includegraphics[height=\figheightexp]{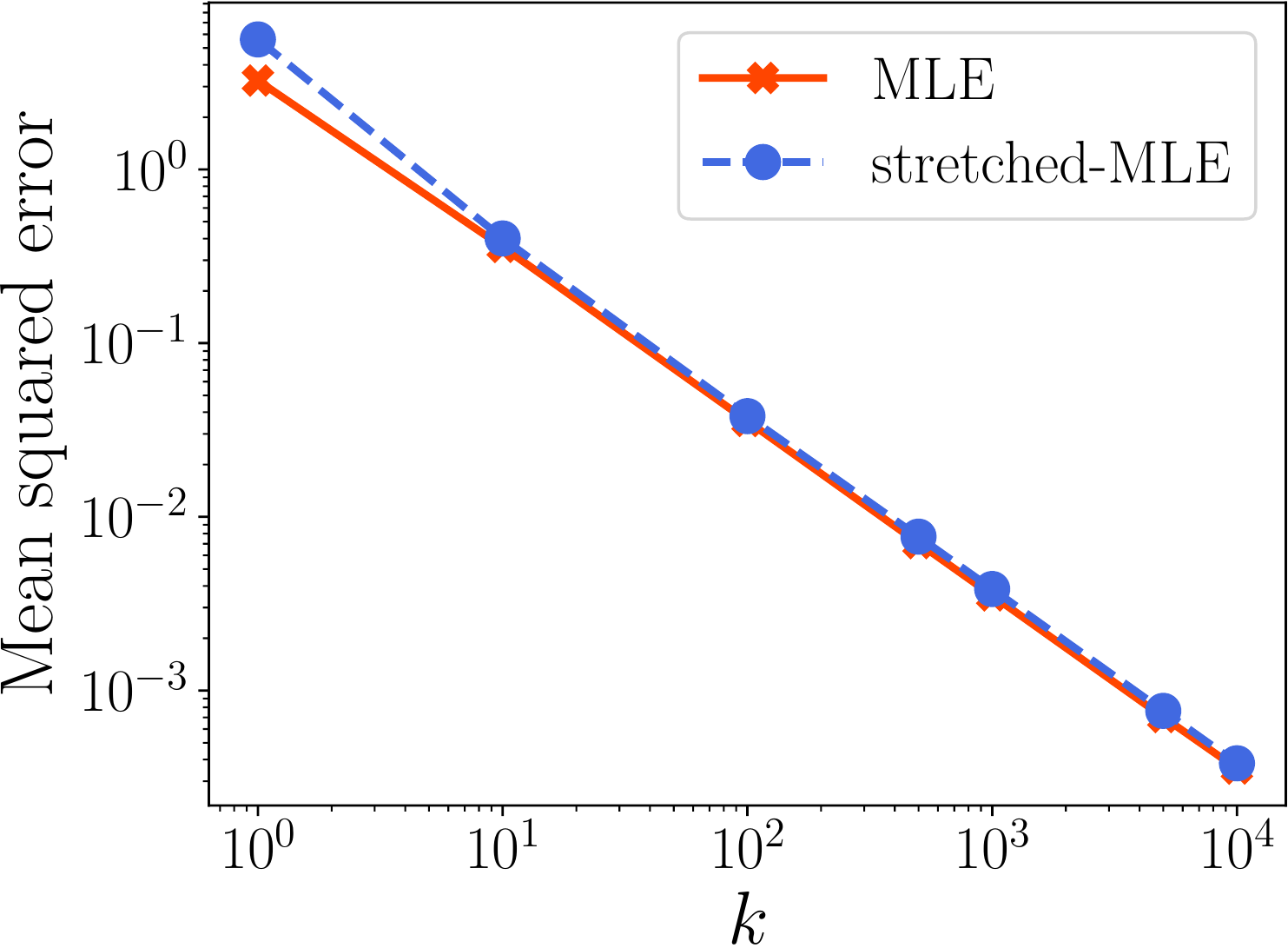}\label{fig:vary_k_mse}}
    \caption{\label{fig:rate_k}
        Performance of estimators for various values of $\numcomparisons$, with $\numitems=10$ and $\boundused = 2$. Each point is a mean over $10000$ iterations.
    }
\end{figure}

%%%
\item \textbf{Dependence on $\numcomparisons$:} We vary the number of comparisons $\numcomparisons$ per pair of items, while fixing $\numitems=10$. The results are shown in Fig.~\ref{fig:rate_k}. As in the simulation~\ref{item:simulation_vary_d} with varying $\numitems$, we observe that the \expandedmle has a significantly smaller bias, and performs on par with the \mle in terms of the mean squared error. Moreover, the simulations also suggest the rate of bias as of order $\frac{1}{\sqrt{\numcomparisons}}$ for the \mle and $\frac{1}{\numcomparisons}$ for the \expandedmle, as predicted by our theoretical results.

%%%
\item \textbf{Different values of $\boundused$:} In our theoretical analysis, we proved bounds that hold for all constant $\boundused$ such that $\boundused > \boundgt$. In this simulation, we empirically compare the performance of the \expandedmle for different values of $\boundused$ (note that setting $\boundused=1$ is equivalent to the standard \mle). We fix $\numitems=10$, varying $\boundused\in [0.5, 3]$ and $\numcomparisons$ from $1$ to $100$. The results are shown in Fig.~\ref{fig:rate_A}. For the bias, we observe that the bias keeps decreasing in the range of $\boundused\in [0.5, 1]$. This is because as we increase $\boundused$ to $1$, the negative bias introduced by the ``clipping'' effect is reduced. The optimal value of $\boundused$ for all settings of $\numcomparisons$ is always greater than $1$. Moreover, the optimal $\boundused$ seems to be closer to $1$ when we increase $\numcomparisons$. This agrees with the intuition in Section~\ref{sec:intuition_tradeoff}. When $\numcomparisons$ is larger, the estimate becomes more concentrated around the true parameter. Then the ``clipping'' effect becomes smaller and can be accommodated by a smaller $\boundused$. The mean squared error is insensitive to the choice of $\boundused$ as long as $\boundused\ge 1$.

\begin{figure}
    \centering
    \includegraphics[height=\figheightlegend]{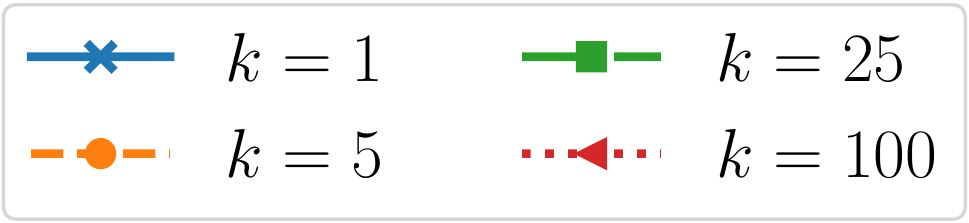}\vspace{-5pt}
    \subfloat[Bias]{\includegraphics[height=\figheightexp]{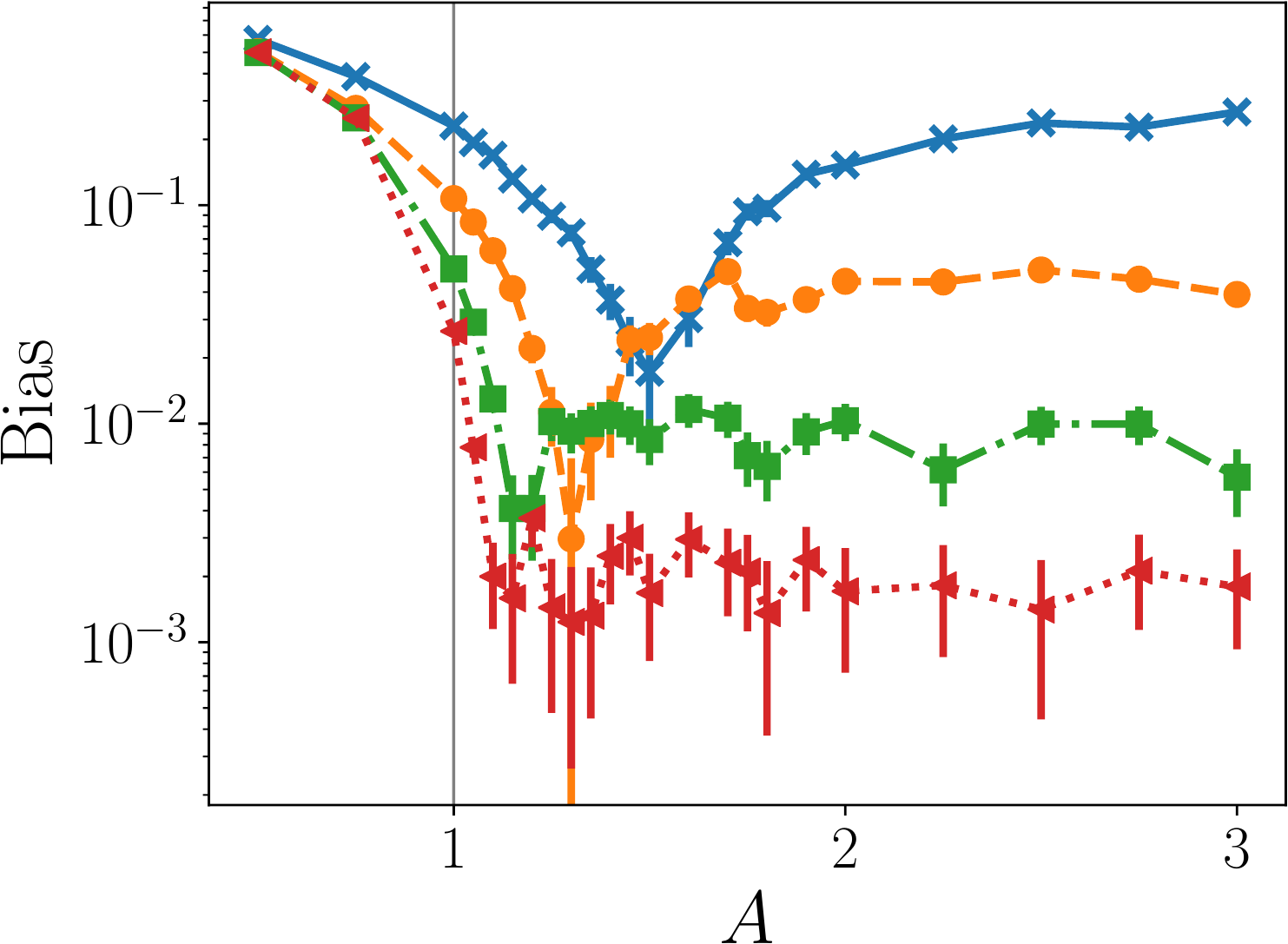}\label{fig:vary_A_bias}}~~~~~~~
    \subfloat[Mean squared error]{\includegraphics[height=\figheightexp]{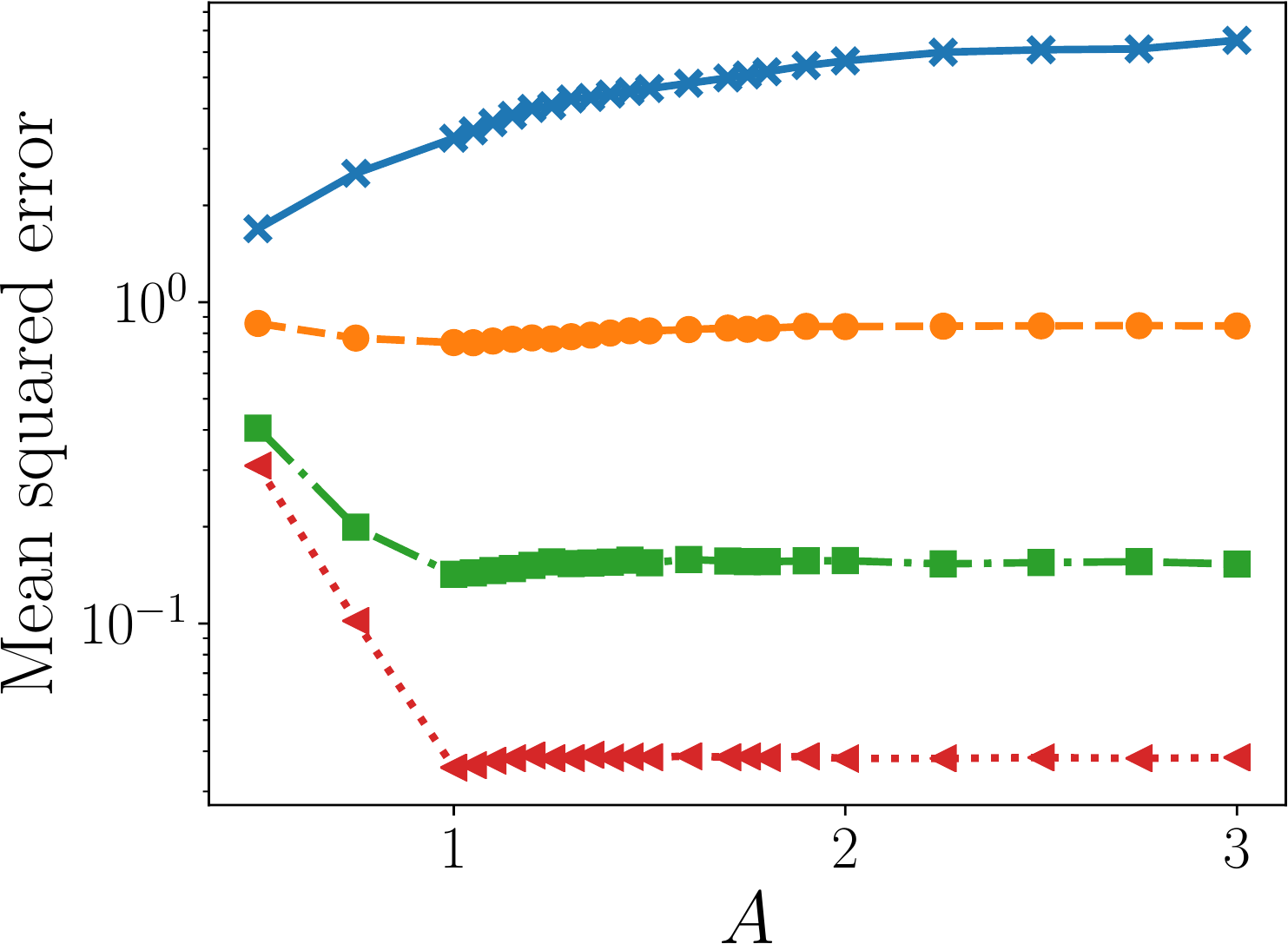}\label{fig:vary_A_mse}}
    \caption{\label{fig:rate_A}
    Performance of estimators for various values of $\boundused$ and $\numcomparisons$, with $\numitems=10$. Setting $\boundused=1$ is equivalent to the standard \mle. Each point is a mean over $5000$ iterations.
    }
\end{figure}

\begin{figure}
    \centering
    \subfloat[Bias]{\includegraphics[height=\figheightexp]{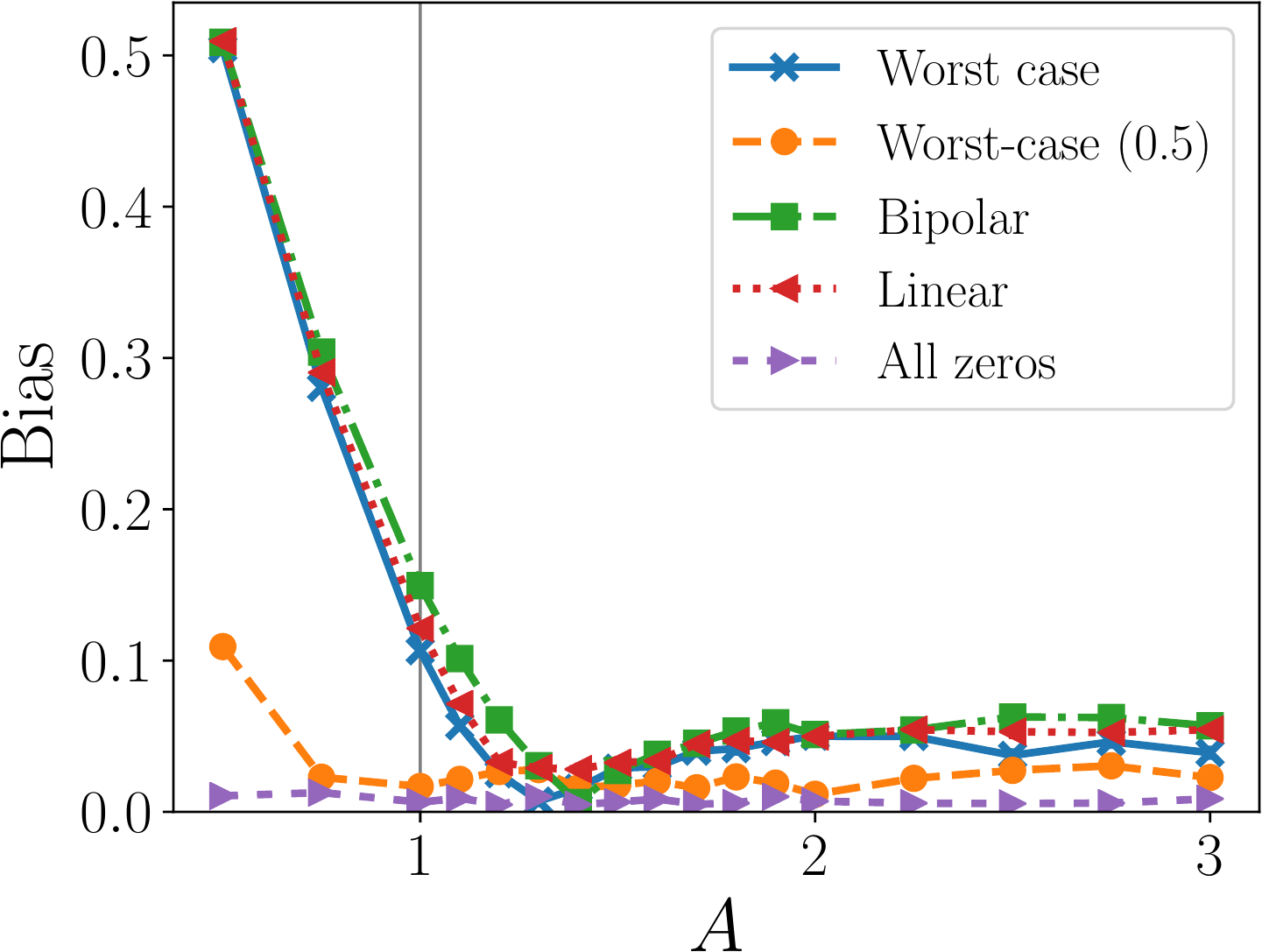}\label{fig:vary_A_theta_bias}}~~~~~~~
    \subfloat[Mean squared error]{\includegraphics[height=\figheightexp]{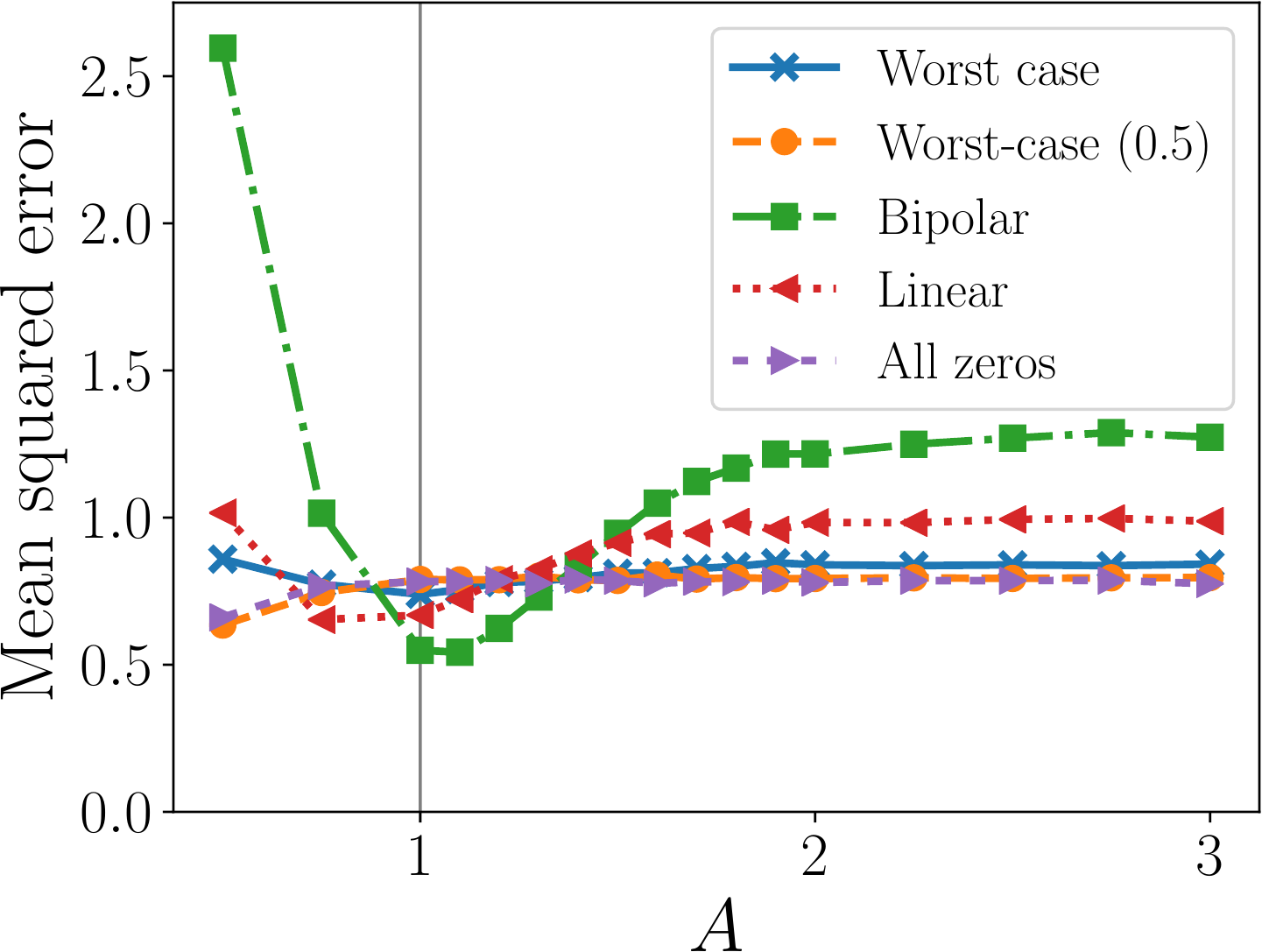}\label{fig:vary_A_theta_mse}}
    \caption{\label{fig:rate_A_theta}
        Performance of estimators for various values of $\boundused$ and various settings of $\paramgt$, with $\numitems=10$ and $\numcomparisons=5$. Setting $\boundused=1$ is equivalent to the standard \mle. Each point is a mean over $5000$ iterations.
    }
\end{figure}

%%%
\item \textbf{Different settings of the true parameter $\paramgt$:} Our theoretical result considers the worst-case bias and accuracy. In this simulation, we empirically compare the performance of the \expandedmle under different settings of the true parameter vector $\paramgt$ (again, recall that setting $\boundused=1$ is equivalent to the \standardmle). Specifically, we consider the following values of $\paramgt$:
\begin{itemizex}
    \item \emph{Worst case:} $\paramgt = [1, -\frac{1}{\numitems-1}, \ldots, -\frac{1}{\numitems-1}]$.
    
    \item \emph{Worst case (0.5): } $\paramgt = [0.5, -\frac{0.5}{\numitems-1}, \ldots, -\frac{0.5}{\numitems-1}]$.
    
    \item \emph{Bipolar:} half of the values are $1$, and the other half are $-1$.
    
    \item \emph{Linear:} the values are equally spaced in the interval $[-1, 1]$.
    
    \item \emph{All zeros: } all parameters are $0$.
\end{itemizex}

We fix $\numitems=10$ and $\numcomparisons = 5$, varying $\boundused\in [0.5, 3]$ under different settings of the true parameter vector $\paramgt$. The results are shown in Fig.~\ref{fig:rate_A_theta}. Two high-level takeaways from the empirical evaluations are that the bias generally reduces with an increase in $\boundused$ till past $\boundgt$, and that the mean squared error remains relatively constant beyond $\boundused = 1$ in the plotted range. In more detail, for the bias, we observe that the performance primarily depends on the largest magnitude of the items (that is, $\norminf{\paramgt}$). For the settings \emph{worst case, bipolar} and \emph{linear} (where $\norminf{\paramgt} = 1$), the bias keeps decreasing when A is past $\boundgt=1$. For the setting \emph{worst-case (0.5)} (where $\norminf{\paramgt} = 0.5$), the bias keeps decreasing when A is past $0.5$. This makes sense since in this case we effectively have $B = 0.5$ (although the algorithm would not know this in practice). The bias for the setting \emph{all zeros} stays small across values of $\boundused$. For the mean squared error, the increase when A is past $1$ is relatively small under most of the settings of the true parameter vector $\paramgt$. The \emph{bipoloar} setting has the largest increase in the mean squared error. Under this setting, all parameters $\paramgt_\idxitem$ take values at the boundaries $\pm \boundgt$, and therefore the estimates of all parameters are affected by the box constraint.

%%%
\item \label{item:simulation_sparse}
\textbf{Sparse observations:} So far we have considered a league format where $\numcomparisons$ comparisons are observed between any pair of items. Now we consider a random-design setup, where $\numcomparisons$ comparisons are observed between any pair of items independently with probability $\pobs\in (0, 1)$, and none otherwise~\cite{negahban2016centrality,chen2019topk}. In our simulations, we set $\pobs = \frac{1}{\sqrt{\numitems}}$ and $\numcomparisons=5$. We discard an iteration if the graph is not connected, since the problem is not identifiable under such a graph. The results are shown in Figure~\ref{fig:sparse}. We observe that the \expandedmle continues to outperform \mle in terms of bias, and perform on par in terms of the mean squared error.

\begin{figure}
    \centering
    \subfloat[Bias]{\includegraphics[height=\figheightexp]{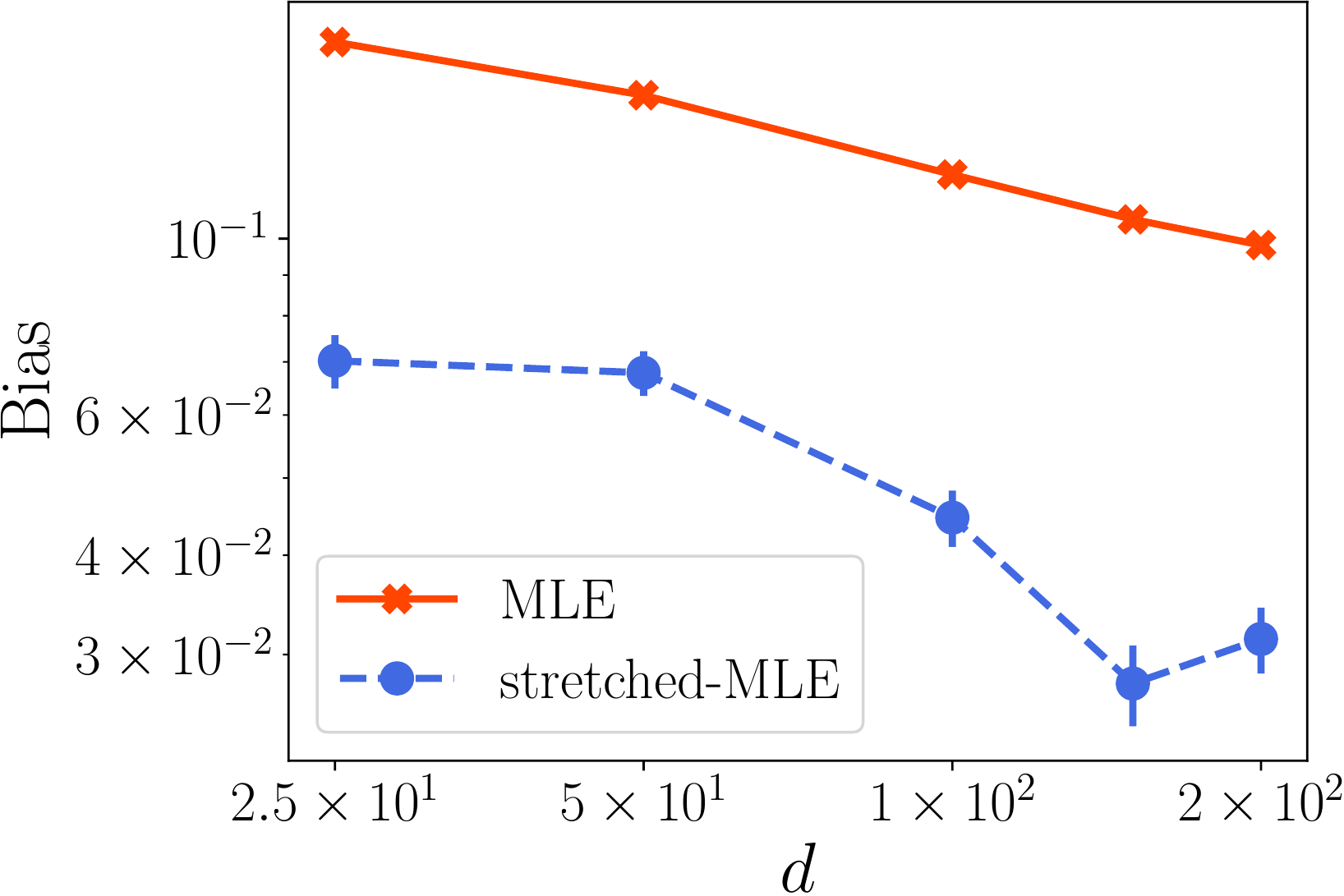}\label{fig:sparse_bias}}~~~~~~~
    \subfloat[Mean squared error]{\includegraphics[height=\figheightexp]{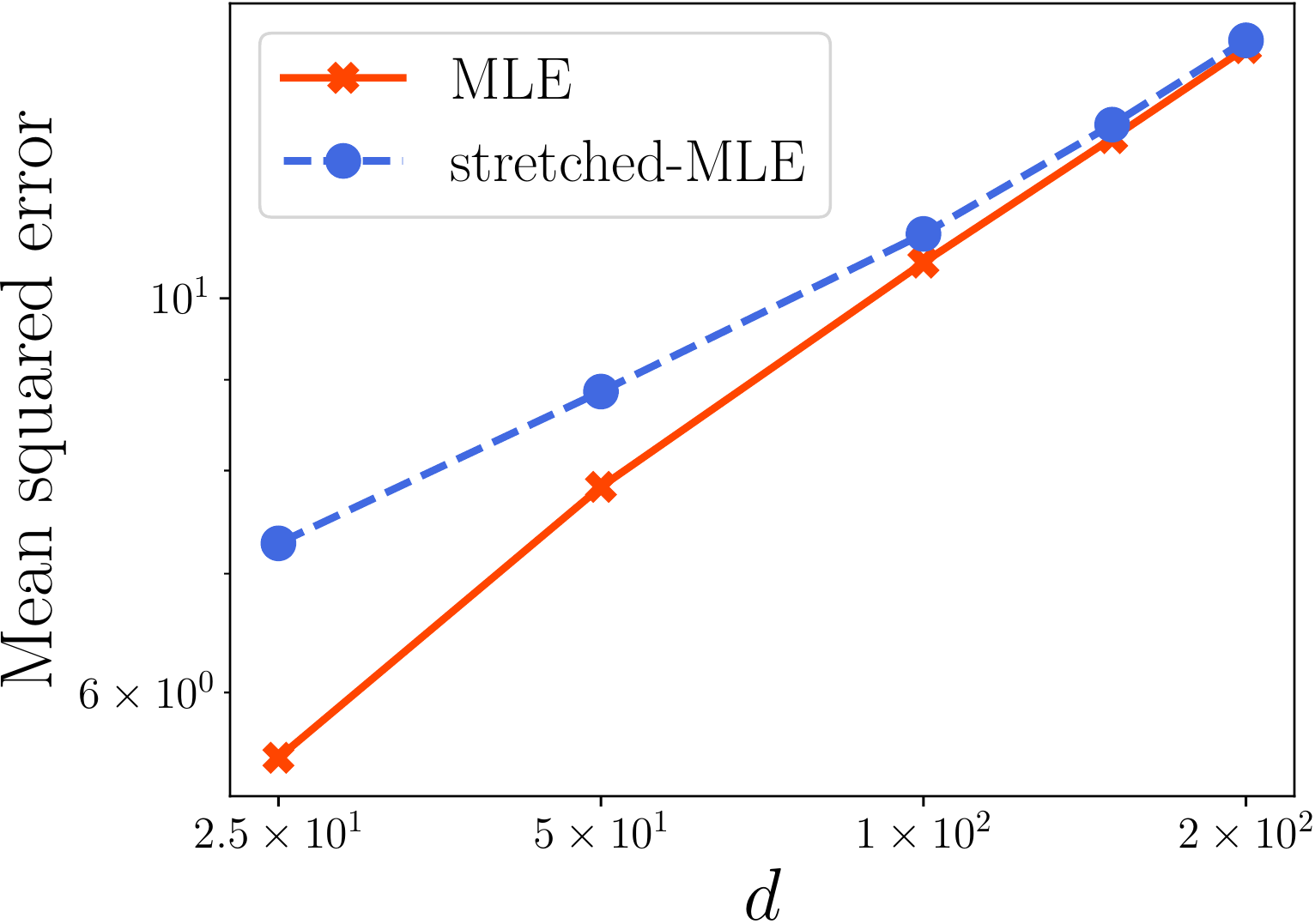}\label{fig:sparse_mse}}
    \caption{\label{fig:sparse}
    Performance of estimators for various values of $\numitems$ under sparse observations, with $\boundused=2$. A number of $\numcomparisons=5$ comparisons are observed between any pair independently with probability $\pobs = \frac{1}{\sqrt{\numitems}}$ and none otherwise. Each point is a mean over $10000$ iterations.
    }
\end{figure}

\end{enumeratex}

%%%%%%
\section{Conclusions and discussions}\label{sec:conclusion}

In this work, we show that the widely-used \mle is suboptimal in terms of bias, and propose a class of estimators called the ``\expandedmle'', which provably reduces the bias while maintaining the minimax-optimality in terms of accuracy.
These results on the performance of the \mle and the \expandedmle are of both theoretical and practical interest. From the theoretical point of view, our analysis and proofs provide insights on the cause of the bias, explain why \stretching the box alleviates this cause, and prove theoretical guarantees in bias reduction by \stretching the box. Our results on the benefits of the \expandedmle thus suggest theoreticians to consider the \expandedmle for analysis instead of the standard \mle. 

From the practical point of view, the constant $\boundgt$ is often unknown, and practitioners oten estimate the value of $\boundgt$ by fitting the data or from past experience. Our results thus suggest that one should estimate $\boundgt$ leniently, as an estimation smaller than or equal to the true $\boundgt$ causes significant bias. Moreover, our proposed estimator is a simple modification to the \mle, which can be incorporated into any existing implementation at ease.

Our results lead to several open problems. First, it is of interest to extend our theoretical analysis to settings where the observations are sparse. For example, one may consider a random-design setup, where $\numcomparisons$ comparisons are observed between any pair independently with probability $\pobs$ and none otherwise~\cite{negahban2016centrality,chen2019topk} (also see simulation~\ref{item:simulation_sparse} in Section~\ref{sec:simulation}). In terms of the bias under this random-design setup, we think that the lower-bound for \mle and the upper-bound for our \expandedmle  also depend on $\numitems$  and $\numcomparisons$ as  $\bigOmega(\frac{1}{\sqrt{\numitems\numcomparisons}})$ and $\bigOlog(\frac{1}{\numitems\numcomparisons})$  respectively; we also think that the dependence of the \expandedmle on $\pobs$ is no worse than that of the \standardmle. Second, it is of interest to extend our results to other parametric models such as the Thurstone model~\cite{thurstone1927comparative}, and we envisage similar results to hold across a variety of such models. Finally, the ideas and techniques developed in this paper may also help in improving the Pareto efficiency on other learning and estimation problems, in terms of the bias-accuracy tradeoff.

\section*{Acknowledgements}
The work of JW and NBS was supported in part by NSF grants 1755656 and 1763734. The work of RR was supported in part by NSF grant 1527032.

{
\small
\bibliographystyle{plain}
\bibliography{references}
}

\appendix

%% Proofs 

%%%%%%%%%%%%%
\section{Proof of Theorem~\ref{thm:mle_ub_lb}}\label{sec:proof_bias_ub_lb}

In this \app, we present the proof of Theorem~\ref{thm:mle_ub_lb}. We first introduce notation and preliminaries in \App~\ref{sec:preliminaries}, to be used subsequently in proving both parts of Theorem~\ref{thm:mle_ub_lb}. The proof of Theorem~\ref{thm:mle_ub_lb}\ref{part:ub} is presented in \App~\ref{sec:proof_ub}. The proof of Theorem~\ref{thm:mle_ub_lb}\ref{part:lb} is presented in \App~\ref{sec:proof_lb}. We first present the proof of Theorem~\ref{thm:mle_ub_lb}\ref{part:ub} followed by Theorem~\ref{thm:mle_ub_lb}\ref{part:lb}, because the proof of Theorem~\ref{thm:mle_ub_lb}\ref{part:lb} depends on the proof of Theorem~\ref{thm:mle_ub_lb}\ref{part:ub}.

In the proof of Theorem~\ref{thm:mle_ub_lb}\ref{part:lb}, the constants are allowed to depend only on the constant $\boundgt$. In the proof of Theorem~\ref{thm:mle_ub_lb}\ref{part:ub}, the constants are allowed to depend only on the constants $\boundused$ and $\boundgt$. The proofs for all the lemmas are presented in \App~\ref{sec:proof_lemma}.

\subsection{Notation and preliminaries}\label{sec:preliminaries}

In this \app, we introduce notation and preliminaries that are used subsequently in the proofs of both Theorem~\ref{thm:mle_ub_lb}\ref{part:ub} and Theorem~\ref{thm:mle_ub_lb}\ref{part:lb}.

\begin{enumeratex}[label={\bfseries (\roman*)}, ref=\roman*]
    %%%
    \item \textbf{Notation}
    
    Recall that $\numitems$ denotes the number of items, and $\numcomparisons$ denotes the number of comparisons per pair of items. The $\numitems$ items are associated to a true parameter vector $\paramgt = [\paramgt_1, \ldots, \paramgt_\numitems]$. We have the set $\rangeparam_\boundgt = \{\param\in \reals^\numitems \given \norm{\param}_\infty \le \boundgt, \sum_{\idxitem=1}^\numitems \param_\idxitem = 0 \}$ and the set $\rangeparam_\boundused = \{\param\in \reals^\numitems \given \norm{\param}_\infty \le \boundused, \sum_{\idxitem=1}^\numitems \param_\idxitem = 0 \}$, where $\boundused$ and $\boundgt$ are finite constants such that $\boundused>\boundgt > 0$. The true parameter vector satisfies $\paramgt\in \rangeboxgt$.

Denote $\meangt_{\idxitem\idxitemalt}$ as the probability that item $\idxitem\in [\numitems]$ beats item $\idxitemalt\in [\numitems]$. Under the BTL model, we have
\begin{align}
    \meangt_{\idxitem\idxitemalt}= \frac{1}{1+ e^{-(\paramgt_\idxitem-\paramgt_\idxitemalt)}}\label{eq:btl}.
\end{align} 
For every $\idxcomparison\in [\numcomparisons]$, denote the outcome of the $\idxcomparison^{th}$ comparison between item $\idxitem\in [\numitems]$ and item $\idxitemalt\in [\numitems]$ as \begin{align*}
    \resultcomparison_{\idxitem\idxitemalt}^{(\idxcomparison)} \defn \indicator\{\text{item $\idxitem$ beats item $\idxitemalt$ in their $\idxcomparison^{th}$ comparison}\}.
\end{align*}
We have $\resultcomparison_{\idxitem\idxitemalt}^{(\idxcomparison)}  \sample \Bernoulli(\meangt_{\idxitem\idxitemalt})$, independent across all $\idxcomparison\in [\numcomparisons]$ and all $\idxitem < \idxitemalt$. Recall that $\numwins_{\idxitem\idxitemalt}$ denotes the number of times that item $\idxitem $ beats $\idxitemalt$. We have $\numwins_{\idxitem\idxitemalt}= \sum_{\idxcomparison=1}^\numcomparisons \resultcomparison_{\idxitem\idxitemalt}^{(\idxcomparison)}$ and therefore $\numwins_{\idxitem\idxitemalt} \sample \binomial(\numcomparisons, \meangt_{\idxitem\idxitemalt})$. Denote $\meanemp_{\idxitem\idxitemalt}$ as the fraction of times that item $\idxitem$ beats item $\idxitemalt$. That is,
\begin{align}
    \meanemp_{\idxitem\idxitemalt} \defn \frac{1}{\numcomparisons}\numwins_{\idxitem\idxitemalt} = \frac{1}{\numcomparisons}\sum_{\idxcomparison=1}^\numcomparisons \resultcomparison_{\idxitem\idxitemalt}^{(\idxcomparison)}.\label{eq:notation_observed_fraction_of_wins}
\end{align}
We have $\meanemp_{\idxitem\idxitemalt}\sample \frac{1}{\numcomparisons}\binomial(\numcomparisons, \meangt_{\idxitem\idxitemalt})$, independent across all $\idxitem<\idxitemalt$.

Finally, we use $\const, \const', \const_1, \const_2$, etc. to denote finite constants whose values may change from line to line. We write $\func(n) \lessorder \funcalt(n)$  if there exists a constant $\const$ such that $\func(n) \le \const\cdot \funcalt(n)$ for all $n \ge 1$. The notation $\func(n) \greaterorder \funcalt(n)$ is defined analogously.

    \item \textbf{Notion of conditioning}
    
    Let $\event$ be any event. The conditional bias of any estimator $\paramest$ conditioned on the event $\event$ is defined as:
    \begin{align*}
        \bias(\paramest \given \event) \defn \sup_{\paramgt \in \rangeboxgt} \norm{\Expect[\paramest \given \event] - \paramgt}_\infty.
    \end{align*}

    We use ``\whpone'' to denote that an event $\event$ happens with probability at least
    \begin{align*}
        \Prob(\event) > 1 - \frac{\const}{\numitems\numcomparisons},
    \end{align*}
    for all $\numitems\ge \constitems$ and $\numcomparisons\ge \constcomparisons$, where $\constitems, \constcomparisons$ and $\const$ are positive constants.

    Similarly, we use ``\whponecondition{\event}'' to denote that conditioned on some event $\event$, some other event $\event'$ happens with probability at least
    \begin{align*}
            \Prob(\event' \given \event) \ge 1-\frac{\const}{\numitems\numcomparisons},
    \end{align*}
    for all $\numitems\ge \constitems$ and $\numcomparisons\ge \constcomparisons$, where $\constitems, \constcomparisons$ and $\const$ are positive constants.

    %%%
    \item \textbf{The negative log-likelihood function and its derivative}

    Recall that $\negloglikelihood$ denotes the negative log-likelihood function. Under the BTL model, we have
    \begin{align}
        \negloglikelihood(\param) \defn \negloglikelihood(\{\numwins_{\idxitem\idxitemalt}\}; \param) & = - \sum_{1 \leq \idxitem < \idxitemalt \leq \numitems} \left[\numwins_{\idxitem\idxitemalt} \log\left(\frac{1}{1 + e^{-(\param_\idxitem-\param_\idxitemalt)}}\right) +  \numwins_{\idxitemalt\idxitem}\log\left(\frac{1}{1 + e^{-(\param_\idxitemalt-\param_\idxitem)}}\right)\right] \nonumber\\
        & = - \numcomparisons\sum_{1 \leq \idxitem < \idxitemalt \leq \numitems} \left[\meanemp_{\idxitem\idxitemalt} \log\left(\frac{1}{1 + e^{-(\param_\idxitem-\param_\idxitemalt)}}\right) +  \meanemp_{\idxitemalt\idxitem}\log\left(\frac{1}{1 + e^{-(\param_\idxitemalt-\param_\idxitem)}}\right)\right] \nonumber\\
        & = \numcomparisons \sum_{1 \le \idxitem < \idxitemalt \le \numitems} \left[\log(e^{\param_\idxitem} + e^{\param_\idxitemalt}) -\meanemp_{\idxitem\idxitemalt} \param_\idxitem -  \meanemp_{\idxitemalt\idxitem}\param_\idxitemalt\right].\label{eq:loglikelihood_expression}
    \end{align}
    Since $\{\meanemp_{\idxitem\idxitemalt}\}$ is simply a normalized version of $\{\numwins_{\idxitem\idxitemalt}\}$, we equivalently denote the negative log-likelihood function as $\negloglikelihood(\{\meanemp_{\idxitem\idxitemalt}\}; \param)$.
        
    From the expression of $\negloglikelihood$ in~\eqref{eq:loglikelihood_expression}, we compute the gradient $\lfrac{\partial \negloglikelihood}{\partial \param_\idxitemthree}$ for every $\idxitemthree\in [\numitems]$ as
    \begin{align}
        \lfrac{\partial \negloglikelihood}{\partial \param_\idxitemthree} & = \numcomparisons \sum_{\idxitem\ne \idxitemthree} \left( \frac{1}{1 + e^{-(\param_\idxitemthree-\param_\idxitem )}}-\meanemp_{\idxitemthree\idxitem}\right).\label{eq:log_likelihood_gradient_raw}
    \end{align}

    Finally, the following lemma from~\cite{hunter2004mm} shows the strict convexity of the negative log-likelihood function $\negloglikelihood$.

        \begin{lemma}[Lemma 2(a) from~\cite{hunter2004mm}]\label{lem:log_likelihood_convex}
            The negative log-likelihood function $\negloglikelihood(\param)$ is strictly convex in $\param\in \reals^\numitems$.
        \end{lemma}
    \item \textbf{The sigmoid function and its derivatives}
    
    Denote the function $\funcsigmoid: (-\infty, \infty) \rightarrow (0, 1)$ as the sigmoid function $\funcsigmoid(\sigvar) = \frac{1}{1 + e^{-\sigvar}}$. It is straightforward to verify that the function $\funcsigmoid$ has the following two properties.
    \begin{itemizex}
        \item The first derivative $\funcsigmoid'$ is positive on $(-\infty, \infty)$. Moreover, on any bounded interval, the first derivative $\funcsigmoid'$ is bounded above and below. That is, for any constants $\const_1 < \const_2$, there exist constants $\const_3, \const_4 > 0$ such that
        \begin{subequations}
        \begin{align}
            0<\const_3< \funcsigmoid'(\sigvar) < \const_4, \qquad \text{for all } \sigvar\in (\const_1, \const_2).\label{eq:property_first_derivative_bounded}
        \end{align}
        \item The second derivative $\func''$ is bounded on any bounded interval. That is, for any constants $\const_1 < \const_2$, there exists a constant $\const_5$ such that
        \begin{align}
            \abs{\funcsigmoid''(\sigvar)} < \const_5, \qquad \text{for all } \sigvar\in (\const_1, \const_2).\label{eq:property_second_derivative_bounded}
        \end{align}
        \end{subequations}
    \end{itemizex}

    %%%%%%%%%%%%
    \item \textbf{Existence and uniqueness of \mle}

    Recall that the \mle~\eqref{eq:intro_mle}, the unconstrained \mle~\eqref{eq:intro_mle_unconstrain}, and the \expandedmle~\eqref{eq:intro_expmle} are respectively defined as:
    \begin{align}
        \paramestboundgt(\{\meanemp_{\idxitem\idxitemalt}\}) & = \argmin_{\param\in \rangeboxgt} \negloglikelihood(\{\meanemp_{\idxitem\idxitemalt}\};\param), \label{eq:mle_standard}\\
        \paramestunconstrain(\{\meanemp_{\idxitem\idxitemalt}\}) & = \argmin_{\param\in \rangeboxinf} \negloglikelihood(\{\meanemp_{\idxitem\idxitemalt}\}; \param),\label{eq:unconstrained_mle}\\
        \paramestbound{\boundused}(\{\meanemp_{\idxitem\idxitemalt}\}) & = \argmin_{\param\in \rangeboxused} \negloglikelihood(\{\meanemp_{\idxitem\idxitemalt}\}; \param).\label{eq:mle_constrained}
\end{align}
The following lemma shows the existence and uniqueness of the \expandedmle $\paramestbound{\boundused}$~\eqref{eq:mle_constrained} for any constant $\boundused > 0$, which incorporates the \standardmle $\paramestbound{\boundgt}$ by setting $\boundused = \boundgt$.
        
    \begin{lemma}\label{lem:unique_solution_constrained}
        For any finite constant $\boundused > 0$, there always exists a unique solution $\paramestbound{\boundused}$ to the \expandedmle~\eqref{eq:mle_constrained}.
    \end{lemma}
    See \App~\ref{sec:proof_lem_unique_solution_constrained} for the proof of Lemma~\ref{lem:unique_solution_constrained}.

    For the unconstrained \mle, due to the removal of the box constraint in~\eqref{eq:unconstrained_mle}, a finite solution $\paramestunconstrain$ may not exist. However, the following lemma shows that a unique finite solution exists with high probability.

    \begin{lemma}\label{lem:finite_solution}
        There exists a unique finite solution $\paramestunconstrain$ to the unconstrained \mle~\eqref{eq:unconstrained_mle} \whpone.
    \end{lemma}
    See \App~\ref{sec:proof_lem_finite_solution} for the proof of Lemma~\ref{lem:finite_solution}.
    
    In the subsequent proofs of Theorem~\ref{thm:mle_ub_lb}\ref{part:ub} and Theorem~\ref{thm:mle_ub_lb}\ref{part:lb}, we heavily use the unconstrained \mle as an intermediate quantity to analyze the \mle and the \expandedmle.

\end{enumeratex}

%%%%%%%%%%%%%%%%%%%%%%%%
\subsection{Proof of Theorem~\ref{thm:mle_ub_lb}\ref{part:ub}}\label{sec:proof_ub}

In this \app, we present the proof of Theorem~\ref{thm:mle_ub_lb}\ref{part:ub}. To describe the main steps involved, we first present a proof sketch of a simple case of $\numitems=2$ items (\App~\ref{sec:toy_ub}), followed by the complete proof of the general case (\App~\ref{sec:actual_proof_ub}). The reader may pass to the complete proof in \App~\ref{sec:actual_proof_ub} without loss of continuity.

%%%%%%
\subsubsection{Simple case: 2 items}\label{sec:toy_ub}
We first present an informal proof sketch for a simple case where there are $\numitems=2$ items. The proof for the general case in \App~\ref{sec:actual_proof_ub} follows the same outline. In the case of $\numitems=2$ items, due to the centering constraint on the true parameter vector $\paramgt$, we have $\paramgt_2 = -\paramgt_1$. Similarly, we have $\paramest_2 = -\paramest_1$ for any estimator that satisfies the centering constraint (in particular, for the \expandedmle $\paramestboundused$ and the unconstrained \mle $\paramestunconstrain$). Therefore, it suffices to focus only on item $1$. Since there are only two items, for ease of notation, we denote $\meanemp = \meanemp_{12}$ and $\meangt = \meangt_{12}$. We now present the main steps of the proof sketch.~~\\

\textbf{Proof sketch of the $2$-item case (informal):}

In the proof sketch, we fix any $\paramgt\in \rangeboxgt$, and any finite constants $\boundused$ and $\boundgt$ such that $\boundused >\boundgt > 0$.

\begin{enumeratex}[label=\bfseries Step \arabic*:, ref=\arabic*]
    %%%%%%
    \item \textbf{Establish concentration of $\mean$} 
    
    By Hoeffding's inequality, we have
    \begin{align}\label{eq:toy_bound}
        \abs{\mean -\meangt} \lessorder \sqrt{\frac{\log\numcomparisons}{\numcomparisons}}, \qquad \text{w.h.p.}
    \end{align}
    
    Since $\abs{\paramgt} \le \boundgt$, we have that $\meangt$ is bounded away from $0$ and $1$ by a constant. Hence, for sufficiently large $\numcomparisons$, there exist constants $\constubsketchlower, \constubsketchupper$ where $0 < \constubsketchlower < \constubsketchupper < 1$, such that 
    \begin{align}
        \mean, \meangt\in (\constubsketchlower, \constubsketchupper).\label{eq:toy_ub_mean_bounded}
    \end{align}

    %%%%%%
    \item \textbf{Write the first-order optimality condition for $\paramestunconstrain$}\label{step:toy_mle}
    
    The unconstrained \mle $\paramestunconstrain$ minimizes the negative log-likelihood $\negloglikelihood$. If a finite unconstrained \mle $\paramestunconstrain$ exists\footnote{
        For the proof sketch, we ignore the high-probability nature of Lemma~\ref{lem:finite_solution}, and assume that a finite $\paramestunconstrain$ always exists. It is made precise in the complete proof in \App~\ref{sec:actual_proof_ub}.
    }, we have $\del_{\param = \paramestunconstrain} \negloglikelihood(\param) = 0$. Setting $\idxitemthree=1$ in the gradient expression~\eqref{eq:log_likelihood_gradient_raw} and plugging in $\paramestunconstrain$,  we have
    \begin{align}
        \left.\lfrac{\partial \negloglikelihood}{\partial \param_1}\right|_{\param = \paramestunconstrain} & = \numcomparisons\left(\frac{1}{1 + e^{-(\paramestunconstrain_1 - \paramestunconstrain_2)}} - \mean_{12}\right) \nonumber\\
        & = \numcomparisons \left(\frac{1}{1 + e^{-2\paramestunconstrain_1}} - \mean\right).\label{eq:toy_ub_gradient}
    \end{align}
    Setting the derivative~\eqref{eq:toy_ub_gradient} to $0$, we have
    \begin{align}
        \paramestunconstrain_1 & = -\frac{1}{2}\log\left(\frac{1}{\meanemp} - 1\right).\label{eq:toy_mle}
    \end{align}
    By the definition of $\{\meangt_{\idxitem\idxitemalt}\}$ in~\eqref{eq:btl}, we have $\meangt = \frac{1}{1 + e^{-(\paramgt_1 - \paramgt_2)}} = \frac{1}{1+ e^{-2\paramgt_1}}$, which can be written as
    \begin{align}
        \paramgt_1 = -\frac{1}{2}\log\left(\frac{1}{\meangt}-1\right).\label{eq:toy_true_log}
    \end{align}
    Define a function $\funclog: [0, 1] \rightarrow \reals \union \{\pm \infty\}$ as
    \begin{align}
        \funclog(\varmean) = -\frac{1}{2}\log\left(\frac{1}{\varmean} - 1\right).\label{eq:toy_ub_def_func}
    \end{align}
    Subtracting~\eqref{eq:toy_true_log} from~\eqref{eq:toy_mle} and using the definition of $\funclog$ from~\eqref{eq:toy_ub_def_func}, we have
    \begin{align}
        \paramestunconstrain_1 - \paramgt_1 = \funclog({\mean}) - \funclog({\meangt}).\label{eq:toy_relation}
    \end{align}
    
    %%%%%%
    \item  \textbf{Bound the difference between $\paramestunconstrain$ and $\paramgt$, by the first-order mean value theorem}
    
    It can be verified that $\funclog$ has positive first-order derivative on $(0, 1)$. Moreover, there exists some constant $\const_1$ such that $0 < \funclog'(\varmean) < \const_1$ for all $\varmean \in (\constubsketchlower, \constubsketchupper)$. Applying the first-order mean value theorem on~\eqref{eq:toy_relation}, we have the deterministic relation
    \begin{align}
        \paramestunconstrain_1 - \paramgt_1 & = \funclog'(\meanvalue)\cdot (\mean - \meangt), \label{eq:toy_mvt_order_one_raw}
    \end{align}
    where $\meanvalue$ is a random variable that depends on $\mean$ and $\meangt$, and takes values between $\mean$ and $\meangt$. By~\eqref{eq:toy_ub_mean_bounded}, we have $\meanvalue\in (\constubsketchlower, \constubsketchupper)$. From~\eqref{eq:toy_mvt_order_one_raw} we have
    
    \begin{align}
        \abs{\paramestunconstrain_1 - \paramgt_1} & \le \const_1 \abs{\mean - \meangt}.\label{eq:toy_mvt_order_one}
    \end{align}
     Combining~\eqref{eq:toy_mvt_order_one} with~\eqref{eq:toy_bound}, we have 
    \begin{align}
        \abs{\paramestunconstrain_1 - \paramgt_1} & \lessorder \sqrt{\frac{\log\numcomparisons}{\numcomparisons}}, \qquad \text{w.h.p.}\label{eq:toy_bound_param_whp}
    \end{align}
    
    %%%%%%
    \item \textbf{Bound the \emph{expected} difference between $\paramestunconstrain$ and $\paramgt$, by the second-order mean value theorem}
    
    By the second-order mean value theorem on~\eqref{eq:toy_relation}, we have the deterministic relation
        \begin{align}
            \paramestunconstrain_1 - \paramgt_1 = \funclog({\mean}) - \funclog({\meangt}) = \funclog'(\meangt) \cdot(\mean - \meangt) + \funclog''(\meanvaluealt)\cdot (\mean - \meangt)^2,\label{eq:toy_mvt_order_two}
        \end{align}
        where $\meanvaluealt$ is a random variable that depends on $\mean$ and $\meangt$, and takes values between $\mean$ and $\meangt$. By~\eqref{eq:toy_ub_mean_bounded}, we have $\meanvaluealt\in (\constubsketchlower, \constubsketchupper)$.
        
        It can be verified that $\funclog$ has bounded second-order derivative. That is, $\abs{\funclog''(\varmean)} < \const_2$ for all $\varmean\in (\constubsketchlower, \constubsketchupper)$. Taking an expectation over~\eqref{eq:toy_mvt_order_two}, we have
        \begin{align}
            \Expect[\paramestunconstrain_1] - \paramgt_1 & = \funclog'(\meangt) \cdot (\Expect[\meanemp] - \meangt) +\Expect[ \funclog''(\meanvaluealt)\cdot (\meanemp - \meangt)^2 ]\label{eq:toy_mvt_order_two_expect}\\
            & \stackrel{\stepone}{\le} \const_2 \Expect[(\meanemp - \meangt)^2] \nonumber\\
            & \stackrel{\steptwo}{\lessorder} \frac{\log\numcomparisons}{\numcomparisons},\label{eq:toy_final}
        \end{align}
        where \stepone is true because $\Expect[\mean] = \meangt$ combined with the fact that $\abs{\funclog''} < \const_2$ on $(\constubsketchlower, \constubsketchupper)$, and \steptwo is true\footnote{
            For the proof sketch, we ignore the high-probability nature of~\eqref{eq:toy_bound} and treat it as a deterministic relation. It is made precise in the complete proof in \App~\ref{sec:actual_proof_ub}.
        } by~\eqref{eq:toy_bound}.
        
        \item \textbf{Connect $\paramestunconstrain$ back to $\paramestboundused$}
        
        From~\eqref{eq:toy_bound_param_whp}, we have $\abs{\paramestunconstrain_1 -\paramgt_1} \le \boundused-\boundgt$ \whp for sufficiently large $\numcomparisons$. Hence,
    \begin{align*}
        \abs{\paramestunconstrain_1} \le \abs{\paramgt_1} + \abs{\paramestunconstrain_1 - \paramgt_1} \le \boundgt + (\boundused-\boundgt) = \boundused, \qquad \text{\whp}
    \end{align*}
    
    Moreover, we have $\abs*{\paramestunconstrain_2} = \abs*{\paramestunconstrain_1} \le \boundused$. Therefore, with high probability, the unconstrained \mle $\paramestunconstrain$ does not violate the box constraint at $\boundused$, and therefore $\paramestunconstrain$ is identical to the \expandedmle $\paramestboundused$. Hence, the bound~\eqref{eq:toy_final} holds\footnote{
        For the proof sketch, we ignore the high-probability nature of the fact that $\paramestunconstrain=\paramestboundused$, and treat it as a deterministic relation. It is made precise in the complete proof in \App~\ref{sec:actual_proof_ub}.
    } for the \expandedmle, completing the proof sketch.
\end{enumeratex}

%%%%%%%%%%%%
\subsubsection{Complete Proof}\label{sec:actual_proof_ub}

In this \app, we present the proof of Theorem~\ref{thm:mle_ub_lb}\ref{part:ub}, by formally extending the $5$ steps outlined for the simple case in \App~\ref{sec:toy_ub}. In the general case, one notable challenge is that one can no longer write a closed-form solution of the \mle as we did in~\eqref{eq:toy_mle} of  Step~\ref{step:toy_mle}. The first-order optimality condition now becomes a system of equations that describe an implicit relation between $\param$ and $\mean$, requiring more involved analysis.  

In the proof, we fix any $\paramgt\in \rangeboxgt$, and fix any finite constants $\boundused$ and $\boundgt$ such that $\boundused>\boundgt > 0$.

\begin{enumeratex}[label={\bfseries Step \arabic*:}, ref=\arabic*]
    %%%%%%
    \item \label{step:concentration} \textbf{Establish concentration of $\{\meanemp_{\idxitem
    \idxitemalt}\}$}
    
    We first use standard concentration inequalities to establish the following lemma, to be used in the subsequent steps of the proof.
    \begin{lemma}\label{lem:concentration}
    There exists a constant $\const > 0$, such that
    \begin{align*}
        \abs*{\sum_{\idxitem\ne \idxitemthree}\meanemp_{\idxitemthree\idxitem} - \sum_{\idxitem\ne \idxitemthree} \meangt_{\idxitemthree\idxitem}} \le \const\sqrt{\frac{\numitems(\log\numitems+\log\numcomparisons)}{\numcomparisons}},
    \end{align*}
    simultaneously for all $\idxitemthree\in [\numitems]$ \whpone.
    \end{lemma}
    See \App~\ref{sec:proof_lem_concentration} for the proof of Lemma~\ref{lem:concentration}.
        
    Recall that Lemma~\ref{lem:finite_solution} states that a finite unconstrained \mle $\paramestunconstrain$ exists \whpone. We denote $\eventlemmaub$ as the event that Lemma~\ref{lem:finite_solution} and Lemma~\ref{lem:concentration} both hold. For the rest of the proof, we condition on $\eventlemmaub$. Since both Lemma~\ref{lem:finite_solution} and Lemma~\ref{lem:concentration} hold \whpone, taking a union bound, we have that $\eventlemmaub$ holds \whpone. That is,
    \begin{align}
        \Prob(\eventlemmaub) \ge 1-\frac{\const}{\numitems\numcomparisons}, \qquad \text{for some constant } \const > 0.\label{eq:ub_lemma_whp}
    \end{align}
    
    %%%%%%
    \item \label{step:first_order_optimality}
    \textbf{Write the first-order optimality condition for the unconstrained \mle $\paramestunconstrain$}
    
    Recall from Lemma~\ref{lem:log_likelihood_convex} that the negative log-likelihood function $\negloglikelihood$ is convex in $\param$. In this step, we first justify that the whenever a finite unconstrained \mle  $\paramestunconstrain$ exists, it satisfies the first-order optimality condition $\del_{\param = \paramestunconstrain} \negloglikelihood(\param) = 0$. (Note that for any optimization problem with constraints, it is in general not true that the derivative of the convex objective equals $0$ at the optimal solution.) Then we derive a specific form of the first-order optimality condition, to be used in subsequent steps of the proof.
    
    Given that we have conditioned on $\eventlemmaub$ (and therefore on Lemma~\ref{lem:finite_solution}), a finite solution $\paramestunconstrain$ to the unconstrained \mle exists. To show that $\paramestunconstrain$ satisfies the first-order optimality condition, we show that $\paramestunconstrain$ is also a solution to the following \mle without any constraint at all (that is, we remove the centering constraint too):
    \begin{align}\label{eq:totally_unconstrained_mle}
        \argmin_{\param\in \reals^\numitems} \negloglikelihood(\param).
    \end{align}
    
    If the unconstrained \mle $\paramestunconstrain$ is a solution to~\eqref{eq:totally_unconstrained_mle}, then it satisfies the first-order condition $\del_\param \negloglikelihood(\paramestunconstrain) = 0$. Now we prove that $\paramestunconstrain$ is a solution to~\eqref{eq:totally_unconstrained_mle}. Note that the solutions to~\eqref{eq:totally_unconstrained_mle} are shift-invariant. That is, if $\param$ is a solution to~\eqref{eq:totally_unconstrained_mle}, then $\param + \const \onevector$ is also a solution, where $\onevector$ is the $\numitems$-dimensional all-one vector, and $\const$ is any constant. Now suppose by contradiction that $\paramestunconstrain$ is not a solution to~\eqref{eq:totally_unconstrained_mle}. Then there exists some finite $\param\in \reals^\numitems$ such that $\negloglikelihood(\param) < \negloglikelihood(\paramestunconstrain)$. Now consider $\param' \defn \param - (\frac{1}{\numitems}\sum_{\idxitem=1}^\numitems\param_\idxitem) \onevector$. We have $\param' \in \rangeboxinf$ because it satisfies the centering constraint, and we have $\negloglikelihood(\param') = \negloglikelihood(\param) < \negloglikelihood(\paramestunconstrain)$ because the solutions to~\eqref{eq:totally_unconstrained_mle} are shift-invariant. The construction of $\param'$ thus contradicts the assumption that $\paramestunconstrain$ is optimal for the unconstrained \mle. Hence, $\paramestunconstrain$ is a solution to~\eqref{eq:totally_unconstrained_mle}, and $\paramestunconstrain$ satisfies the first-order optimality condition.
    
    Now we derive a specific form of the first-order optimality condition. Plugging $\paramestunconstrain$ into the gradient expression~\eqref{eq:log_likelihood_gradient_raw} and setting the gradient to $0$, we have the deterministic equality
        \begin{align}\label{eq:first_order_optimality_raw}
             \sum_{\idxitem\ne \idxitemthree} \frac{1}{1 + e^{-(\paramestunconstrain_\idxitemthree-\paramestunconstrain_\idxitem)}}=\sum_{\idxitem\ne \idxitemthree} \meanemp_{\idxitemthree\idxitem}, \qquad \text{for every } \idxitemthree \in [\numitems].
        \end{align}
    In words, the first-order optimality condition~\eqref{eq:first_order_optimality_raw} means that for any item $\idxitemthree\in [\numitems]$, the probability that item $\idxitemthree$ wins (among all comparisons in which item $\idxitemthree$ is involved) as predicted by the unconstrained \mle $\paramestunconstrain$ equals the fraction of wins by item $\idxitemthree$ from the observed comparisons. We now subtract~\eqref{eq:btl} from both sides of ~\eqref{eq:first_order_optimality_raw}:
    \begin{align}
        \sum_{\idxitem\ne \idxitemthree}\left( \frac{1}{1 + e^{-(\paramestunconstrain_\idxitemthree-\paramestunconstrain_\idxitem)}}-\frac{1}{1 + e^{-(\paramgt_\idxitemthree-\paramgt_\idxitem)}}\right)=\sum_{\idxitem\ne \idxitemthree} (\meanemp_{\idxitemthree\idxitem}- \meangt_{\idxitemthree\idxitem})\nonumber\\
        \sum_{\idxitem=1}^\numitems\left( \frac{1}{1 + e^{-(\paramestunconstrain_\idxitemthree-\paramestunconstrain_\idxitem)}}-\frac{1}{1 + e^{-(\paramgt_\idxitemthree-\paramgt_\idxitem)}}\right)=\sum_{\idxitem\ne \idxitemthree} (\meanemp_{\idxitemthree\idxitem}- \meangt_{\idxitemthree\idxitem}).\label{eq:first_order_optimality}
    \end{align}
    For ease of notation, we denote the random vector $\diforacle \defn \paramestunconstrain - \paramgt$. Equivalently, we have $\paramestunconstrain = \paramgt+\diforacle$. Using the definition of $\diforacle$, we rewrite~\eqref{eq:first_order_optimality} as:
    \begin{align}
        \sum_{\idxitem=1}^\numitems \left(\frac{1}{1 + e^{-( \paramgt_\idxitemthree-\paramgt_\idxitem  +\diforacle_\idxitemthree-\diforacle_\idxitem)}} - \frac{1}{1 + e^{-(\paramgt_\idxitemthree-\paramgt_\idxitem)}}\right) = \sum_{\idxitem\ne \idxitemthree}(\meanemp_{\idxitemthree\idxitem}-\meangt_{\idxitemthree\idxitem}).\label{eq:first_order_optimality_recall_raw}
    \end{align}
    Using the definition of the sigmoid function $\func(x) = \frac{1}{1 + e^{-x}}$, we rewrite~\eqref{eq:first_order_optimality_recall_raw} as:
    \begin{align}
        \sum_{\idxitem=1}^\numitems \left[\funcsigmoid( \paramgt_\idxitemthree-\paramgt_\idxitem  +\diforacle_\idxitemthree-\diforacle_\idxitem) - \funcsigmoid(\paramgt_\idxitemthree-\paramgt_\idxitem)\right] = \sum_{\idxitem\ne \idxitemthree}(\meanemp_{\idxitemthree\idxitem}-\meangt_{\idxitemthree\idxitem}).\label{eq:first_order_optimality_recall}
    \end{align}
    In the rest of the proof, we primarily work with the first-order optimality condition in the form of~\eqref{eq:first_order_optimality_recall}.

    %%%%%%
    \item \textbf{Bound the difference between the unconstrained \mle $\paramestunconstrain$ and the true parameter vector $\paramgt$}\label{step:ub_bound_uniform}
    
    The first-order optimality condition~\eqref{eq:first_order_optimality_recall} can be thought of as a system of equations that describes some implicit relation between the unconstrained \mle $\paramestunconstrain$ and the observations $\{\mean_{\idxitemthree\idxitem}\}$. Intuitively, the concentration of $\{\mean_{\idxitemthree\idxitem}\}$ on the RHS of~\eqref{eq:first_order_optimality_recall} (by Lemma~\ref{lem:concentration}) should imply the concentration of the unconstrained \mle $\paramestunconstrain$ on the LHS. The following lemma formalizes this intuition about the concentration of $\paramestunconstrain$.

    \begin{lemma}\label{lem:bound_diff_each}
        Conditioned on $\eventlemmaub$, we have the deterministic relation
        \begin{align*}
            \abs{\diforacle_\idxitemthree} = \abs{\paramestunconstrain_\idxitemthree - \paramgt_\idxitemthree} \lessorder \sqrt{\frac{\log\numitems+\log\numcomparisons}{\numitems\numcomparisons}}, \qquad \text{for every }  \idxitemthree \in [\numitems],
        \end{align*}
        for all $\numitems\ge \constitems$ and $\numcomparisons\ge \constcomparisons$, where $\constitems$ and $\constcomparisons$ are constants.
        \end{lemma}
    See \App~\ref{sec:proof_lem_bound_diff_each} for the proof of Lemma~\ref{lem:bound_diff_each}.
        
    This lemma provides a deterministic bound on the difference between $\paramestunconstrain$ and $\paramgt$. Now we move to analyze the difference between $\paramestunconstrain$ and $\paramgt$ in expectation.

    \item \textbf{Bound the \emph{expected} difference between the unconstrained \mle $\paramestunconstrain$ and the true parameter vector $\paramgt$, using the second-order mean value theorem}\label{step:ub_bound_expectation}
    
    In Step~\ref{step:concentration} we bound the difference between $\{\meanemp_{\idxitemthree\idxitem}\}$ and $\{\meangt_{\idxitemthree\idxitem}\}$ with high-probability. However, if we consider the difference in expectation, we have $\Expect[\meanemp_{\idxitemthree\idxitem}] = \meangt_{\idxitemthree\idxitem}$. The \emph{expected} difference between $\{\meanemp_{\idxitemthree\idxitem}\}$ and $\{\meangt_{\idxitemthree\idxitem}\}$ is $0$, significantly smaller than the high-probability bound in Step~\ref{step:concentration}. Intuitively, we may also expect that the \emph{expected} difference between $\paramestunconstrain$ and $\paramgt$ is smaller than the deterministic bound in Lemma~\ref{lem:bound_diff_each}. In this step, we formalize this intuition.

    By the second-order mean value theorem on the LHS of the first-order optimality condition~\eqref{eq:first_order_optimality_recall}, we have the deterministic relation that for every $\idxitemthree\in[\numitems]$,
    \begin{align}
        \sum_{\idxitem=1}^\numitems \left[ \funcsigmoid'(\paramgt_\idxitemthree-\paramgt_\idxitem) \cdot ( \diforacle_\idxitemthree-\diforacle_\idxitem)+ \frac{1}{2}\funcsigmoid''(\meanvalue_{\idxitemthree\idxitem})\cdot (\diforacle_\idxitemthree-\diforacle_\idxitem)^2 \right] & = \sum_{\idxitem\ne \idxitemthree}(\meanemp_{\idxitemthree\idxitem}-\meangt_{\idxitemthree\idxitem})\nonumber\\
        \sum_{\idxitem=1}^\numitems \funcsigmoid'(\paramgt_\idxitemthree-\paramgt_\idxitem) \cdot ( \diforacle_\idxitemthree-\diforacle_\idxitem) & = \sum_{\idxitem\ne \idxitemthree}(\meanemp_{\idxitemthree\idxitem}-\meangt_{\idxitemthree\idxitem})-\frac{1}{2}\sum_{\idxitem=1}^\numitems \funcsigmoid''(\meanvalue_{\idxitemthree\idxitem})\cdot (\diforacle_\idxitemthree-\diforacle_\idxitem)^2,\label{eq:ub_mvt_order_two}
    \end{align}
    where each $\meanvalue_{\idxitemthree\idxitem}$ is a random variable that takes values between $\paramgt_\idxitemthree - \paramgt_\idxitem$ and $\paramgt_\idxitemthree-\paramgt_\idxitem+(\diforacle_\idxitemthree - \diforacle_\idxitem)$. Taking an expectation over~\eqref{eq:ub_mvt_order_two} conditional on $\eventlemmaub$, we have that for every $\idxitemthree\in [\numitems]$:
    \begin{align}
        \sum_{\idxitem=1}^\numitems \funcsigmoid'(\paramgt_\idxitemthree-\paramgt_\idxitem) \cdot \Expect\left[ \diforacle_\idxitemthree-\diforacle_\idxitem \given \eventlemmaub \right] & = 
        \sum_{\idxitem\ne \idxitemthree}(\Expect[\meanemp_{\idxitemthree\idxitem}\given \eventlemmaub]-\meangt_{\idxitemthree\idxitem}) - %
        \frac{1}{2}\sum_{\idxitem=1}^\numitems\Expect[\funcsigmoid''(\meanvalue_{\idxitemthree\idxitem})(\diforacle_\idxitemthree-\diforacle_\idxitem)^2
        \given \eventlemmaub].\label{eq:ub_mvt_order_two_expectation_raw}
    \end{align}
    Denote the vector $\difexpect \defn \Expect[\diforacle \given \eventlemmaub] = \Expect[\paramestunconstrain \given \eventlemmaub] - \paramgt$. Plugging this definition of $\difexpect$ into~\eqref{eq:ub_mvt_order_two_expectation_raw} yields
    \begin{align}
        \sum_{\idxitem=1}^\numitems \funcsigmoid'(\paramgt_\idxitemthree-\paramgt_\idxitem) \cdot (\difexpect_\idxitemthree-\difexpect_\idxitem) & = 
        \sum_{\idxitem\ne \idxitemthree}(\Expect[\meanemp_{\idxitemthree\idxitem}\given \eventlemmaub]-\meangt_{\idxitemthree\idxitem}) - %
        \frac{1}{2}\sum_{\idxitem=1}^\numitems\Expect[\funcsigmoid''(\meanvalue_{\idxitemthree\idxitem})(\diforacle_\idxitemthree-\diforacle_\idxitem)^2
        \given \eventlemmaub].\label{eq:ub_mvt_order_two_expectation}
    \end{align}
    We first bound the RHS of~\eqref{eq:ub_mvt_order_two_expectation}, and then derive a bound regarding $\difexpect_\idxitem$ on the LHS accordingly.

    To bound the RHS of~\eqref{eq:ub_mvt_order_two_expectation}, we first consider the term $\Expect[\meanemp_{\idxitemthree\idxitem}\given \eventlemmaub]-\meangt_{\idxitemthree\idxitem}$. In what follows, we state a lemma that is slightly more general than what is needed here. The more general version is used in the subsequent proof of Theorem~\ref{thm:mle_ub_lb}\ref{part:lb}. To state the lemma, recall the definition that an event $\event'$ happens \whponecondition{\event}, if the conditional probability $\Prob(\event' \given \event) \ge 1-\frac{\const}{\numitems\numcomparisons}$,
    for some constant $\const > 0$.

    \begin{lemma}\label{lem:ub_conditional_expectation}
        Let $\event$ be any event, and let $\event'$ be any event that happens \whponecondition{\event}. Then for any $\idxitemthree \ne \idxitem$, we have
        \begin{align}
            \abs*{\Expect[\meanemp_{\idxitemthree\idxitem} \given \event', \event] - \Expect[\meanemp_{\idxitemthree\idxitem} \given \event]} \lessorder \frac{1}{\numitems\numcomparisons}.\label{eq:ub_empirical_prob_expectation_bound}
        \end{align}
    \end{lemma}
    See \App~\ref{sec:proof_lem_ub_conditional_expectation} for the proof of Lemma~\ref{lem:ub_conditional_expectation}.
    
    To apply Lemma~\ref{lem:ub_conditional_expectation}, we set $\event$ to be the (trivial) event of the entire probability space, and set $\event'$ to be $\eventlemmaub$ in~\eqref{eq:ub_empirical_prob_expectation_bound}. We have
     \begin{align}
        \abs*{\Expect[\meanemp_{\idxitemthree\idxitem} \given \eventlemmaub] - \Expect[\meanemp_{\idxitemthree\idxitem}]} = \abs*{\Expect[\meanemp_{\idxitemthree\idxitem} \given \eventlemmaub] - \meangt_{\idxitemthree\idxitem}} \lessorder \frac{1}{\numitems\numcomparisons}.\label{eq:ub_empirical_prob_expectation_bound_apply_lemma}
    \end{align}

   The remaining terms in~\eqref{eq:ub_mvt_order_two_expectation} are handled in the following lemma. This lemma bounds the expected difference between $\paramestunconstrain$ and $\paramgt$ conditioned on $\eventlemmaub$, that is, the quantity $\abs{\difexpect_\idxitemthree} = \abs{\Expect[\paramestunconstrain_\idxitemthree\given \eventlemmaub] - \paramgt_\idxitemthree}$.
    
    \begin{lemma}\label{lem:ub_expectation_bound_each}
    Conditioned on $\eventlemmaub$, we have
    \begin{align*}
        \abs{\difexpect_\idxitemthree} \lessorder\frac{\log\numitems+\log\numcomparisons}{\numitems\numcomparisons},\qquad \text{ for every }\idxitemthree\in [\numitems],
    \end{align*}
    for all $\numitems \ge \constitems$ and all $\numcomparisons\ge \constcomparisons$, where $\constitems$ and $\constcomparisons$ are constants. Equivalently,
     \begin{align}\label{eq:ub_condition_final}
        \bias(\paramestunconstrain \given \eventlemmaub) = \norm{\Expect[\paramestunconstrain \given \eventlemmaub] - \paramgt}_\infty = \norm{\difexpect}_\infty \lessorder\frac{\log\numitems+\log\numcomparisons}{\numitems\numcomparisons},
    \end{align}
    for all $\numitems \ge \constitems$ and all $\numcomparisons\ge \constcomparisons$, where $\constitems$ and $\constcomparisons$ are constants.
    \end{lemma}
    See \App~\ref{sec:proof_lem_ub_expectation_bound_each} for the proof of Lemma~\ref{lem:ub_expectation_bound_each}.
    
    Note that~\eqref{eq:ub_condition_final} yields the desired rate on the quantity $\bias(\paramestunconstrain\given \eventlemmaub)$. It remains to show that $\bias(\paramestunconstrain\given \eventlemmaub)$ is sufficiently close to $\bias(\paramestboundused)$.
    
    \item \textbf{Show that the box constraint at $\boundused$ is vacuous for the unconstrained \mle $\paramestunconstrain$ and hence $\paramestunconstrain$ is the same as the \expandedmle $\paramestboundused$ with high probability, using the deterministic bound in Step~\ref{step:ub_bound_uniform}}\label{step:ub_recenter}
    
    To show that $\bias(\paramestunconstrain\given \eventlemmaub)$ is sufficiently close to $\bias(\paramestboundused)$, we divide the argument into two parts. First, we show that $\bias(\paramestunconstrain\given \eventlemmaub) = \bias(\paramestboundused \given \eventlemmaub)$. Second, we show that $\bias(\paramestboundused\given \eventlemmaub)$ is close to $\bias(\paramestboundused)$.

    We first show that $\bias(\paramestunconstrain\given \eventlemmaub) = \bias(\paramestboundused \given \eventlemmaub)$. Recall that $\boundused$ and $\boundgt$ are constants such that $\boundused>\boundgt$. Recall from Lemma~\ref{lem:bound_diff_each} that $\norm{\paramestunconstrain - \paramgt}_\infty \lessorder \frac{\log\numitems+\log\numcomparisons}{\numitems\numcomparisons}$ conditioned on $\eventlemmaub$. Hence, there exist constants $\constitems$ and $\constcomparisons$, such that for any $\numitems \ge \constitems$ and $\numcomparisons\ge \constcomparisons$, we have $\norm{\paramestunconstrain - \paramgt}_\infty < \boundused-\boundgt$ conditioned on $\eventlemmaub$. In this case, we have 
    \begin{align*}
        \norm{\paramestunconstrain}_\infty \le  \norm{\paramgt}_\infty +\norm{\paramestunconstrain - \paramgt}_\infty < \boundgt + (\boundused-\boundgt) = \boundused, \qquad \text{conditioned on } \eventlemmaub.
    \end{align*}
    Conditioned on $\eventlemmaub$, the unconstrained \mle $\paramestunconstrain$ obeys the box constraint $\norm{\paramestunconstrain}_\infty \le \boundused$. Therefore, $\paramestunconstrain$ is also a solution to the \expandedmle $\paramestboundused$. By the uniqueness of $\paramestboundused$ from Lemma~\ref{lem:unique_solution_constrained}, we have
    \begin{align*}
        \paramestboundused = \paramestunconstrain, \qquad \text{conditioned on } \eventlemmaub.
    \end{align*}
    Hence, we have the relation
    \begin{align}
        \bias(\paramestunconstrain \given \eventlemmaub) = \bias(\paramestboundused\given \eventlemmaub),\label{eq:ub_clipping_vacuous_bias_equal}
    \end{align}
    completing the first part of the argument.~~\\
    
   It remains to show that $\bias(\paramestboundused\given \eventlemmaub)$ is sufficiently close to $\bias(\paramestboundused)$. We have
    \begin{align}
        \bias(\paramestboundused) & = \norm{\Expect[\paramestboundused] - \paramgt}_\infty \nonumber\\
        & \stackrel{\stepone}{=} \norm{\Expect[\paramestboundused \given \eventlemmaub] \cdot \Prob(\eventlemmaub) + \Expect[\paramestboundused \given \eventlemmaubcomp]\cdot \Prob(\eventlemmaubcomp) - \paramgt}_\infty \nonumber\\
        & \stackrel{\steptwo}{\le} \norm{\Expect[\paramestboundused \given \eventlemmaub] - \paramgt}_\infty\cdot \Prob(\eventlemmaub) + \norm{\Expect[\paramestboundused \given \eventlemmaubcomp] - \paramgt}_\infty\cdot \Prob(\eventlemmaubcomp) \nonumber\\
        & = \underbrace{
                \bias(\paramestboundused\given \eventlemmaub)\cdot \Prob(\eventlemmaub)
            }_{\term_1} + \underbrace{
                \norm{\Expect[\paramestboundused \given \eventlemmaubcomp] - \paramgt}_\infty\cdot \Prob(\eventlemmaubcomp)
            }_{\term_2}.\label{eq:ub_bias_expression_final}
    \end{align}
    where step \stepone is true by the law of iterated expectation, and step \steptwo is true by the triangle inequality.
    
    Consider the two terms in~\eqref{eq:ub_bias_expression_final}. For $\term_1$, combining~\eqref{eq:ub_condition_final} and~\eqref{eq:ub_clipping_vacuous_bias_equal} yields \begin{align*}
        \bias(\paramestboundused\given \eventlemmaub) =\bias(\paramestunconstrain\given \eventlemmaub) \lessorder\frac{\log\numitems+\log\numcomparisons}{\numitems\numcomparisons}.
    \end{align*}
    Therefore,
    \begin{align}
        \term_1 \lessorder\frac{\log\numitems + \log\numcomparisons}{\numitems\numcomparisons}.\label{eq:ub_final_term_one_bound}
    \end{align}
    Now consider $\term_2$. By the box constraint $\norm{\paramestboundused}_\infty \le \boundused$, we have \begin{align}
        \norm{\Expect[\paramestboundused \given \eventlemmaubcomp] - \paramgt}_\infty \stackrel{\stepone}{\le} \norm{\Expect[\paramestboundused \given \eventlemmaubcomp]}_\infty + \norm{ \paramgt}_\infty \le \boundused+\boundgt,\label{eq:ub_final_term_two}
    \end{align}
    where step \stepone is true by the triangle inequality. Recall from~\eqref{eq:ub_lemma_whp}, the event $\eventlemmaub$ happens \whpone. Therefore,
    \begin{align}
        \Prob(\eventlemmaubcomp) \lessorder \frac{1}{\numitems\numcomparisons}.\label{eq:ub_final_term_three}
    \end{align}
    Combining~\eqref{eq:ub_final_term_two} and~\eqref{eq:ub_final_term_three} yields
    \begin{align}
        \term_2 \lessorder \frac{1}{\numitems\numcomparisons}.\label{eq:ub_final_term_two_bound}
    \end{align}
    Plugging the term $\term_1$ from~\eqref{eq:ub_final_term_one_bound} and the term $\term_2$ from~\eqref{eq:ub_final_term_two_bound} back into~\eqref{eq:ub_bias_expression_final}, we have
    \begin{align*}
        \bias(\paramestboundused) \lessorder \frac{\log\numitems + \log\numcomparisons}{\numitems\numcomparisons},
    \end{align*}
    completing the proof of Theorem~\ref{thm:mle_ub_lb}\ref{part:ub}.
\end{enumeratex} % end steps

%%%%%%%%%%%%%%%%%%%%%%%%
\subsection{Proof of Theorem~\ref{thm:mle_ub_lb}\ref{part:lb}}\label{sec:proof_lb}

Similar to the proof of Theorem~\ref{thm:mle_ub_lb}\ref{part:ub}, we first present a proof of the simple case of $\numitems=2$ items. It is important to note that although we present proofs of the $2$-item case for both Theorem~\ref{thm:mle_ub_lb}\ref{part:ub} and Theorem~\ref{thm:mle_ub_lb}\ref{part:lb}, their purposes are different. In Theorem~\ref{thm:mle_ub_lb}\ref{part:ub} presented in \App~\ref{sec:proof_ub}, the proof sketch of the $2$-item case is informal. It serves as a guideline for the general case. Then the main work involved in the general case is to generalize the arguments in the $2$-item case step-by-step. On the other hand, in Theorem~\ref{thm:mle_ub_lb}\ref{part:lb}, the proof of the $2$-item case to be presented is formal. It serves as a core sub-problem of the general case. Then the main work involved in the general case is to reduce the problem to the $2$-item case, and then the results from the $2$-item case directly.

\subsubsection{Simple case: 2 items}\label{sec:toy_lb}
As in \App~\ref{sec:toy_ub}, we first consider the simple case where there are $\numitems=2$ items. Again, due to the centering constraint, we have $\paramgt_2 = -\paramgt_1$ for the true parameter vector $\paramgt$, and we have $\paramest_2 = -\paramest_1$ for any estimator $\paramest$ that satisfies the centering constraint (in particular, for the \standardmle $\paramestboundgt$ and the unconstrained \mle $\paramestunconstrain$). Therefore, it suffices to focus only on item $1$. Since there are only two items, for ease of notation, we denote $\meanemp = \meanemp_{12}$ and $\meangt = \meangt_{12}$. 

We consider the true parameter vector $\paramgt =[\boundgt, -\boundgt]$. By the definition of $\{\meangt_{\idxitem\idxitemalt}\}$ in~\eqref{eq:btl}, we have
\begin{align*}
    \meangt = \frac{1}{1 + e^{-(\paramgt_1 - \paramgt_2)}} = \frac{1}{1 + e^{-2\boundgt}}.
\end{align*}
The following proposition now lower bounds the bias of the \standardmle $\paramestboundgt$.
\begin{proposition}\label{prop:lb_two_item_negative_bias}
Under $\paramgt = [\boundgt, -\boundgt]$, the bias of the \mle $\paramestboundgt$ is bounded as
\begin{align*}
    \bias(\paramestboundgt) = \norminf{\Expect[\paramestboundgt] -\paramgt} =  \abs{\Expect[\paramestboundgt_1] -\boundgt} \greaterorder \frac{1}{\sqrt{\numcomparisons}}.
\end{align*}
Specifically, the bias is negative, that is,
\begin{align}\label{eq:toy_lb_desired_negative_bias}
    \Expect[\paramestboundgt_1] -\boundgt \le -\frac{\const}{\sqrt{\numcomparisons}},
\end{align}
for some constant $\const > 0$.
\end{proposition}

The rest of this \app is devoted to proving~\eqref{eq:toy_lb_desired_negative_bias} in Proposition~\ref{prop:lb_two_item_negative_bias}.~~\\

For ease of notation, denote $\meanmax = \meangt = \frac{1}{1 + e^{-2\boundgt}}$, and $\meanmin = 1 - \meangt = \frac{1}{1 + e^{2\boundgt}}$. In the proof sketch of Theorem~\ref{thm:mle_ub_lb}\ref{part:ub} of the case of $\numitems=2$ items (\App~\ref{sec:toy_ub}), we derived the following expression~\eqref{eq:toy_mle} for the unconstrained \mle:
\begin{align*}
    \paramestunconstrain_1(\meanemp) = -\frac{1}{2}\log\left(\frac{1}{\meanemp}-1\right).
\end{align*}
Now consider the \standardmle $\paramestboundgt$. By straightforward analysis, one can derive the following closed-form expression for the \standardmle:
\begin{align}
    \paramestboundgt_1(\mean) = 
    \begin{cases}
        -\boundgt & \text{if } \mean\in [0, \meanmin]\\
        -\frac{1}{2}\log\left(\frac{1}{\mean}- 1\right) & \text{if } \mean \in (\meanmin, \meanmax)\\
        \boundgt & \text{if } \mean \in [\meanmax, 1].
    \end{cases}\label{eq:toy_lb_def_func_log}
\end{align}
For ease of notation, we denote a function $\funclog: [0, 1] \rightarrow [-\boundgt, \boundgt]$ as
\begin{align}
    \funclog(\varmean) = \begin{cases}
        -\boundgt & \text{if } \varmean\in [0, \meanmin]\\
        -\frac{1}{2}\log\left(\frac{1}{\varmean}- 1\right) & \text{if } \varmean\in (\meanmin, \meanmax)\\
        \boundgt & \text{if } \varmean\in [\meanmax, 1],\label{eq:toy_lb_def_func_log_free}
    \end{cases}
\end{align}
where $\funclog(\varmean) = \paramestboundgt_1(\mean = \varmean)$ for any $\varmean\in [0, 1]$. Then the \standardmle~\eqref{eq:toy_lb_def_func_log} can be equivalently written as $\funclog(\mean)$. To make the computation of the bias incurred by $\paramestboundgt$ more tractable, we also define the following auxiliary function $\funclogplus: [0, 1] \rightarrow [-\boundgt, \boundgt]$ as:
 \begin{align}
    \funclogplus(\varmean) \defn \begin{cases}
    \frac{2\boundgt}{\meanmax} \left(\varmean - \meanmax\right) + \boundgt & \text{if } \varmean\in [0, \meanmax)\\
    \boundgt & \text{if } \varmean\in [\meanmax, 1].
    \end{cases}\label{eq:toy_lb_def_func_log_plus}
\end{align}
In words, the function $\funclogplus$ is piecewise linear. On the interval $[0, \meanmax]$, it is a line passing through the points $(0, -\boundgt)$ and $(\meanmax, \boundgt)$. On the interval $[\meanmax, 1]$, its value equals the constant $\boundgt$. The following lemma now states a relation between $\funclogplus(\mean)$ and $\funclog(\mean)$ in expectation with respect to $\mean$.

\begin{lemma}\label{lem:toy_lb_surrogate_greater}
    Under $\paramgt = [\boundgt, -\boundgt]$, we have
    \begin{align}
    \Expect[\funclog(\mean)] \le  \Expect[\funclogplus(\mean)].\label{eq:toy_lb_surrogate_less_bias_raw}
    \end{align}
\end{lemma}
See \App~\ref{sec:proof_lem_toy_lb_surrogate_greater} for the proof of Lemma~\ref{lem:toy_lb_surrogate_greater}.

Now subtracting $\boundgt$ from both sides of~\eqref{eq:toy_lb_surrogate_less_bias_raw}, we have
\begin{align}
    \Expect[\paramestboundgt_1] - \paramgt_1 = \Expect[\funclog(\mean)] -\boundgt \le \Expect[\funclogplus(\mean)] - \boundgt.\label{eq:toy_lb_surrogate_less_bias}
\end{align}
The following lemma states that the bias introduced by $\funclogplus(\mean)$ satisfies the desired rate from Proposition~\ref{prop:lb_two_item_negative_bias}.
\begin{lemma}\label{lem:toy_lb_surrogate_desired_bias}
    Under $\paramgt = [\boundgt, -\boundgt]$, we have
    \begin{align}
        \Expect[\funclogplus(\mean)] - \boundgt \le -\frac{\const}{\sqrt{\numcomparisons}},\label{eq:toy_lb_surrogate_desired_bound}
    \end{align}
    for some constant $\const > 0$.
\end{lemma}
See \App~\ref{sec:proof_lem_toy_lb_surrogate_desired_bias} for the proof of Lemma~\ref{lem:toy_lb_surrogate_desired_bias}.

Combining~\eqref{eq:toy_lb_surrogate_less_bias} and~\eqref{eq:toy_lb_surrogate_desired_bound}, we have
\begin{align*}
    \Expect[\paramestboundgt_1] - \paramgt_1 \le -\frac{\const}{\sqrt{\numcomparisons}},
\end{align*}
completing the proof of~\eqref{eq:toy_lb_desired_negative_bias} in Proposition~\ref{prop:lb_two_item_negative_bias}.

%%%%%%%%%%%%%%%%%%%%%%%%
\subsubsection{Complete Proof}\label{sec:actual_proof_lb}

In this \app, we present the proof of Theorem~\ref{thm:mle_ub_lb}\ref{part:lb}. The proof reduces the general case to the $2$-item case presented in \App~\ref{sec:toy_lb}. In the reduction, we construct an ``oracle'' \mle, such that the oracle \mle yields identical estimates for item $2$ through item $\numitems$. Specifically, we consider an unconstrained oracle denoted by $\paramoracleunconstrain$ (without the box constraint), and a constrained oracle denoted by $\paramoracleclip$ (with the box constraint at $\boundgt$), to be defined precisely in the proof shortly. Then we derive the closed-form expressions for $\paramoracleunconstrain$ and $\paramoracleclip$, which bear resemblance to the expressions of the the unconstrained \mle and the \standardmle in the $2$-item case. Using the proof of the $2$-item case, we prove that the constrained oracle $\paramoracleclip$ incurs a negative bias of  $\bigOmega(\frac{1}{\sqrt{\numitems\numcomparisons}})$. Given this result, it remains to show that $\paramoracleclip$ and $\paramestboundgt$ differ by $\smallo(\frac{1}{\sqrt{\numitems\numcomparisons}})$ in terms of bias. We decompose the difference between $\paramoracleclip$ and $\paramestboundgt$ into three terms: from $\paramoracleclip$ to $\paramoracleunconstrain$, from $\paramoracleunconstrain$ to $\paramestunconstrain$, and from $\paramestunconstrain$ to $\paramestboundgt$, The second term is bounded by
$\bigOlog(\frac{1}{\numitems\numcomparisons})$ by modifying the upper-bound proof of Theorem~\ref{thm:mle_ub_lb}\ref{part:ub}. The first and the third terms are bounded by carefully analyzing the effect of the box constraint on the oracle \mle and the \standardmle, respectively.

In the proof, we fix any constant $\boundgt>0$, and consider the true parameter vector:
\begin{align}\label{eq:construction_param_gt}
    \paramgt=\left[\boundgt, -\frac{\boundgt
    }{\numitems-1}, -\frac{\boundgt}{\numitems-1}, \ldots, -\frac{\boundgt}{\numitems-1}\right].
\end{align}
It can be verified that $\paramgt$ satisfies both the box constraint at $\boundgt$ and the centering constraint, so we have $\paramgt\in \rangeboxgt$. We prove that the bias on item $1$ is negative, and its magnitude is $\bigOmega(\frac{1}{\sqrt{\numitems\numcomparisons}})$. That is, we prove that
\begin{align*}
    \Expect[\paramestboundgt_1] - \paramgt_1 =  \Expect[\paramestboundgt_1]  - \boundgt \le -\frac{\const}{\sqrt{\numitems\numcomparisons}},
\end{align*}
for some constant $\const > 0$. The proof consists of the following $5$ steps.

    \begin{enumeratex}[label={\bfseries Step \arabic*:}, ref=\arabic*]
        \item \textbf{Construct oracle estimators $\paramoracleunconstrain$ (unconstrained) and $\paramoracleclip$ (constrained)}\label{step:lb_oracle}
        
        Recall that $\meanemp_{\idxitem\idxitemalt}\sample \frac{1}{\numcomparisons}\binomial(\numcomparisons, \meangt_{\idxitem\idxitemalt})$ is a random variable representing the fraction of times that item $\idxitem$ beats item $\idxitemalt$. We define $\meanemp_1$ as fraction of wins by item $1$, among all comparisons in which item $1$ is involved:
        \begin{align}
            \meanemp_1\defn \frac{1}{\numitems-1}\sum_{\idxitemthree=2}^{\numitems} \mean_{1\idxitemthree}.\label{eq:lb_definition_mean_item_one}
        \end{align}
        We similarly define the true probability $\meangt_1 = \frac{1}{\numitems-1}\sum_{\idxitemthree=2}^\numitems\meangt_{1\idxitemthree}$. With the construction~\eqref{eq:construction_param_gt} of $\paramgt$, we have $\meangt_1 = \frac{1}{1 + e^{-\frac{\numitems}{\numitems-1}}\boundgt}$. Now we construct the following random quantities $\{\meanoracle_{\idxitem\idxitemalt}\}_{\idxitem\ne \idxitemalt}$ as a function of $\{\meanemp_{\idxitem\idxitemalt}\}_{\idxitem\ne \idxitemalt}$:
        \begin{align}\label{eq:oracle_probabilities}
            \meanoracle_{\idxitem\idxitemalt} = \begin{cases}
                \meanemp_1 & \text{if }\idxitem=1,\; \idxitemalt\in \{2, \ldots, \numitems\}\\
                1 - \meanemp_1 & \text{if } \idxitemalt=1,\; \idxitem\in \{2, \ldots, \numitems\}\\
                \frac{1}{2} & \text{otherwise}.
            \end{cases}
        \end{align}
        Recall that $\paramestunconstrain(\{\meanemp_{\idxitem\idxitemalt}\})$ denotes the unconstrained \mle~\eqref{eq:unconstrained_mle}. Now define an ``unconstrained oracle'' \mle $\paramoracleunconstrain$ as:
        \begin{subequations}
        \begin{align}
            \paramoracleunconstrain(\{\meanemp_{\idxitem\idxitemalt}\}) \defn & \; \paramestunconstrain(\{\meanoracle_{\idxitem\idxitemalt}\}) \nonumber\\
            = & \argmin_{\param\in \rangeboxinf} \negloglikelihood(\{\meanoracle_{\idxitem\idxitemalt}\}; \param).\label{eq:defn_unconstrained_oracle}
        \end{align}
        Similarly, define a ``constrained oracle'' \mle $\paramoracleclip$ as:
        \begin{align}
            \paramoracleclip(\{\meanemp_{\idxitem\idxitemalt}\}) \defn & \; \paramestboundgt(\{\meanoracle_{\idxitem\idxitemalt}\}) \nonumber\\
            = & \argmin_{\param\in \rangeboxgt} \negloglikelihood(\{\meanoracle_{\idxitem\idxitemalt}\}; \param).
        \end{align}
        \end{subequations}
        In the subsequent steps, these oracle estimators are used to reduce the general case to the $2$-item case.

        \item \textbf{Formalize the oracle information contained in the unconstrained oracle $\paramoracleunconstrain$ and the constrained oracle $\paramoracleclip$}\label{step:lb_constrained_unconstrained_oracle}
        
        Note that the construction of $\{\meanoracle_{\idxitem\idxitemalt}\}$ in~\eqref{eq:oracle_probabilities} is symmetric with respect to item $2$ through item $\numitems$, that is, for any two items $\idxitem$ and $\idxitem'$ where $\idxitem, \idxitem' \in \{2, \ldots, \numitems\}$, we have $\meanoracle_{\idxitem\idxitemalt} = \meanoracle_{\idxitem'\idxitemalt}$ and $\meanoracle_{\idxitemalt\idxitem} = \meanoracle_{\idxitemalt\idxitem'}$ for every $\idxitem\in [\numitems]\setminus \{\idxitemalt, \idxitemalt'\}$. Therefore, the construction of $\{\meanoracle_{\idxitem\idxitemalt}\}$ intuitively encodes the ``oracle'' that item $2$ through item $\numitems$ have identical parameters. Formally, define the set $\rangeboxoracle \defn \{\param\in \reals^\numitems \given \param_2 = \cdots = \param_\numitems\}$. The following lemma states that the unconstrained oracle and the constrained oracle incorporate the set $\rangeboxoracle$ into the domain of optimization without altering their solutions.
        \begin{lemma}\label{lem:symmetric_oracle}
        \begin{subequations}
        The unconstrained oracle $\paramoracleunconstrain$ can be equivalently written as
        \begin{align}
            \paramoracleunconstrain = \argmin_{\rangeboxinf \cap \rangeboxoracle} \negloglikelihood(\{\meanoracle_{\idxitem\idxitemalt}\}; \param).\label{eq:defn_unconstrained_oracle_equiv}
        \end{align}
        That is, a solution to~\eqref{eq:defn_unconstrained_oracle} exists if and only if a solution to~\eqref{eq:defn_unconstrained_oracle_equiv} exists. Moreover, when the solutions to~\eqref{eq:defn_unconstrained_oracle} and~\eqref{eq:defn_unconstrained_oracle_equiv} exist, they are identical.
        
        Similarly, the constrained oracle $\paramoracleclip$ can be equivalently written as
        \begin{align}
            \paramoracleclip = & \argmin_{\param\in \rangeboxgt\cap \rangeboxoracle} \negloglikelihood(\{\meanoracle_{\idxitem\idxitemalt}\}; \param).
        \end{align}
        \end{subequations}
        \end{lemma}
        See \App~\ref{sec:proof_lem_symmetric_oracle} for the proof of Lemma~\ref{lem:symmetric_oracle}.
    
        Given Lemma~\ref{lem:symmetric_oracle} combined with the centering constraint, we parameterize the unconstrained oracle $\paramoracleunconstrain$ and the constrained oracle $\paramoracleclip$ as:
        \begin{subequations}
        \begin{align}
            \paramoracleunconstrain & = \left[\paramoracleunconstrain_1, -\frac{1}{\numitems-1}\paramoracleunconstrain_1, \ldots, -\frac{1}{\numitems-1}\paramoracleunconstrain_1\right],\label{eq:reparam_unconstrained_oracle}\\
            \paramoracleclip & = \left[\paramoracleclip_1, -\frac{1}{\numitems-1}\paramoracleclip_1, \ldots, -\frac{1}{\numitems-1}\paramoracleclip_1\right].\label{eq:reparam_constrained_oracle}
        \end{align}
        \end{subequations}
        
        %%%%%%%%%%%%
        \item \textbf{Show that the bias of the constrained oracle $\paramoracleclip$ on item $1$ is bounded by $\Expect[\paramoracleclip_1] - \paramgt_1 \le -\frac{\const}{\sqrt{\numitems\numcomparisons}}$, by making a reduction to the $2$-item case}\label{step:lb_two_item_reduction}
        
        In this step, we modify the proof of Proposition~\ref{prop:lb_two_item_negative_bias} in the $2$-item case to lower bound the bias of the constrained oracle $\paramoracleclip$. Specifically, we show that given $\paramgt = \left[\boundgt, -\frac{\boundgt}{\numitems-1}, \ldots, -\frac{\boundgt}{\numitems-1}\right]$, the bias on item $1$ is bounded as (cf.~\eqref{eq:toy_lb_desired_negative_bias}):
        \begin{align*}
            \Expect[\paramoracleclip_1] - \paramgt \le -\frac{\const}{\sqrt{\numitems\numcomparisons}},
        \end{align*}
        for some constant $\const > 0$.
        
      First, we solve for the unconstrained oracle $\paramoracleunconstrain$ and the constrained oracle $\paramoracleclip$ in closed form. Set $\idxitemthree=1$ in the gradient expression~\eqref{eq:log_likelihood_gradient_raw}. Plugging in the expressions for the unconstrained oracle $\paramoracleunconstrain$~\eqref{eq:reparam_unconstrained_oracle} and the manipulated observations $\{\meanoracle_{\idxitem\idxitemalt}\}$~\eqref{eq:oracle_probabilities}, we have
        \begin{align}
            \left.\lfrac{\partial \negloglikelihood}{\partial \param_1}\right|_{\param = \paramoracleunconstrain} & = \numcomparisons (\numitems-1)\left[\frac{1}{1 + e^{-\frac{\numitems}{\numitems-1}\paramoracleunconstrain_1}}-\meanemp_1\right]\label{eq:lb_derivative_unconstrained_oracle}
        \end{align}
        Setting the derivative~\eqref{eq:lb_derivative_unconstrained_oracle} to $0$, we have
        \begin{align}
           \frac{1}{1 + e^{-\frac{\numitems}{\numitems-1}\paramoracleunconstrain_1}} & = \mean_1 \nonumber\\
            \paramoracleunconstrain_1 & = -\frac{\numitems-1}{\numitems} \log\left(\frac{1}{\meanemp_1} - 1\right).\label{eq:lb_unconstrained_oracle_mle_expression}
        \end{align}

        Denote $\meanmaxd = \meangt_1 = \frac{1}{1 + e^{-\frac{\numitems}{\numitems-1}\boundgt}}$, and $\meanmind = 1 - \meanmaxd = \frac{1}{1 + e^{\frac{\numitems}{\numitems-1}\boundgt}}$. In the notations $\meanmaxd$ and $\meanmind$, the dependency on $\numitems$ is made explicit. When the dependency on $\numitems$ does not need to be emphasized, we also use the shorthand notations $\meanmax$ and $\meanmin$. Now consider the constrained oracle $\paramoracleclip$. By straightforward analysis, one can derive the following closed-form expression for the constrained oracle:
        \begin{align}
            \paramoracleclip_1(\mean_1) & = \begin{cases}
            -\boundgt & \text{if } 0 \le \mean_1 < \meanmind\\
            -\frac{\numitems-1}{\numitems} \log\left(\frac{1}{\mean_1} - 1\right) &\text{if } \meanmind < \mean_1 < \meanmaxd\\
            \boundgt & \text{if } \meanmaxd \le \mean_1 \le 1 .
            \end{cases}\label{eq:lb_func_log_oracle}
        \end{align}
        Note the similarity between $\paramoracleclip$ in~\eqref{eq:lb_func_log_oracle} and the $2$-item case $\paramestboundgt_1$ in~\eqref{eq:toy_lb_def_func_log} from \App~\ref{sec:toy_lb}. Similar to the function $\funclog$ defined in~\eqref{eq:toy_lb_def_func_log_free} of the $2$-item case, we denote a function $\funclog_\numitems: [0, 1]\rightarrow [-\boundgt, \boundgt]$ as:
        \begin{align*}
            \funclog_\numitems(\varmean) & = \begin{cases}
            -\boundgt & \text{if } 0 \le \varmean < \meanmind\\
            -\frac{\numitems-1}{\numitems} \log\left(\frac{1}{\varmean} - 1\right) &\text{if } \meanmind < \varmean < \meanmaxd\\
            \boundgt & \text{if } \meanmaxd \le \varmean \le 1,
            \end{cases}
        \end{align*}
        where $\funclog_\numitems(\varmean) = \paramoracleclip_1(\mean_1=\varmean)$ for any $\varmean\in [0, 1]$. Then the estimator $\paramoracleclip_1(\mean)$ can be equivalently written as $\funclog_\numitems(\mean)$. Similar to the function $\funclogplus$ defined in~\eqref{eq:toy_lb_def_func_log_plus} of the $2$-item case, we define an auxiliary function $\funclogplusd: [0, 1] \rightarrow [-\boundgt, \boundgt]$ as:
         \begin{align*}
           \funclogplusd(\varmean) = \begin{cases}
            \frac{2\boundgt}{\meanmaxd} \left(\varmean - \meanmaxd\right) + \boundgt & \text{if } 0 \le \varmean < \meanmaxd\\
            \boundgt & \text{if } \meanmaxd \le \varmean \le 1.
            \end{cases}
        \end{align*}
        
        Note that in the proofs of Lemma~\ref{lem:toy_lb_surrogate_greater} and Lemma~\ref{lem:toy_lb_surrogate_desired_bias}, we have only relied on the following two facts:
        \begin{itemizex}
            \item There exists a constant $\const$ such that \begin{align*}
            \frac{1}{2} < \meanmax < \const < 1.
            \end{align*}
            \item The random variable $\meanemp$ is sampled as $\meanemp\sample \frac{1}{\numcomparisons}\binomial(\numcomparisons, \meanmax)$.
        \end{itemizex}
        In the general case, it can be verified that
        \begin{itemizex}
            \item There exists a constant $\const$ such that \begin{align*}
            \frac{1}{2} < \meanmaxd < \const < 1, \qquad \text{for all }\numitems\ge 2.
            \end{align*}

            \item The random variable $\meanemp_1$ as defined in~\eqref{eq:oracle_probabilities} is sampled as $\meanemp_1\sample \frac{1}{\numcomparisons'}\binomial(\numcomparisons', \meanmax)$, where $\numcomparisons' \defn (\numitems-1)\numcomparisons$ denotes the total number of comparisons in which item $1$ is involved.
        \end{itemizex}
        To extend the arguments in the $2$-item case to the general case, we replace $\meanemp$ by $\meanemp_1$, replace $\meanmax$ by $\meanmaxd$, replace $\funclogplus$ by $\funclogplusd$, and replace $\numcomparisons$ by $\numcomparisons'$ in the proof of Proposition~\ref{prop:lb_two_item_negative_bias}. It can be verified that the arguments in Lemma~\ref{lem:toy_lb_surrogate_greater} and Lemma~\ref{lem:toy_lb_surrogate_desired_bias} still hold after these replacements. Therefore, extending the arguments in Proposition~\ref{prop:lb_two_item_negative_bias}, we have that at $\paramgt = \left[\boundgt, -\frac{\boundgt}{\numitems-1}, \ldots, -\frac{\boundgt}{\numitems-1}\right]$,
        \begin{align}
            \Expect[\paramoracleclip_1]-\paramgt_1 & \le - \frac{\const}{\sqrt{\numcomparisons'}} = -\frac{\const}{\sqrt{(\numitems-1)\numcomparisons}}  \le -\frac{\const'}{\sqrt{\numitems\numcomparisons}}, \label{eq:negative_bias_two_items}
        \end{align}
        for some constants $\const, \const' > 0$.
        
         \item \textbf{Bound the difference between the unconstrained oracle $\paramoracleunconstrain$ and the unconstrained \mle $\paramestunconstrain$, by modifying the proof of Theorem~\ref{thm:mle_ub_lb}\ref{part:ub}}\label{step:lb_oracle_vs_regular}
        
         Recall that the random variable $\meanemp_1$ denotes the fraction of wins by item $1$. In this step, we fix any real number $\meancondition\in [\frac{1}{2}, \meanmax]$, and denote $\eventmean$ as the event that we observe $\meanemp_1 = \meancondition$. Then we prove that conditioned on the event $\eventmean$, the difference between the unconstrained oracle $\paramoracleunconstrain$ and the unconstrained \mle $\paramestunconstrain$ is small in expectation, by modifying Step~\ref{step:concentration} to Step~\ref{step:ub_bound_expectation} in the upper-bound proof of Theorem~\ref{thm:mle_ub_lb}\ref{part:ub} in \App~\ref{sec:actual_proof_ub}.
        
        We first conceptually explain how to modify the proof of Theorem~\ref{thm:mle_ub_lb}\ref{part:ub}.  Our goal is to bound the difference between $\paramoracleunconstrain$ and $\paramestunconstrain$ in expectation conditioned on the event $\eventmean$. By the definition of $\{\meanoracle_{\idxitem\idxitemalt}\}$ in~\eqref{eq:oracle_probabilities}, the quantities $\{\meanoracle_{\idxitem\idxitemalt}\}$ are fixed (not random) conditioned on $\eventmean$, and hence the unconstrained oracle $\paramoracleunconstrain$ is fixed conditioned on $\eventmean$. We therefore replace the role of the true parameter vector $\paramgt$ in the proof of Theorem~\ref{thm:mle_ub_lb}\ref{part:ub} by the unconstrained oracle $\paramoracleunconstrain$. Then we think of the actual observations $\{\meanemp_{\idxitem\idxitemalt}\}$ as a noisy version of $\{\meanoracle_{\idxitem\idxitemalt}\}$, and think of $\paramestunconstrain$ as the estimate for $\paramoracleunconstrain$. Now we modify the proof of Theorem~\ref{thm:mle_ub_lb}\ref{part:ub} to bound the expected difference between $\paramestunconstrain$ and $\paramoracleunconstrain$ conditioned on $\eventmean$. At the end of this step, we provide more intuition why we need to condition on the event $\eventmean$.
        
        Formally, we denote      $\{\meanoracleevent_{\idxitem\idxitemalt}\}$ as the values of $\{\meanoracle_{\idxitem\idxitemalt}\}$ conditional on $\eventmean$. We denote $\paramoracleevent$ as the unconstrained oracle $\paramoracleunconstrain$ conditional on $\eventmean$. It can be verified that $\{\meanoracleevent_{\idxitem\idxitemalt}\}$ and $\paramoracleevent$ are fixed (not random) given any $\meancondition\in [\frac{1}{2}, \meanmax]$. Conditioned on $\eventmean$, we think of $\paramoracleevent$ \emph{as if} it is the ``true'' parameter vector to be estimated (replacing the role of $\paramgt$), and think of $\{\meanoracleevent_{\idxitem\idxitemalt}\}$ \emph{as if} it is the ``true'' underlying probabilities (replacing the role of $\{\meangt_{\idxitem\idxitemalt}\}$).
        
        Given the definition of $\{\meanoracle_{\idxitem\idxitemalt}\}$ in~\eqref{eq:oracle_probabilities}, we have that conditioned on event $\eventmean$, 
        \begin{align}\label{eq:defn_oracle_probabilities_condition}
            \meanoracleevent_{\idxitem\idxitemalt} = \begin{cases}
                \meancondition & \text{if }\idxitem=1,\; \idxitemalt\in \{2, \ldots, \numitems\}\\
                1 - \meancondition & \text{if } \idxitemalt=1,\; \idxitem\in \{2, \ldots, \numitems\}\\
                \frac{1}{2} & \text{otherwise}.
            \end{cases}
        \end{align}

        From the expression~\eqref{eq:lb_unconstrained_oracle_mle_expression} of the unconstrained oracle $\paramoracleunconstrain$, it can be verified that $\paramoracleunconstrain$ satisfies  the deterministic equality
        \begin{align}
            \frac{1}{1+ e^{-\left(\paramoracleunconstrain_\idxitem-\paramoracleunconstrain_\idxitemalt\right)}} = \meanoracle_{\idxitem\idxitemalt} ,\qquad \text{for all } \idxitem\ne \idxitemalt.\label{eq:oracle_probabilities_equals_unconstrained_oracle_estimate}
        \end{align}
        Now we start to replicate Step~\ref{step:concentration} to Step~\ref{step:ub_bound_expectation} in the proof of Theorem~\ref{thm:mle_ub_lb}\ref{part:ub} presented in \App~\ref{sec:actual_proof_ub}. 
        
        %%%%%%%%%%%%
        To replicate \emph{Step~\ref{step:concentration}} of Theorem~\ref{thm:mle_ub_lb}\ref{part:ub}, recall that in the proof of Theorem~\ref{thm:mle_ub_lb}\ref{part:ub}, we condition on Lemma~\ref{lem:finite_solution} and Lemma~\ref{lem:concentration}. We first establish the modified versions of these two lemmas, when conditioned on $\eventmean$.
        
        \begin{lemma}[Conditional version of Lemma~\ref{lem:finite_solution}]\label{lem:finite_solution_conditioning}
            Conditioned on the event $\eventmean$, there exists a finite solution $\paramestunconstrain$ to the unconstrained \mle~\eqref{eq:unconstrained_mle} \whponeconditionmean.
        \end{lemma}
        See \App~\ref{sec:proof_lem_finite_solution_conditioning} for the proof of Lemma~\ref{lem:finite_solution_conditioning}.
        
        \begin{lemma}[Conditional version of Lemma~\ref{lem:concentration}]\label{lem:concentration_conditioning}
        Conditioned on the event $\eventmean$, there exists a constant $\const > 0$, such that
        \begin{align}
            \abs*{\sum_{\idxitem\ne \idxitemthree}\meanemp_{\idxitemthree\idxitem} -\sum_{\idxitem\ne \idxitemthree} \meanoracleevent_{\idxitemthree\idxitem}} \le \const\sqrt{\frac{\numitems(\log\numitems+\log\numcomparisons)}{\numcomparisons}},\label{eq:lb_mcdiardmid_desired}
        \end{align}
        simultaneously for all $\idxitemthree \in [\numitems]$ \whponeconditionmean.
        \end{lemma}
        See \App~\ref{sec:proof_lem_concentration_conditioning} for the proof of Lemma~\ref{lem:concentration_conditioning}.
        
        Recall that we have conditioned on the event $\eventmean$. Denote $\eventlemmalb$ as the event that  Lemma~\ref{lem:finite_solution_conditioning} and Lemma~\ref{lem:concentration_conditioning} both hold. (Note that the event $\eventlemmalb$ is defined for some fixed $\meancondition$, so to be precise, the event $\eventlemmalb$ should be denoted as $\eventlemmalb{}_{,\meancondition}$. For ease of notation, we drop the subscript $\meancondition$.) Taking a union bound of Lemma~\ref{lem:finite_solution_conditioning} and Lemma~\ref{lem:concentration_conditioning}, we have that $\eventlemmalb$ happens \whponeconditionmean. For the rest of the proof, we condition on the events $(\eventlemmalb, \eventmean)$.
        
        %%%%%%%%%%%%
        To replicate \emph{Step~\ref{step:first_order_optimality}} of Theorem~\ref{thm:mle_ub_lb}\ref{part:ub}, we subtract equality~\eqref{eq:oracle_probabilities_equals_unconstrained_oracle_estimate} from both sides of~\eqref{eq:first_order_optimality_raw}. We obtain the (unconditional) deterministic equality:
        \begin{align}
            \sum_{\idxitem=1}^\numitems \left(\frac{1}{1 + e^{-( \paramestunconstrain_\idxitemthree- \paramestunconstrain_\idxitem)}} - \frac{1}{1 + e^{-(\paramoracleunconstrain_\idxitemthree - \paramoracleunconstrain_\idxitem)}}\right) = \sum_{\idxitem\ne \idxitemthree} (\meanemp_{\idxitemthree\idxitem} - \meanoracle_{\idxitemthree\idxitem}),\qquad \text{for every } \idxitemthree\in [\numitems].\label{eq:lb_first_order_optimality_before_condition}
        \end{align}
        Conditioning~\eqref{eq:lb_first_order_optimality_before_condition} on $(\eventlemmalb, \eventmean)$, we have the following deterministic equality, as a modified version of~\eqref{eq:first_order_optimality}:
        \begin{align}
            \sum_{\idxitem=1}^\numitems \left(\frac{1}{1 + e^{-(\paramestunconstrain_\idxitemthree-\paramestunconstrain_\idxitem)}} - \frac{1}{1 + e^{-(\paramoracleevent_\idxitemthree-\paramoracleevent_\idxitem)}}\right) = \sum_{\idxitem\ne \idxitemthree} (\meanemp_{\idxitemthree\idxitem} - \meanoracleevent_{\idxitemthree\idxitem}), \qquad \text{conditioned on } (\eventlemmaub, \eventmean).\label{eq:lb_first_order_optimality_condition}
        \end{align}
        
        To replicate \emph{Step~\ref{step:ub_bound_uniform}} of Theorem~\ref{thm:mle_ub_lb}\ref{part:ub}, note that $\meancondition$ is bounded as $\meancondition \in [\frac{1}{2}, \meanmax]$. By the expression~\eqref{eq:lb_unconstrained_oracle_mle_expression} of $\paramoracleunconstrain$ (and hence of $\paramoracleevent$), it can be verified that $\paramoracleevent$ is bounded as $\abs{\paramoracleevent} \le \const$ for some constant $\const$. Denote $\difcondition = \paramestunconstrain - \paramoracleevent$. Using the same arguments as in Lemma~\ref{lem:bound_diff_each}, we have the deterministic relation that
        \begin{align}\label{eq:lb_analogy_uniform_bound}
            \norm{\paramestunconstrain - \paramoracleevent}_\infty = \norm{\difcondition}_\infty \lessorder \sqrt{\frac{\log\numitems + \log\numcomparisons}{\numitems\numcomparisons}}, \qquad \text{conditioned on } (\eventlemmalb, \eventmean).
        \end{align}
        
        To replicate \emph{Step~\ref{step:ub_bound_expectation}} of Theorem~\ref{thm:mle_ub_lb}\ref{part:ub}, we first apply the second-order mean value theorem on~\eqref{eq:lb_first_order_optimality_condition}, and then take an expectation conditional on ($\eventlemmalb, \eventmean)$. The following equation establishes a modified version of~\eqref{eq:ub_mvt_order_two_expectation_raw}:
        \begin{align}
             \sum_{\idxitem=1}^\numitems \funcsigmoid'(\paramoracleevent_\idxitemthree-\paramoracleevent_\idxitem)\cdot \Expect\left[ \difcondition_\idxitem - \difcondition_\idxitemthree\given \eventlemmalb, \eventmean \right]  &= \nonumber\\
            \sum_{\idxitem\ne \idxitemthree}(\Expect[\meanemp_{\idxitemthree\idxitem}\given & \eventlemmalb, \eventmean]-\meanoracleevent_{\idxitemthree\idxitem}) - %
            \frac{1}{2}\sum_{\idxitem=1}^\numitems\Expect[\funcsigmoid''(\meanvalue_{\idxitemthree\idxitem})(\difcondition_\idxitemthree-\difcondition_\idxitem)^2
            \given \eventlemmalb, \eventmean],\label{eq:lb_expect_diff_mvt_order_two}
        \end{align}
        where each $\meanvalue_{\idxitemthree\idxitem}$ is a random variable that takes values between $\paramoracleevent_\idxitemthree-\paramoracleevent_\idxitem$ and $\paramoracleevent_\idxitemthree-\paramoracleevent_\idxitem + \difcondition_\idxitemthree-\difcondition_\idxitem$.
        To apply Lemma~\ref{lem:ub_conditional_expectation}, we set $\event$ as $\eventmean$, and set $\event'$ as $\eventlemmalb$ in~\eqref{eq:ub_empirical_prob_expectation_bound}:
        \begin{align}
            \abs{\Expect[\meanemp_{\idxitem\idxitemalt} \given \eventlemmalb, \eventmean] - \Expect[\meanemp_{\idxitem\idxitemalt}\given \eventmean]} \lessorder\frac{1}{\numitems\numcomparisons}.\label{eq:lb_mean_condition_bound}
        \end{align}
        It can be verified that
        \begin{align}
            \Expect[\meanemp_{\idxitem\idxitemalt} \given \eventmean] = \meanoracleevent_{\idxitem\idxitemalt}.\label{eq:lb_mean_oracle_condition_equality}
        \end{align}
        Plugging~\eqref{eq:lb_mean_oracle_condition_equality} into~\eqref{eq:lb_mean_condition_bound}, we have     
        \begin{align*}
         \abs{\Expect[\meanemp_{\idxitem\idxitemalt} \given \eventlemmalb, \eventmean] - \meanoracleevent_{\idxitem\idxitemalt}} \lessorder\frac{1}{\numitems\numcomparisons}.
        \end{align*}
        Using the same arguments as in Lemma~\ref{lem:ub_expectation_bound_each} to handle the remaining terms in~\eqref{eq:lb_expect_diff_mvt_order_two}, we have the following upper bound as a modified version of~\eqref{eq:ub_condition_final}:
        \begin{align}\label{eq:lb_bound_diff_est_oracle_expectation}
            \norm{\Expect[\paramestunconstrain  - \paramoracleevent\given \eventlemmalb, \eventmean]}_\infty  = \norm{\Expect[\paramestunconstrain  - \paramoracleunconstrain\given \eventlemmalb, \eventmean]}_\infty \lessorder \frac{\log\numitems+\log\numcomparisons}{\numitems\numcomparisons}.
        \end{align}~\\

        Now that we have established the desired result~\eqref{eq:lb_bound_diff_est_oracle_expectation} of this step, we conclude this step with some intuition why we need to condition on $\eventmean$. Without conditioning on $\eventmean$, we could still have utilized the proof of Theorem~\ref{thm:mle_ub_lb}\ref{part:ub}, and could have established a result of the form (cf.~\eqref{eq:lb_bound_diff_est_oracle_expectation}):
        \begin{align}
            \norm{\Expect[\paramestunconstrain -\paramoracleevent \given \eventlemmalb]}_\infty =\norm{\Expect[\paramestunconstrain -\paramoracleunconstrain \given \eventlemmalb]}_\infty \lessorder \frac{\log\numitems+\log\numcomparisons}{\numitems\numcomparisons}.\label{eq:lb_wrong}
        \end{align}
        Our goal here is to bound the constrained oracle $\paramestboundgt$ and the constrained \mle $\paramoracleclip$ in expectation. However, the fact that two \emph{unconstrained} estimators are close in expectation does not imply that their \emph{constrained} counterparts are close in expectation\footnote{
            For example, consider the following two univariate estimators. The first estimator always outputs a value within $[-\boundgt, \boundgt]$. The second estimator sometimes outputs a value within $[-\boundgt, \boundgt]$, and sometimes outputs a value greater than $\boundgt$. The two estimators could be constructed such that they are close (or equal) in expectation. However, now consider their constrained counterparts. The first estimator is not affected by a box constraint at $\boundgt$, whereas the expected value of second estimator can become significantly smaller due to the box constraint. Therefore, the constrained counterparts of these two estimators may not be close in expectation. 
        }.
        Therefore, a bound of the form~\eqref{eq:lb_wrong} is not sufficient for our goal, and instead we need to establish some ``pointwise'' control between $\paramestunconstrain$ and $\paramoracleunconstrain$. That is, whenever the box constraint has little effect on $\paramoracleunconstrain$, we want to show that the box constraint also has little effect on $\paramestunconstrain$. Thus, we condition on the event $\eventmean$ for any $\meancondition\in [\frac{1}{2}, \meanmax]$, and bound the difference between $\paramestunconstrain$ and $\paramoracleunconstrain$ in expectation conditioned on $\eventmean$ (that is, the bound in~\eqref{eq:lb_bound_diff_est_oracle_expectation}). Given this pointwise result, we then integrate over $\meancondition$ to establish the desired result that $\paramestboundgt$ and $\paramoracleclip$ are close in expectation, to be presented in the subsequent step of the proof.

        \item \textbf{Bound the expected difference between $\paramestboundgt$ and $\paramoracleclip$, by making a connection between $\paramestboundgt-\paramoracleclip$ and $\paramestunconstrain-\paramoracleunconstrain$}\label{step:lb_constrained_unconstrained_mle}

        We decompose the bias of the \standardmle $\paramestboundgt$ as 
        \begin{align}\label{eq:bias_decompose}
            \Expect[\paramestboundgt_1] - \paramgt_1 = 
                (\Expect[\paramoracleclip_1] - \paramgt_1)
            + \Expect[\paramestboundgt_1 - \paramoracleclip_1].
        \end{align}
        Recall from~\eqref{eq:negative_bias_two_items} that
        \begin{align}
            \Expect[\paramoracleclip_1] - \paramgt_1\le -\frac{\const}{\sqrt{\numitems\numcomparisons}}.\label{eq:negative_bias_two_items_recall}
        \end{align}
        In what follows, we prove that
        \begin{align}
            \Expect[\paramestboundgt_1 - \paramoracleclip_1] \le \const'\frac{\log\numitems+\log\numcomparisons}{\numitems\numcomparisons}.\label{eq:lb_diff_constrained_unconstrained_expectation_desired}
        \end{align}
        Then plugging~\eqref{eq:negative_bias_two_items_recall} and~\eqref{eq:lb_diff_constrained_unconstrained_expectation_desired} back into~\eqref{eq:bias_decompose} yields
        \begin{align*}
            \Expect[\paramestboundgt_1] - \paramgt_1 \le -\frac{\const}{\sqrt{\numitems\numcomparisons}} + \const'\frac{\log\numitems+\log\numcomparisons}{\numitems\numcomparisons} \le -\frac{\const''}{\sqrt{\numitems\numcomparisons}},
        \end{align*}
        for all $\numitems\ge \constitems$ and $\numcomparisons \ge \constcomparisons$ where $\constitems$ and $\constcomparisons$ are constants, completing the proof of Theorem~\ref{thm:mle_ub_lb}\ref{part:lb}.~~\\
        
        The rest of this step is devoted to proving~\eqref{eq:lb_diff_constrained_unconstrained_expectation_desired}. To bound $\Expect[\paramestboundgt_1 - \paramoracleclip_1]$, we make a connection between $\paramestboundgt_1 - \paramoracleclip_1$ and $\paramoracleunconstrain_1 - \paramestunconstrain_1$, and then we evoke the bound on $\paramoracleunconstrain_1 - \paramestunconstrain_1$ from~\eqref{eq:lb_bound_diff_est_oracle_expectation} in Step~\ref{step:lb_oracle_vs_regular}.
        
        Recall that $\meanemp_1$ is a discrete random variable representing the fraction of wins by item $1$. By the law of iterated expectation, we have
        \begin{align}
            \Expect[\paramestboundgt_1 - \paramoracleclip_1]  = & \underbrace{%
                \Expect\left[\paramestboundgt_1 - \paramoracleclip_1 \given \meanempboundary < \meanemp_1 < \meangt_1\right]\cdot \Prob\left(\meanempboundary < \meanemp_1 < \mean_1^*\right)
            }_{\term_1}\nonumber\\
            \qquad & + \underbrace{%
                \Expect[\paramestboundgt_1 - \paramoracleclip_1 \given \meanemp_1 \ge \mean_1^*]\cdot \Prob\left(\meanemp_1 \ge \mean_1^* \right)
            }_{\term_2} + \underbrace{%
                \Expect\left[\paramestboundgt_1 - \paramoracleclip_1\given \meanemp_1 < \meanempboundary\right]\cdot \Prob\left(\meanemp_1 < \meanempboundary\right)
            }_{\term_3}.\label{eq:bias_decompose_three}
        \end{align}
        In what follows, we bound the terms $\term_1, \term_2$ and $\term_3$ separately.
        
        Consider the term $\term_2$. From the expression of $\paramoracleclip$ in~\eqref{eq:lb_func_log_oracle}, we have $\paramoracleclip_1 = \boundgt$ when $\meanemp_1 \ge \meangt_1$. Therefore, 
        \begin{align*}
             \Expect[\paramestboundgt_1 - \paramoracleclip_1 \given \meanemp_1 \ge \mean_1^*] = \Expect[\paramestboundgt_1 \given \meanemp_1 \ge \mean_1^*] - \boundgt \stackrel{\stepone}{\le} 0,
        \end{align*}
        where \stepone is true due to the box constraint $\abs{\paramestboundgt_1} \le \boundgt$. Hence, 
        \begin{align}
            \term_2 \le 0.\label{eq:lb_term_two}
        \end{align}

        Consider the term $\term_3$, we have $\Expect[\meanemp_1] = \meangt_1 =\frac{1}{1 + e^{-\frac{\numitems}{\numitems-1}\boundgt}}$, and therefore it can be verified that there exists a constant $\tau > 0$, such that $\meangt_1 > \frac{1}{2} + \tau$ for all  $\numitems\ge 2$. By Hoeffding's inequality, we have
        \begin{align}
            \Prob\left(\mean_1 < \frac{1}{2}\right) & < \Prob\left(\abs*{\meanemp_1 - \meangt_1} > \tau\right)\nonumber\\
            &  \le 2 \exp\left(-2(\numitems-1)\numcomparisons\tau^2\right) \lessorder \frac{1}{\numitems\numcomparisons}.\label{eq:lb_prob_mean_one_below_boundary}
        \end{align}
        Therefore, we have
        \begin{align}
            \term_3 & = \Expect\left[\paramestboundgt_1 - \paramoracleclip_1\given \meanemp_1 < \meanempboundary\right]\cdot \Prob\left(\meanemp_1 < \meanempboundary\right) \nonumber\\
            & \stackrel{\stepone}{\le} 2\boundgt \cdot \Prob\left(\meanemp_1 < \meanempboundary\right) \nonumber\\
            & \stackrel{\steptwo}{\lessorder} \frac{1}{\numitems\numcomparisons},\label{eq:lb_term_three}
        \end{align}
        where \stepone is true because $\abs{\paramestboundgt_1 - \paramoracleclip_1} \le \abs{\paramestboundgt_1} +\abs{\paramoracleclip_1} \le 2\boundgt$ by the box constraint, and \steptwo is true due to~\eqref{eq:lb_prob_mean_one_below_boundary}.
        
        Now consider the term $\term_1$. Denote $\eventlemmalbcomp$ as the complement of the event $\eventlemmalb$. Using the law of iterated expectation again, we have\begin{align}
            \term_1 =\Expect\left[\paramestboundgt_1 - \paramoracleclip_1 \given \frac{1}{2} < \meanemp_1 < \meangt_1\right] & \cdot \Prob\left(\meanempboundary < \meanemp_1 < \meangt_1\right)  =\nonumber\\ & \underbrace{%
                \Expect\left[\paramestboundgt_1 - \paramoracleclip_1  \given \eventlemmalb, \frac{1}{2} < \meanemp_1 < \meangt_1\right] \cdot \Prob\left(\eventlemmalb, \meanempboundary < \meanemp_1 < \meangt_1 \right)
            }_{\term_{11}} \nonumber\\
            +& \underbrace{%
                \Expect\left[\paramestboundgt_1 - \paramoracleclip_1  \given \eventlemmalbcomp, \meanempboundary < \meanemp_1 < \meangt_1\right] \cdot \Prob\left(\eventlemmalbcomp, \meanempboundary < \meanemp_1 < \meangt_1\right)
            }_{\term_{12}}\label{eq:lb_term_one_decompose}
        \end{align}
        
        Consider the term $\term_{12}$. We have
        \begin{align}
            \Prob\left(\eventlemmalbcomp, \meanempboundary < \meanemp_1 < \meangt_1\right) & = \sum_{\meancondition \in (\meanempboundary, \meangt_1)} \Prob(\eventlemmalbcomp\given \eventmean) \cdot \Prob(\eventmean)\nonumber\\
            & \stackrel{\stepone}{\le} \frac{\const}{\numitems\numcomparisons} \sum_{\meancondition \in (\meanempboundary, \meangt_1)}  \Prob(\eventmean) \nonumber\\
            & \lessorder \frac{1}{\numitems\numcomparisons},\label{eq:lb_term_one_two_prob}
        \end{align}
        where \stepone is true because $\eventlemmalb$ happens \whponecondition{\eventmean}.
        Combining~\eqref{eq:lb_term_one_two_prob} with the fact that $\abs{\paramestboundgt_1-\paramoracleclip_1} \le 2\boundgt$ due to the box constraint, we have
        \begin{align}
            \term_{12} \lessorder\frac{1}{\numitems\numcomparisons}.\label{eq:lb_term_one_two_bound}
        \end{align}

        Now consider the term $\term_{11}$. 
        We first analyze the constrained oracle $\paramoracleclip$. By the expression of $\paramoracleclip$ in~\eqref{eq:lb_func_log_oracle} and the expression of $\paramoracleunconstrain$ in~\eqref{eq:lb_unconstrained_oracle_mle_expression}, we have 
        \begin{align}
            \paramoracleclip = \paramoracleunconstrain, \qquad \text{conditioned on } \meanempboundary <\mean_1 < \meangt_1.\label{eq:lb_param_oracle_constrained_unconstrained}
        \end{align}
        Moreover, given $\frac{1}{2} < \meanemp_1 < \meangt_1$, by the expression of $\paramoracleclip$ in~\eqref{eq:lb_func_log_oracle},  we have 
        \begin{subequations}
        \begin{align*}
            0 < \paramoracleclip_1 < \boundgt
        \end{align*}
        and therefore by the parameterization of $\paramoracleclip$ in~\eqref{eq:reparam_constrained_oracle},
        \begin{align*}
            & \abs{\paramoracleclip_\idxitem} \le \frac{1}{\numitems-1}\boundgt\qquad \text{for every } \idxitem \in \{2,\ldots,\numitems\}.
        \end{align*}
        \end{subequations}
        Hence, there exists a constant $\tau' > 0$ such that
        \begin{subequations}\label{eq:lb_oracle_constrained_value_analysis}
        \begin{align}
            & \paramoracleclip_1 > -\boundgt+\tau'
        \end{align}
        and
        \begin{align}
            -\boundgt+\tau' < & \; \paramoracleclip_\idxitem < \boundgt-\tau' \qquad \text{for every } \idxitem\in \{2, \ldots, \numitems\}.
        \end{align}
        \end{subequations}
        
        Now we analyze the \standardmle $\paramestboundgt$. Recall that $\eventmean$ denotes the event that $\meanemp_1 = \meancondition$. We have that for every $\meancondition\in \left(\frac{1}{2}, \meangt_1\right)$,
        \begin{align}
        \norm{\paramestunconstrain_1 - \paramoracleclip_1}_\infty \stackrel{\stepone}{=} \norm{\paramestunconstrain_1 - \paramoracleunconstrain}_\infty \stackrel{\steptwo}{\lessorder} \sqrt{\frac{\log\numitems+\log\numcomparisons}{\numitems\numcomparisons}}, \qquad \text{conditioned on } (\eventlemmalb, \eventmean), \label{eq:lb_analogy_uniform_bound_recall}
        \end{align}
        where \stepone is true by~\eqref{eq:lb_param_oracle_constrained_unconstrained}, and \steptwo is true by~\eqref{eq:lb_analogy_uniform_bound} from Step~\ref{step:lb_oracle_vs_regular}. By~\eqref{eq:lb_analogy_uniform_bound_recall}, we have that for every $\meancondition\in \left(\frac{1}{2}, \meangt_1\right)$,
        \begin{align}\label{eq:lb_unconstrained_mle_constrained_oracle_diff}
            \norm{\paramestunconstrain_1 - \paramoracleclip_1}_\infty \le \tau',\qquad \text{conditioned on } (\eventlemmalb, \eventmean),
        \end{align}
        for all $\numitems\ge \constitems$ and all $\numcomparisons\ge \constcomparisons$, where $\constitems$ and $\constcomparisons$ are constants. Combining~\eqref{eq:lb_unconstrained_mle_constrained_oracle_diff} with~\eqref{eq:lb_oracle_constrained_value_analysis}, if the unconstrained \mle $\paramestunconstrain$ violates the box constraint, then only possible case is $\paramestunconstrain_1 > \boundgt$. Then either $\paramestunconstrain_1 = \paramestboundgt_1$ (when $\paramestunconstrain$ does not violate the box constraint) or $\paramestunconstrain_1 > \boundgt \ge \paramestboundgt_1$ (when $\paramestunconstrain$ violates the box constraint). Hence, for every $\meancondition\in (\meanempboundary, \meangt_1)$,
        \begin{align}
            \paramestunconstrain_1 \ge \paramestboundgt_1, \qquad \text{conditioned on } (\eventlemmalb, \eventmean) .\label{eq:lb_param_mle_constrained_unconstrained}
        \end{align}
        
        Combining~\eqref{eq:lb_param_oracle_constrained_unconstrained} and~\eqref{eq:lb_param_mle_constrained_unconstrained}, we have that for every $\meancondition\in (\meanempboundary, \meangt_1)$,
        \begin{align}\label{eq:lb_constrained_unconstrained_diff_inequality}
            \paramestboundgt - \paramoracleclip \le \paramestunconstrain - \paramoracleunconstrain,\qquad \text{conditioned on } (\eventlemmalb, \eventmean).
        \end{align}
       By the law of iterated expectation again, we have
        \begin{align}
            \term_{11} & =
            \sum_{\meancondition\in (\meanempboundary, \meangt_1)} \Expect[\paramestboundgt_1 - \paramoracleclip_1  \given \eventlemmalb,  \meanemp_1=\meancondition] \cdot \Prob(\eventlemmalb, \meanemp_1 = \meancondition) \nonumber\\
            & = \sum_{\meancondition\in (\meanempboundary, \meangt_1)} \Expect[\paramestboundgt_1 - \paramoracleclip_1  \given \eventlemmalb,  \eventmean] \cdot \Prob(\eventlemmalb, \eventmean) \nonumber\\
            & \stackrel{\stepone}{\le} \sum_{\meancondition\in (\meanempboundary, \meangt_1)} \Expect[\paramestunconstrain_1 - \paramoracleunconstrain_1  \given \eventlemmalb,  \eventmean] \cdot \Prob(\eventlemmalb, \eventmean) \nonumber\\
            & \stackrel{\steptwo}{\lessorder} \frac{\log\numitems+\log\numcomparisons}{\numitems\numcomparisons} \sum_{\meancondition\in (\meanempboundary, \meangt_1)}  \Prob(\eventlemmalb, \eventmean) \nonumber\\
            & \lessorder \frac{\log\numitems+\log\numcomparisons}{\numitems\numcomparisons},\label{eq:lb_term_one_one_bound}
        \end{align}
        where \stepone is true due to~\eqref{eq:lb_constrained_unconstrained_diff_inequality}, and \steptwo is true due to the bound~\eqref{eq:lb_bound_diff_est_oracle_expectation} from Step~\ref{step:lb_oracle_vs_regular}.
        
        Plugging the term $\term_{11}$ from~\eqref{eq:lb_term_one_one_bound} and $\term_{12}$ from~\eqref{eq:lb_term_one_two_bound} back to~\eqref{eq:lb_term_one_decompose}, we have
        \begin{align}
        \term_1  = \term_{11} + \term_{12} \lessorder \frac{\log\numitems+\log\numcomparisons}{\numitems\numcomparisons}.\label{eq:lb_term_one}
        \end{align}
        Finally, plugging the terms $\term_1$ from~\eqref{eq:lb_term_one}, $\term_2$ from~\eqref{eq:lb_term_two}, and $\term_3$ from~\eqref{eq:lb_term_three} back into~\eqref{eq:bias_decompose_three} yields
        \begin{align*}
            \Expect[\paramestboundgt_1 - \paramoracleclip_1] \lessorder \frac{\log\numitems+\log\numcomparisons}{\numitems\numcomparisons},
        \end{align*}
        completing the proof of~\eqref{eq:lb_diff_constrained_unconstrained_expectation_desired}.

    \end{enumeratex} % end proof steps
    
    \subsection{Proofs of Lemmas}\label{sec:proof_lemma}
    
    %% Proofs of lemmas

%%%%%%%%%%%%
In this \app, we present the proofs of all the lemmas used for proving Theorem~\ref{thm:mle_ub_lb}.

%%%%%%%%%%%%
\subsubsection{Proof of Lemma~\ref{lem:unique_solution_constrained}}\label{sec:proof_lem_unique_solution_constrained}

We fix any constant $\boundused>0$.

The \expandedmle~\eqref{eq:mle_constrained} is an optimization over the compact set $\rangeboxused$, and the negative log-likelihood function $\negloglikelihood$ is continuous. By the Extreme Value Theorem~\cite[Theorem 4.16]{rudin1976real}, a solution $\paramestbound{\boundused}$ is guaranteed to exist. 

It remains to prove the uniqueness of $\paramestbound{\boundused}$. Assume for contradiction that there exist two solutions $\paramest, \paramest'\in \rangeboxused$ to the \expandedmle~\eqref{eq:mle_constrained} and $\paramest \ne \paramest'$. By Lemma~\ref{lem:log_likelihood_convex}, the negative log-likelihood function $\negloglikelihood$ is strictly convex. Therefore,
\begin{align}\label{eq:ub_mle_uniqueness_convexity}
    \frac{1}{2}\left(\negloglikelihood(\paramest) + \negloglikelihood(\paramest')\right) > \negloglikelihood\left(\frac{\paramest + \paramest'}{2}\right).
\end{align}
It can be verified that $\frac{\paramest+\paramest'}{2}\in \rangeboxused$. Moreover,~\eqref{eq:ub_mle_uniqueness_convexity} along with the fact that $\negloglikelihood(\paramest) = \negloglikelihood(\paramest')$ implies that $\frac{\paramest + \paramest'}{2}$ attains a strictly smaller function value than both $\paramest$ and $\paramest'$. This contradicts the assumption that $\paramest$ and $\paramest'$ are both optimal solutions to the \expandedmle~\eqref{eq:mle_constrained}.

%%%%%%%%%%%%
\subsubsection{Proof of Lemma~\ref{lem:finite_solution}}\label{sec:proof_lem_finite_solution}

We first define a ``comparison graph'' $\graphcomparison(\{\numwins_{\idxitem\idxitemalt}\})$ as a function of the pairwise-comparison outcomes $\{\numwins_{\idxitem\idxitemalt}\}$. Let each item $\idxitem\in [\numitems]$ be a node of the graph. Let there be a directed edge  $(\idxitem\rightarrow \idxitemalt) \in \graphcomparison$, if and only if there exists a comparison where item $\idxitem$ beats item $\idxitemalt$. A directed graph is called strongly-connected if and only if there exists a path from every node $\idxitem$ to every other node $\idxitemalt$.

The following lemma from~\cite{ford1957exist} relates the existence and uniqueness of a finite unconstrained \mle $\paramestunconstrain$ to the strong connectivity of the comparison graph $\graphcomparison$. This lemma is based on a different parameterization of the BTL model. In this parameterization, each item has a weight $\weightgt_\idxitem > 0$, and the probability that item $\idxitem$ beats item $\idxitemalt$ equals $\frac{\weightgt_\idxitem}{\weightgt_\idxitem + \weightgt_\idxitemalt}$.
\begin{lemma}[Section 2 from~\cite{ford1957exist}]
If the comparison graph $\graphcomparison(\{\numwins_{\idxitem\idxitemalt}\})$ is strongly-connected, then there exists a unique solution to the following \mle:
\begin{align*}
    \weightestmle = \argmin_{\substack{\weight\in \reals^\numitems \\ \weight_\idxitem > 0,\; \sum_{\idxitem=1}^\numitems \weight_\idxitem = 1}} \negloglikelihoodreparam(\{\numwins_{\idxitem\idxitemalt}\}; \weight),
\end{align*}
where the negative log-likelihood function $\negloglikelihoodreparam$ is defined as \begin{align*}
    \negloglikelihoodreparam(\weight) = -\sum_{1\le \idxitem < \idxitemalt \le \numitems} \left(\numwins_{\idxitem\idxitemalt}\log\left(\frac{\weight_\idxitem}{\weight_\idxitem + \weight_\idxitemalt}\right) + \numwins_{\idxitemalt\idxitem}\log\left(\frac{\weight_\idxitemalt}{\weight_\idxitem+\weight_\idxitemalt}\right)\right).
\end{align*}
\end{lemma}

It can be seen that $\param$ and $\weight$ are simply different parameterizations of the same problem. There is a one-to-one mapping between $\param$ and $\weight$, by taking $\param_\idxitem=\log(\weight_\idxitem)$ and re-centering accordingly (or in the inverse direction, by taking $\weight_\idxitem=e^{\param_\idxitem}$ and normalizing accordingly). Therefore, the existence and the uniqueness of the \mle $\weightestmle$ in Lemma~\ref{lem:finite_solution} carries over to our unconstrained MLE $\paramestunconstrain$ in~\eqref{eq:unconstrained_mle}. That is, if the comparison graph $\graphcomparison$ is strongly-connected, then there exists a unique solution $\paramestunconstrain$ to the unconstrained MLE. It remains to show that the comparison graph $\graphcomparison$ 
is strongly-connected \whpone.~~\\

We first construct an undirected graph $\graphundirected(\{\numwins_{\idxitem
\idxitemalt}\})$ as follows. Let each item $\idxitem\in [\numitems]$ be a node of the graph $\graphundirected$. Let there be an undirected edge $(\idxitem, \idxitemalt)\in \graphcomparison'$, if and only if in the directed graph $\graphcomparison$ we have both $(\idxitem\rightarrow \idxitemalt) \in \graphcomparison$ and $(\idxitemalt\rightarrow \idxitem)\in \graphcomparison$. Equivalently, there exists an undirected edge $(\idxitem, \idxitemalt)\in \graphcomparison'$, if and only if $0 < \meanemp_{\idxitem\idxitemalt} < 1$. It can be verified that the connectivity of the undirected graph $\graphundirected$ implies the strong connectivity of the directed graph $\graphcomparison$. Therefore,
\begin{align}
    \Prob(\graphcomparison \text{ strongly-connected}) \ge \Prob(\graphundirected \text{ connected}).\label{eq:graph_directed_undirected}
\end{align}

The probability that $(\idxitem, \idxitemalt)\in \graphundirected$ is $\Prob(0 < \meanemp_{\idxitem\idxitemalt} < 1)$. By Hoeffding's inequality, we have that for any $t > 0$,
\begin{align*}
   \Prob(\abs{\meanemp_{\idxitem\idxitemalt} - \meangt_{\idxitem\idxitemalt}} > t) < 2e^{-\numcomparisons t^2}, \qquad \text{for all } 1 \le \idxitem < \idxitemalt \le \numitems.
\end{align*}
We have $0<\frac{1}{1 + e^{2\boundgt}} \le \meangt_{\idxitem\idxitemalt} \le \frac{1}{1 + e^{-2\boundgt}} < 1$, for any $\idxitem<\idxitemalt$. Since $\boundgt$ is a constant, we have that $\meangt_{\idxitem\idxitemalt}$ is bounded away from $0$ and $1$ by a constant. Set $t = \tau$ where $\tau$ is any constant such that $0 < \tau < \frac{1}{1 + e^{2\boundgt}}$. Then for all $1 \le \idxitem < \idxitemalt \le \numitems$, we have
\begin{align*}
     \Prob(0 < \meanemp_{\idxitem\idxitemalt} < 1) & > \Prob(\meangt_{\idxitem\idxitemalt} - \tau < \meanemp_{\idxitem\idxitemalt} < \meangt_{\idxitem\idxitemalt} + \tau)\\
     & \ge 1- \Prob(\abs{\meanemp_{\idxitem\idxitemalt} - \meangt_{\idxitem\idxitemalt}} > \tau) \\
     & >1- 2e^{-\const\numcomparisons },
\end{align*}
for some constant $\const>0$ .

Recall that the random variables $\{\meanemp_{\idxitem\idxitemalt}\}$ are independent across all $1 \le \idxitem < \idxitemalt \le \numitems$. Hence, the probability of the undirected graph $\graphundirected$ being connected is at least the probability of an (undirected) \Erdos-\Renyi \xspace random graph being connected, where each edge independently exists with probability $1-2e^{-\const\numcomparisons}$.

The following lemma from~\cite{gilbert1959graph} provides an upper bound on the probability of an  (undirected) \Erdos-\Renyi \xspace random graph being disconnected (and hence a lower bound on the probability of the graph being connected).

\begin{lemma}[Theorem 1 from~\cite{gilbert1959graph}]\label{lem:erdos_renyi_graph}
For an (undirected) \Erdos-\Renyi \xspace graph of $\numitems$ nodes, where each edge independently exists with probability $\probedge$. Let $\probedgecomp \defn 1 - \probedge$. Then the probability of the graph being disconnected is at most
\begin{align*}
    \left(1- \frac{\numitems-1}{2}\probedgecomp^{\numitems-1}\right)\numitems\probedgecomp^{\numitems-1} .
\end{align*}
\end{lemma}

To apply Lemma~\ref{lem:erdos_renyi_graph}, we set $\probedge=1-2e^{-\const\numcomparisons}$ and therefore $\probedgecomp = 2e^{-\const\numcomparisons}$. Then we have
\begin{align}
    \Prob[\graphundirected \text{ disconnected}] & \le \left(1- \frac{\numitems-1}{2}\probedgecomp^{\numitems-1}\right)\numitems\probedgecomp^{\numitems-1} \nonumber \\
    & \le \numitems \probedgecomp^{\numitems-1} \nonumber \\
    & =\numitems e^{-\const\numcomparisons(\numitems-1)} \nonumber \\
    &\le \frac{\const'}{\numitems\numcomparisons},\qquad \text{for some constant } \const' > 0.\label{eq:undirected_disconnected_bound}
\end{align}
Combining~\eqref{eq:graph_directed_undirected} and~\eqref{eq:undirected_disconnected_bound} completes the proof of the lemma.

%%%%%%%%%%%%
\subsubsection{Proof of Lemma~\ref{lem:concentration}}\label{sec:proof_lem_concentration}
We first consider any fixed $\idxitemthree\in [\numitems]$. By the definition of $\{\mean_{\idxitem\idxitemalt}\}$ in~\eqref{eq:notation_observed_fraction_of_wins}, we have \begin{align}
    \sum_{\idxitem\ne \idxitemthree}\meanemp_{\idxitemthree\idxitem} = \frac{1}{\numcomparisons } \sum_{\idxitem\ne \idxitemthree} \sum_{\idxcomparison=1}^\numcomparisons \resultcomparison_{\idxitemthree\idxitem}^{(\idxcomparison)}.\label{eq:ub_mean_sum_expression}
\end{align}
There are $(\numitems-1)\numcomparisons$ terms of the form $\resultcomparison_{\idxitemthree\idxitem}^{(\idxcomparison)}$ in~\eqref{eq:ub_mean_sum_expression}. It can be verified that the terms $\resultcomparison_{\idxitemthree\idxitem}^{(\idxcomparison)}$ involved in~\eqref{eq:ub_mean_sum_expression} are independent. Moreover, since $\resultcomparison_{\idxitemthree\idxitem}^{(\idxcomparison)}\in\{0, 1\}$, changing the value of a single term $\resultcomparison_{\idxitemthree\idxitem}^{(\idxcomparison)}$ changes the value of~\eqref{eq:ub_mean_sum_expression} by $\frac{1}{\numcomparisons}$. By McDiarmid's inequality, we have that for any $t > 0$,
\begin{align}
    \Prob\left[\abs*{\sum_{\idxitem\ne \idxitemthree}\meanemp_{\idxitemthree\idxitem} -\sum_{\idxitem\ne\idxitemthree}\meangt_{\idxitemthree\idxitem}} > t\right] \le 2\exp\left(-\frac{2t^2}{(\numitems-1)\numcomparisons\cdot (\frac{1}{\numcomparisons})^2}\right) = 2\exp\left(-\frac{2\numcomparisons t^2}{(\numitems-1)} \right).\label{eq:mcdiardmid_raw}
\end{align}

Setting $t = \const\sqrt{\frac{\numitems(\log\numitems + \log\numcomparisons)}{\numcomparisons}}$ in~\eqref{eq:mcdiardmid_raw}, we have
\begin{align}
    \Prob\left[\abs*{\sum_{\idxitem\ne \idxitemthree}\meanemp_{\idxitemthree\idxitem} -\sum_{\idxitem\ne\idxitemthree}\meangt_{\idxitemthree\idxitem}} \le \const\sqrt{\frac{\numitems(\log\numitems + \log\numcomparisons)}{\numcomparisons}}\right] & \ge 1 - 2\exp\left(-\const'\frac{\numitems}{\numitems-1}(\log\numitems + \log\numcomparisons)\right)\nonumber\\
    & \ge 1- \frac{\const''}{\numitems^{2} \numcomparisons},\label{eq:mcdiarmid}
\end{align}
for some constants $\const', \const'' > 0$, provided that the constant $\const>0$ is sufficiently large.

Taking a union bound over $\idxitemthree\in [\numitems]$ on~\eqref{eq:mcdiarmid} completes the proof.

%%%%%%
\subsubsection{Proof of Lemma~\ref{lem:bound_diff_each}}\label{sec:proof_lem_bound_diff_each}
Denote the random variables $\idxitemthreemax \defn  \argmax_{\idxitem\in [\numitems]} \diforacle_\idxitem$ and $\idxitemthreemin \defn \argmin_{\idxitem\in [\numitems]} \diforacle_\idxitem$. When there are multiple maximizers or minimizers, we arbitrarily choose one. 

Setting $\idxitemthree=\idxitemthreemax$ in the first-order optimality condition~\eqref{eq:first_order_optimality_recall}, we have
\begin{align}
    \underbrace{%
        \sum_{\idxitem=1}^\numitems \left[\funcsigmoid( \paramgt_\idxitemthreemax-\paramgt_\idxitem  +\diforacle_\idxitemthreemax-\diforacle_\idxitem) - \funcsigmoid(\paramgt_\idxitemthreemax-\paramgt_\idxitem)\right]
    }_{\termmax} = \sum_{\idxitem\ne \idxitemthreemax} (\meanemp_{\idxitemthree \idxitem} - \meangt_{\idxitemthree\idxitem}) \stackrel{\stepone}{\lessorder} \sqrt{\frac{\numitems(\log\numitems+\log\numcomparisons)}{\numcomparisons}},\label{eq:ub_term_plus_concentration}
\end{align}
where \stepone is true by Lemma~\ref{lem:concentration} (recall that the lemma statement is conditioned on the event $\eventlemmaub$ that both Lemma~\ref{lem:finite_solution} and Lemma~\ref{lem:concentration} hold).

Denote the function $\funcsigmoiddiff(\siganchor, \sigdif) \defn \funcsigmoid(\siganchor + \sigdif)- \funcsigmoid(\siganchor)  =\frac{1}{1 + e^{-(\siganchor+\sigdif)}} -\frac{1}{1 + e^{-\siganchor}}$. The following lemma states three properties for the function $\funcsigmoiddiff$, which are used in later parts of the proof.

\begin{lemma}\label{lem:sigmoid_diff_properties}
We have the following properties for the function $\funcsigmoiddiff$.
\begin{subequations}\label{eq:sigmoid_diff_properties}
\begin{align}
    \funcsigmoiddiff(\siganchor, \sigdif) & = -\funcsigmoiddiff(-\siganchor, -\sigdif),\qquad \text{for all } \siganchor, \sigdif \in \reals\label{eq:sigmoid_negate}\\
    \funcsigmoiddiff(\siganchor, \sigdif) & \ge \funcsigmoiddiff(\sigbound, \sigdif) > 0,\qquad \text{for all } \sigbound > 0,\text{ } \sigdif > 0, \text{ and all } \siganchor \text{ such that } -\sigbound\le \siganchor \le \sigbound \label{eq:sigmoid_diff_pos}\\
    \funcsigmoiddiff(\sigbound, \sigdif_1) + \funcsigmoiddiff(\sigbound, \sigdif_2) & \ge \funcsigmoiddiff(\sigbound, \sigdif_1 + \sigdif_2),\qquad \text{ for all } \sigbound > 0, \text{and all } \sigdif_1, \sigdif_2 \ge 0.\label{eq:sigmoid_diff_pos_neg}
\end{align}
\end{subequations}
\end{lemma}

Lemma~\ref{lem:sigmoid_diff_properties} can be verified by straightforward algebra. For completeness, we include the proof of Lemma~\ref{lem:sigmoid_diff_properties} at the end of this \app.

By the definition of $\idxitemthreemax$, we have $\diforacle_\idxitemthreemax = \max_{\idxitem\in [\numitems]} \diforacle_\idxitem$, and therefore $\diforacle_\idxitemthreemax - \diforacle_\idxitem \ge 0$ for all $\idxitem\in [\numitems]$. Hence, we have
\begin{align}
  \termmax & = \sum_{\idxitem=1}^\numitems \funcsigmoid(\paramgt_\idxitemthreemax-\paramgt_\idxitem+\diforacle_\idxitemthreemax- \diforacle_\idxitem) - \funcsigmoid(\paramgt_\idxitemthreemax-\paramgt_\idxitem) \nonumber\\
  & =\sum_{\idxitem=1}^\numitems \funcsigmoiddiff(\paramgt_\idxitemthreemax-\paramgt_\idxitem, \diforacle_\idxitemthreemax-\diforacle_\idxitem) \nonumber\\
 & \stackrel{\stepone}{\ge} \sum_{\idxitem=1}^\numitems \funcsigmoiddiff(2\boundgt, \diforacle_\idxitemthreemax - \diforacle_\idxitem),\label{eq:ub_term_plus}
\end{align}
where \stepone is true by~\eqref{eq:sigmoid_diff_pos} combined with the fact that $\abs{\paramgt_{\idxitem}-\paramgt_\idxitemalt} \le \abs{\paramgt_{\idxitem}} + \abs{\paramgt_\idxitemalt} \le 2\boundgt$ for all $\idxitem,\idxitemalt\in [\numitems]$. 

Similarly, setting $\idxitemthree=\idxitemthreemin$ in the first-order optimality condition~\eqref{eq:first_order_optimality_recall}, we have
\begin{align}
    \underbrace{%
       \sum_{\idxitem=1}^\numitems \left[\funcsigmoid( \paramgt_\idxitemthreemin-\paramgt_\idxitem  +\diforacle_\idxitemthreemin-\diforacle_\idxitem) - \funcsigmoid(\paramgt_\idxitemthreemin - \paramgt_\idxitem)\right]
    }_{\termmin} \lessorder \sqrt{\frac{\numitems(\log\numitems + \log\numcomparisons)}{\numcomparisons}}.\label{eq:ub_term_minus_concentration}
\end{align}
By the definition of $\idxitemthreemin$, we have $\diforacle_\idxitemthreemin = \min_{\idxitem\in [\numitems]} \diforacle_\idxitem$, and therefore $\diforacle_\idxitem - \diforacle_\idxitemthreemin \ge 0$ for all $\idxitem\in [\numitems]$. Hence, we have
\begin{align}
     \termmin & = \sum_{\idxitem=1}^\numitems \funcsigmoid(\paramgt_\idxitemthreemin-\paramgt_\idxitem +\diforacle_\idxitemthreemin- \diforacle_\idxitem) - \funcsigmoid(\paramgt_\idxitemthreemin-\paramgt_\idxitem) \nonumber\\
     & =\sum_{\idxitem=1}^\numitems \funcsigmoiddiff(\paramgt_\idxitemthreemin-\paramgt_\idxitem, \diforacle_\idxitemthreemin-\diforacle_\idxitem) \nonumber\\
     & \stackrel{\stepone}{=}\sum_{\idxitem=1}^\numitems -\funcsigmoiddiff(\paramgt_\idxitem- \paramgt_\idxitemthreemin, \diforacle_\idxitem-\diforacle_\idxitemthreemin) \nonumber\\
     & \stackrel{\steptwo}{\le} \sum_{\idxitem=1}^\numitems -\funcsigmoiddiff(2\boundgt, \diforacle_\idxitem - \diforacle_\idxitemthreemin),\label{eq:ub_term_minus}
\end{align}
where \stepone is true by~\eqref{eq:sigmoid_negate}, and \steptwo is true by~\eqref{eq:sigmoid_diff_pos} combined with the fact that $\abs{\paramgt_\idxitem - \paramgt_\idxitemalt}\le 2\boundgt$ for all $\idxitem,\idxitemalt\in [\numitems]$.

Combining~\eqref{eq:ub_term_plus} and~\eqref{eq:ub_term_minus}, we have
\begin{align}
    \termmax - \termmin & \ge \sum_{\idxitem=1}^\numitems \funcsigmoiddiff(2\boundgt, \diforacle_\idxitemthreemax - \diforacle_\idxitem) +  \sum_{\idxitem=1}^\numitems \funcsigmoiddiff(2\boundgt, \diforacle_\idxitem - \diforacle_\idxitemthreemin) \nonumber\\
   & \stackrel{\stepone}{\ge} \sum_{\idxitem=1}^\numitems\funcsigmoiddiff(2\boundgt, \diforacle_\idxitemthreemax - \diforacle_\idxitemthreemin) \nonumber\\
   & = \numitems\cdot \funcsigmoiddiff(2\boundgt, \diforacle_\idxitemthreemax - \diforacle_\idxitemthreemin) \stackrel{\steptwo}{ \ge} 0 ,\label{eq:ub_term_plus_minus_expression}
\end{align}
where \stepone is true due to~\eqref{eq:sigmoid_diff_pos_neg} since $\diforacle_{\idxitemthreemax} -\diforacle_\idxitem  \ge 0$ and $\diforacle_\idxitem -\diforacle_{\idxitemthreemin}  \ge 0$ for all $\idxitem\in [\numitems]$, and \steptwo is true since $\diforacle_\idxitemthreemax - \diforacle_\idxitemthreemin \ge 0$. On the other hand, combining~\eqref{eq:ub_term_plus_concentration} and~\eqref{eq:ub_term_minus_concentration}, we have
\begin{align}
    \termmax - \termmin \lessorder \sqrt{\frac{\numitems(\log\numitems +\log\numcomparisons)}{\numcomparisons}}.\label{eq:ub_term_plus_minus_concentration}
\end{align}
Combining~\eqref{eq:ub_term_plus_minus_expression} and~\eqref{eq:ub_term_plus_minus_concentration}, we have 
\begin{align}
    0 \le \numitems\cdot \funcsigmoiddiff(2\boundgt, \diforacle_\idxitemthreemax - \diforacle_\idxitemthreemin) \le \termmax - \termmax & \lessorder\sqrt{\frac{\numitems(\log\numitems +\log\numcomparisons)}{\numcomparisons}}\nonumber\\
    \funcsigmoiddiff(2\boundgt, \diforacle_\idxitemthreemax - \diforacle_\idxitemthreemin) & \lessorder\sqrt{\frac{\log\numitems +\log\numcomparisons}{\numitems\numcomparisons}}\nonumber\\
    \funcsigmoid(2\boundgt + \diforacle_\idxitemthreemax - \diforacle_\idxitemthreemin) - \funcsigmoid(2\boundgt) & \lessorder \sqrt{\frac{\log\numitems +\log\numcomparisons}{\numitems\numcomparisons}}.\label{eq:ub_diforacle_concentration}
\end{align}
By the first-order mean value theorem on the LHS of~\eqref{eq:ub_diforacle_concentration}, we have
\begin{align}
 \funcsigmoid(2\boundgt + \diforacle_\idxitemthreemax - \diforacle_\idxitemthreemin) - \funcsigmoid(2\boundgt) = \funcsigmoid'(\meanvalue)\cdot (\diforacle_\idxitemthreemax - \diforacle_\idxitemthreemin)   \le \const\sqrt{\frac{\log\numitems+\log\numcomparisons}{\numitems\numcomparisons}},\label{eq:ub_mvt_order_one}
 \end{align}
 where $\meanvalue$ is a random variable that takes values in the interval $[2\boundgt, 2\boundgt+\diforacle_\idxitemthreemax - \diforacle_\idxitemthreemin]$.

Let $\epsilon$ be any constant such that $0 < \epsilon < 1-\funcsigmoid(2\boundgt)$. Then there exists a constant $\tau>0$ such that $\funcsigmoid(2\boundgt + \tau) - \funcsigmoid(2\boundgt) = \epsilon$. On the other hand, there exist constants $\constitems >0$ and $\constcomparisons>0$ such that 
\begin{align}
   \const\sqrt{\frac{\log\numitems+\log\numcomparisons}{\numitems\numcomparisons}} < \epsilon, \qquad \text{for any }\numitems \ge \constitems \text{ and } \numcomparisons\ge \constcomparisons.\label{eq:ub_sufficiently_large_constant}
\end{align}
Combining~\eqref{eq:ub_mvt_order_one} and~\eqref{eq:ub_sufficiently_large_constant}, we have
\begin{align}
    \funcsigmoid(2\boundgt + \diforacle_\idxitemthreemax - \diforacle_\idxitemthreemin) - \funcsigmoid(2\boundgt) & \le \const\sqrt{\frac{\log\numitems+\log\numcomparisons}{\numitems\numcomparisons}} < \epsilon = \funcsigmoid(2\boundgt+\tau) - \funcsigmoid(2\boundgt) \nonumber\\
    \funcsigmoid(2\boundgt + \diforacle_\idxitemthreemax - \diforacle_\idxitemthreemin) & \le \funcsigmoid(2\boundgt+\tau).\label{eq:bound_sigmoid_max_min_dif}
\end{align}
By~\eqref{eq:property_first_derivative_bounded}, we have $\funcsigmoid' > 0$ on $(-\infty, \infty)$, and hence the function $\funcsigmoid$ is monotonically increasing. Hence, from~\eqref{eq:bound_sigmoid_max_min_dif}, we have $\diforacle_\idxitemthreemax - \diforacle_\idxitemthreemin \le \tau$, and therefore the interval $[2\boundgt, 2\boundgt + \diforacle_\idxitemthreemax - \diforacle_\idxitemthreemin]$ is bounded. By the property~\eqref{eq:property_first_derivative_bounded} of the sigmoid function $\funcsigmoid$, we have $\funcsigmoid' > \const_3 > 0$ for some constant $\const_3 > 0$ in the bounded interval $[2\boundgt, 2\boundgt + \diforacle_\idxitemthreemax - \diforacle_\idxitemthreemin]$. Recall that $\meanvalue$ takes values in the interval $[2\boundgt, 2\boundgt+\diforacle_\idxitemthreemax-\diforacle_\idxitemthreemin]$. Therefore, we have
\begin{align}
    \const_3 ( \diforacle_\idxitemthreemax - \diforacle_\idxitemthreemin)< \funcsigmoid'(\meanvalue)\cdot (\diforacle_\idxitemthreemax - \diforacle_\idxitemthreemin) .\label{eq:ub_bounded_first_derivative}
\end{align}
Combining~\eqref{eq:ub_mvt_order_one} and~\eqref{eq:ub_bounded_first_derivative}, we have 
\begin{align}
   \const_3 (\diforacle_\idxitemthreemax - \diforacle_\idxitemthreemin) & < \funcsigmoid'(\meanvalue)\cdot (\diforacle_\idxitemthreemax - \diforacle_\idxitemthreemin)  \le \const\sqrt{\frac{\log\numitems+\log\numcomparisons}{\numitems\numcomparisons}}\nonumber\\
   \diforacle_\idxitemthreemax - \diforacle_\idxitemthreemin & \lessorder \sqrt{\frac{\log\numitems+\log\numcomparisons}{\numitems\numcomparisons}}.\label{eq:ub_diff_bound_max_min}
\end{align}

By the assumption that $\paramgt\in \rangeboxgt$, we have $\sum_{\idxitem=1}^\numitems \paramgt_\idxitem= 0$. Similarly, by the centering constraint on the unconstrained MLE $\paramestunconstrain$ in~\eqref{eq:unconstrained_mle}, we have $\sum_{\idxitem=1}^\numitems \paramestunconstrain_\idxitem=0$. Hence, we have the deterministic relation
\begin{align}\label{eq:diforacle_sum_zero}
    \sum_{\idxitem=1}^\numitems \paramestunconstrain_\idxitem - \sum_{\idxitem=1}^\numitems \paramgt_\idxitem = \sum_{\idxitem=1}^\numitems\diforacle_\idxitem=0.
\end{align}
Hence, $\diforacle_\idxitemthreemax \ge 0$ and $\diforacle_\idxitemthreemin \le 0$. By~\eqref{eq:ub_diff_bound_max_min}, we have
\begin{align*}
    \diforacle_\idxitemthreemax - \diforacle_\idxitemthreemin =\abs{\diforacle_\idxitemthreemax } +\abs{\diforacle_\idxitemthreemin}\lessorder \sqrt{\frac{\log\numitems+\log\numcomparisons}{\numitems\numcomparisons}}.
\end{align*}
Hence, $\abs{\diforacle_\idxitemthreemax} \lessorder\sqrt{\frac{\log\numitems+\log\numcomparisons}{\numitems\numcomparisons}}$ and $\abs{\diforacle_\idxitemthreemin} \lessorder \sqrt{\frac{\log\numitems+\log\numcomparisons}{\numitems\numcomparisons}}$. Therefore,
\begin{align*}
    \abs{\diforacle_\idxitemthree} \lessorder \sqrt{\frac{\log\numitems+\log\numcomparisons}{\numitems\numcomparisons}}, \qquad\text{for all }\idxitemthree\in [\numitems],
\end{align*}
completing the proof of the lemma.~~\\

%%%%%%
\paragraph{Proof of Lemma~\ref{lem:sigmoid_diff_properties}:}

We prove the three parts of the lemma separately.
\begin{enumeratex}[(a)]
    %%%%%%%%%%%%
    \item It can be verified that $\funcsigmoid(x) = 1 - \funcsigmoid(-x)$. Hence,
    \begin{align*}
        \funcsigmoiddiff(\siganchor, \sigdif) = \funcsigmoid(\siganchor + \sigdif) - \funcsigmoid(\siganchor) & = [1 - \funcsigmoid(-\siganchor - \sigdif)] - [1 - \funcsigmoid(-\siganchor)]\\
        & = -[\funcsigmoid(-\siganchor - \sigdif) - \funcsigmoid(-\siganchor)] = -\funcsigmoiddiff(-\siganchor, -\sigdif).
    \end{align*}
    \item We prove the two parts of the inequality separately.
    
    We first prove that $\funcsigmoiddiff(\sigbound, \sigdif) > 0$. By~\eqref{eq:property_first_derivative_bounded}, the function $\funcsigmoid$ is strictly increasing. Therefore, for any $\sigdif > 0$, we have
    \begin{align*}
        \funcsigmoiddiff(\sigbound, \sigdif) = \funcsigmoid(\sigbound +\sigdif)-\funcsigmoid(\sigbound)  > 0.
    \end{align*}
    Now we prove that $\funcsigmoiddiff(\siganchor, \sigdif) \ge \funcsigmoiddiff(\sigbound, \sigdif)$. We have
    \begin{align}
        \funcsigmoiddiff(\siganchor, \sigdif) - \funcsigmoiddiff(\sigbound, \sigdif) & = \funcsigmoid(\siganchor + \sigdif)-\funcsigmoid(\siganchor) - [\funcsigmoid(\sigbound + \sigdif)-\funcsigmoid(\sigbound)] \nonumber\\
        & = \int_{\siganchor}^{\siganchor+\sigdif} \funcsigmoid'(u) \di u - \int_{\sigbound}^{\sigbound+\sigdif} \funcsigmoid'(u)\di u \nonumber\\
        & = \int_{0}^{\sigdif} \funcsigmoid'(\siganchor+u) \di u - \int_{0}^{\sigdif} \funcsigmoid'(\sigbound+ u)\di u\nonumber \\
        & = \int_{0}^\sigdif [\funcsigmoid'(\siganchor+u)-\funcsigmoid'(\sigbound +u)] \di u.\label{eq:sigmoid_diff_integral}
    \end{align}
    By~\eqref{eq:sigmoid_diff_integral}, it remains to prove that
    \begin{align}
         \funcsigmoid'(\siganchor +u)  \ge \funcsigmoid'(\sigbound + u),\qquad \text{for any } u\in [0, t].\label{eq:sigmoid_property_desired}
    \end{align}
    Fix any $u\in [0, t]$. By assumption we have $\sigbound > 0$. Hence, $\sigbound + u > 0$. Now we consider the sign of $(\siganchor+u)$.
    
    If $\sigvar + u \ge 0$, then by the assumption that $\siganchor\le \sigbound$, we have $0 \le \sigvar + u \le \sigbound + u$. It can be verified that $\funcsigmoid'$ is decreasing on $[0, \infty)$. Therefore,
    \begin{align}
        \funcsigmoid'(\siganchor + u) \ge \funcsigmoid'(\sigbound + u).\label{eq:sigmoid_diff_integrand_one}
    \end{align}
    
    If $\siganchor + u < 0$, we have 
    \begin{align}
        0 < -\siganchor -u \stackrel{\stepone}{\le} \sigbound - u \stackrel{\steptwo}{\le} \sigbound + u,\label{eq:sigmoid_diff_argument_two}
    \end{align}
    where \stepone is true by the assumption that $\siganchor \ge -\sigbound$, and \steptwo is true because $u\in [0, t]$ and therefore $u \ge 0$. We have
    \begin{align}
         \funcsigmoid'(\siganchor+u) \stackrel{\stepone}{=}  \funcsigmoid'(-\siganchor-u)\stackrel{\steptwo}{\ge} \funcsigmoid'(\sigbound +u),\label{eq:sigmoid_diff_integrand_two}
    \end{align}
    where \stepone holds because it can be verified that $\funcsigmoid'(x) = \funcsigmoid'(-x)$ for any $x\in \reals$, and \steptwo is true by combining~\eqref{eq:sigmoid_diff_argument_two} with the fact that $\funcsigmoid'$ is decreasing on $[0, \infty)$.
    
    Combining the two cases of~\eqref{eq:sigmoid_diff_integrand_one} and~\eqref{eq:sigmoid_diff_integrand_two} completes the proof of~\eqref{eq:sigmoid_property_desired}.

    \item We have \begin{align*}
        \funcsigmoiddiff(\sigbound, \sigdif_1) + \funcsigmoiddiff(\sigbound, \sigdif_2) & =  \funcsigmoid(\sigbound + \sigdif_1) -\funcsigmoid(\sigbound) + \funcsigmoid(\sigbound + \sigdif_2)-\funcsigmoid(\sigbound)\\
        & = \int_{\sigbound}^{\sigbound + \sigdif_1} \funcsigmoid'(u) \di u + \int_{\sigbound}^{\sigbound + \sigdif_2} \funcsigmoid'(u) \di u\\
        & \stackrel{\stepone}{\ge } \int_{\sigbound}^{\sigbound + \sigdif_1} \funcsigmoid'(u) \di u + \int_{\sigbound+\sigdif_1}^{\sigbound + \sigdif_1 +\sigdif_2} \funcsigmoid'(u) \di u\\
        & = \int_{\sigbound}^{\sigbound + \sigdif_1 + \sigdif_2} \funcsigmoid'(u) \di u \\
        & = \funcsigmoid(\sigbound + \sigdif_1 + \sigdif_2) - \funcsigmoid(\sigbound) = \funcsigmoiddiff(\sigbound, \sigdif_1 + \sigdif_2),
    \end{align*}
    where \stepone is true because $\funcsigmoid'$ is decreasing on $(0, \infty)$, and because $\tau > 0$ and $\sigdif_1, \sigdif_2 \ge 0$ by assumption.
\end{enumeratex}

%%%%%%
\subsubsection{Proof of Lemma~\ref{lem:ub_conditional_expectation}}\label{sec:proof_lem_ub_conditional_expectation}
We fix any $\idxitem, \idxitemalt\in[\numitems]$ where $\idxitem\ne \idxitemalt$. By the law of iterated expectation, we have
\begin{align}
    \Expect[\meanemp_{\idxitem\idxitemalt}\given \event] & =  \Expect[\meanemp_{\idxitem\idxitemalt}\given \event', \event] \cdot \Prob(\event'\given \event) + \Expect[\meanemp_{\idxitem\idxitemalt}\given \eventcomplement', \event] \cdot \Prob(\eventcomplement'\given \event).\label{eq:ub_mean_iterated_expectation}
\end{align}
Subtracting $\Expect[\meanemp_{\idxitem\idxitemalt}\given \event', \event]$ from both sides of~\eqref{eq:ub_mean_iterated_expectation}, we have
\begin{align}
    \Expect[\meanemp_{\idxitem\idxitemalt}\given \event]  -  \Expect[\meanemp_{\idxitem\idxitemalt}\given \event', \event] &= \Expect[\meanemp_{\idxitem\idxitemalt}\given \event', \event] \cdot [\Prob(\event'\given \event)-1] + \Expect[\meanemp_{\idxitem\idxitemalt}\given \eventcomplement', \event] \cdot \Prob(\eventcomplement'\given \event) \nonumber\\
    & = (-\Expect[\meanemp_{\idxitem\idxitemalt} \given \event', \event] + \Expect[\meanemp_{\idxitem\idxitemalt} \given \eventcomplement', \event])\cdot \Prob(\eventcomplement' \given \event).\label{eq:ub_mean_iterated_expectation_subtract}
\end{align}
Taking an absolute value on~\eqref{eq:ub_mean_iterated_expectation_subtract}, we have
\begin{align*}
    \abs*{\Expect[\meanemp_{\idxitem\idxitemalt}\given \event] -  \Expect[\meanemp_{\idxitem\idxitemalt}\given \event', \event]} &  =
      \abs*{-\Expect[\meanemp_{\idxitem\idxitemalt} \given \event', \event] + \Expect[\meanemp_{\idxitem\idxitemalt} \given \eventcomplement', \event]}\cdot \Prob(\eventcomplement' \given \event)\\
    & \stackrel{\stepone}{\lessorder} \frac{1}{\numitems\numcomparisons},
\end{align*}
    where \stepone is true due to the deterministic inequality $0\le \meanemp_{\idxitem\idxitemalt} \le 1$ and the fact that event $\event'$ happens \whponecondition{\event}.
     
%%%%%%
\subsubsection{Proof of Lemma~\ref{lem:ub_expectation_bound_each}}\label{sec:proof_lem_ub_expectation_bound_each}
Denote $\idxitemthreemax \defn \argmax_{\idxitem\in [\numitems]} \difexpect_\idxitem$ and $\idxitemthreemin \defn \argmin_{\idxitem\in [\numitems]} \difexpect_\idxitem$. When there are multiple maximizers or minimizers, we arbitrarily choose one. The proof works similarly in spirit to the proof of Lemma~\ref{lem:bound_diff_each}. We first show that $\difexpect_\idxitemthreemax - \difexpect_\idxitemthreemin$ satisfies the desired upper bound. Then we show that $\difexpect_\idxitemthreemax$ and $\difexpect_\idxitemthreemin$ have different signs, and therefore the desired upper bound holds on $\abs{\difexpect_\idxitemthree}$ uniformly across all $\idxitemthree\in [\numitems]$.

Recall from~\eqref{eq:ub_mvt_order_two_expectation} that for every $\idxitemthree\in [\numitems]$,
\begin{align}
     \sum_{\idxitem=1}^\numitems \funcsigmoid'(\paramgt_\idxitemthree-\paramgt_\idxitem) \cdot (\difexpect_\idxitemthree-\difexpect_\idxitem) & = \underbrace{%
        \sum_{\idxitem\ne \idxitemthree}(\Expect[\meanemp_{\idxitemthree\idxitem}\given \eventlemmaub]-\meangt_{\idxitemthree\idxitem})
     }_{\term_1} -  \underbrace{%
        \frac{1}{2}\sum_{\idxitem=1}^\numitems\Expect[\funcsigmoid''(\meanvalue_{\idxitemthree\idxitem})(\diforacle_\idxitemthree-\diforacle_\idxitem)^2
        \given \eventlemmaub]
    }_{\term_2},\label{eq:ub_mvt_order_two_expectation_recall}
\end{align}
where $\meanvalue_{\idxitemthree\idxitem}$ is a random variable that takes values between $\paramgt_\idxitemthree-\paramgt_\idxitem$ and $\paramgt_\idxitemthree-\paramgt_\idxitem + (\diforacle_\idxitemthree-\diforacle_\idxitem)$.

We consider the two terms on the RHS of~\eqref{eq:ub_mvt_order_two_expectation} separately. For the term $\term_1$, recall from~\eqref{eq:ub_empirical_prob_expectation_bound_apply_lemma} that
\begin{align*}
        \abs*{\Expect[\meanemp_{\idxitemthree\idxitem} \given \eventlemmaub] - \meangt_{\idxitemthree\idxitem}} \lessorder \frac{1}{\numitems\numcomparisons}.
\end{align*}
Therefore,
\begin{align}
    \abs{\term_1} \lessorder (\numitems-1)\cdot \frac{1}{\numitems\numcomparisons} \lessorder\frac{1}{\numcomparisons}.\label{eq:ub_mvt_rhs_mean_diff}
\end{align}

Now consider the term $\term_2$. Recall that $\paramgt\in \rangeboxgt$. Therefore, for every $\idxitemthree\in [\numitems]$, we have $\abs{\paramgt_\idxitemthree} \le \boundgt$. Recall from Lemma~\ref{lem:bound_diff_each} that for every $\idxitemthree\in [\numitems]$, we have
\begin{align}
    \abs{\diforacle_\idxitemthree} \lessorder\sqrt{\frac{\log\numitems+\log\numcomparisons}{\numitems\numcomparisons}},\qquad \text{conditioned on $\eventlemmaub$}. \label{eq:ub_oracle_diff_bound_recall}
\end{align}
Let $\const> 0$ be any constant. By~\eqref{eq:ub_oracle_diff_bound_recall}, we have $\abs{\diforacle_\idxitemthree}\le \const$, for all $\numitems\ge \constitems$ and $\numcomparisons\ge \constcomparisons$, where $\constitems$ and $\constcomparisons$ are constants which may only depend on $\const$. Hence, conditioned on $\eventlemmaub$, the interval between $\paramgt_\idxitemthree-\paramgt_\idxitem$ and $\paramgt_\idxitemthree -\paramgt_\idxitem + (\diforacle_\idxitemthree - \diforacle_\idxitem)$ is contained in the interval $[-2\boundgt-2\const, 2\boundgt + 2\const]$. By the property~\eqref{eq:property_second_derivative_bounded} of the sigmoid function $\funcsigmoid$, we have
\begin{align*}
    \abs{\funcsigmoid''} < \const_5, \qquad \text{on the bounded interval } [-2\boundgt-2\const, 2\boundgt + 2\const].
\end{align*}
Therefore,
\begin{align*}
    \abs*{\Expect\left[ \funcsigmoid''(\meanvalue_{\idxitemthree\idxitem})\cdot (\diforacle_\idxitemthree-\diforacle_\idxitem)^2\given \eventlemmaub\right]} \le \const_5 \cdot \Expect[(\diforacle_\idxitemthree-\diforacle_\idxitem)^2\given \eventlemmaub] \stackrel{\stepone}{\lessorder} \frac{\log\numitems+\log\numcomparisons}{\numitems\numcomparisons}, \qquad \text{for all } \idxitem, \idxitemthree\in [\numitems],
\end{align*}
where \stepone is again by~\eqref{eq:ub_oracle_diff_bound_recall}. Therefore,
\begin{align}
    \abs{\term_2} \lessorder \numitems \cdot \frac{\log\numitems+\log\numcomparisons}{\numitems\numcomparisons} = \frac{\log\numitems+\log\numcomparisons}{\numcomparisons}.\label{eq:ub_mvt_rhs_second_order}
\end{align}
    
Taking an absolute value on~\eqref{eq:ub_mvt_order_two_expectation_recall} and using the triangle inequality, we have
\begin{align}
    \abs*{\sum_{\idxitem=1}^\numitems \funcsigmoid'(\paramgt_\idxitemthree-\paramgt_\idxitem) \cdot (\difexpect_\idxitemthree-\difexpect_\idxitem)} &\le \abs{\term_1} + \abs{\term_2} \stackrel{\stepone}{\lessorder} \frac{\log\numitems+\log\numcomparisons}{\numcomparisons},\label{eq:ub_first_order_expectation_bound}
\end{align}
where \stepone is true by combining the term $\term_1$ from~\eqref{eq:ub_mvt_rhs_mean_diff} and the term $\term_2$ from~\eqref{eq:ub_mvt_rhs_second_order}. Taking $\idxitemthree = \idxitemthreemax$ in~\eqref{eq:ub_first_order_expectation_bound}, we have
\begin{align}
    \sum_{\idxitem=1}^\numitems \funcsigmoid'(\paramgt_\idxitemthreemax-\paramgt_\idxitem) \cdot (\difexpect_\idxitemthreemax-\difexpect_\idxitem) & \le \const\frac{\log\numitems+\log\numcomparisons}{\numcomparisons}.\label{eq:ub_first_order_expectation_bound_max}
\end{align}
Taking $\idxitemthree = \idxitemthreemin$ in~\eqref{eq:ub_first_order_expectation_bound}, we have
\begin{align*}
    \sum_{\idxitem=1}^\numitems \funcsigmoid'(\paramgt_\idxitemthreemin-\paramgt_\idxitem) \cdot (\difexpect_\idxitemthreemin-\difexpect_\idxitem) & \ge -\const\frac{\log\numitems+\log\numcomparisons}{\numcomparisons}
\end{align*}
and hence
\begin{align}
    \sum_{\idxitem=1}^\numitems \funcsigmoid'(\paramgt_\idxitemthreemin-\paramgt_\idxitem) \cdot (\difexpect_\idxitem-\difexpect_\idxitemthreemin) & \le \const\frac{\log\numitems+\log\numcomparisons}{\numcomparisons}.\label{eq:ub_first_order_expectation_bound_min}
\end{align} 
Adding~\eqref{eq:ub_first_order_expectation_bound_max} and~\eqref{eq:ub_first_order_expectation_bound_min}, we have
\begin{align}
    \underbrace{
        \sum_{\idxitem=1}^\numitems \funcsigmoid'(\paramgt_\idxitemthreemax-\paramgt_\idxitem) \cdot (\difexpect_\idxitemthreemax - \difexpect_\idxitem) + \sum_{\idxitem=1}^\numitems \funcsigmoid'(\paramgt_\idxitemthreemin-\paramgt_\idxitem)\cdot(\difexpect_\idxitem-\difexpect_\idxitemthreemin)
    }_{\term}
    & \le \const\frac{\log\numitems+\log\numcomparisons}{\numcomparisons}.\label{eq:ub_first_order_bound_expectation_max_min}
\end{align}

Consider the term $\term$. We have $\abs{\paramgt_\idxitemthree-\paramgt_\idxitem} \le 2\boundgt$ for all $\idxitem, \idxitemthree\in [\numitems]$. By the property~\eqref{eq:property_first_derivative_bounded} of the sigmoid function, there exists some constant $\const_3$, such that
\begin{align}
    \funcsigmoid'(\paramgt_\idxitemthree - \paramgt_\idxitem) > \const_3 > 0, \qquad \text{for all } \idxitem, \idxitemthree\in [\numitems].\label{eq:ub_first_order_expectation_bounded_first_derivative}
\end{align}
By the definition of $\idxitemthreemax$ and $\idxitemthreemin$, we have $\difexpect_\idxitemthreemax - \difexpect_\idxitem \ge 0$ and $\difexpect_\idxitem - \difexpect_\idxitemthreemin \ge 0$ for every $\idxitem\in [\numitems]$. Plugging~\eqref{eq:ub_first_order_expectation_bounded_first_derivative} into~\eqref{eq:ub_first_order_bound_expectation_max_min}, combined with the fact that $\difexpect_\idxitemthreemax - \difexpect_\idxitem \ge 0$ and $\difexpect_\idxitem- \difexpect_\idxitemthreemin \ge 0$, we have
\begin{align}
    \const_3 \left[\sum_{\idxitem=1}^\numitems(\difexpect_\idxitemthreemax - \difexpect_\idxitem) + \sum_{\idxitem=1}^\numitems (\difexpect_\idxitem - \difexpect_\idxitemthreemin)\right] & \le \term \le     \const\frac{\log\numitems+\log\numcomparisons}{\numcomparisons} \nonumber\\
    \const_3 \numitems \cdot (\difexpect_\idxitemthreemax- \difexpect_\idxitemthreemin) & \le     \const\frac{\log\numitems+\log\numcomparisons}{\numcomparisons} \nonumber\\
    \difexpect_\idxitemthreemax - \difexpect_\idxitemthreemin & \lessorder \frac{\log\numitems + \log\numcomparisons}{\numitems\numcomparisons}.\label{eq:ub_diforacle_difference_bound}
\end{align}

By~\eqref{eq:diforacle_sum_zero} in the proof of Lemma~\ref{lem:bound_diff_each}, we have the deterministic relation
\begin{align}
    \sum_{\idxitem=1}^\numitems \diforacle_\idxitem=0.\label{eq:diforacle_sum_zero_recall}
\end{align}
Taking an expectation over~\eqref{eq:diforacle_sum_zero_recall} conditional on $\eventlemmaub$, we have
\begin{align*}
    \sum_{\idxitem=1}^\numitems \difexpect_\idxitem = 0.
\end{align*}
Hence, $\difexpect_\idxitemthreemax\ge 0$ and $\difexpect_\idxitemthreemin \le 0$. By~\eqref{eq:ub_diforacle_difference_bound}, we have
\begin{align*}
    \difexpect_\idxitemthreemax - \difexpect_\idxitemthreemin = \abs{\difexpect_\idxitemthreemax} + \abs{\difexpect_\idxitemthreemin} \lessorder \frac{\log\numitems + \log\numcomparisons}{\numitems\numcomparisons}.
\end{align*}
Hence, $\abs{\difexpect_\idxitemthreemax} \lessorder\frac{\log\numitems + \log\numcomparisons}{\numitems\numcomparisons}$ and $\abs{\difexpect_\idxitemthreemin} \lessorder\frac{\log\numitems + \log\numcomparisons}{\numitems\numcomparisons}$. Therefore,
\begin{align*}
    \abs{\difexpect_\idxitemthree} \lessorder \frac{\log\numitems + \log\numcomparisons}{\numitems\numcomparisons}, \qquad \text{for all } \idxitemthree\in [\numitems].
\end{align*}

%%%%%%
\subsubsection{Proof of Lemma~\ref{lem:toy_lb_surrogate_greater}}\label{sec:proof_lem_toy_lb_surrogate_greater}
\begin{figure}
    \centering
    \includegraphics[width=0.45\linewidth]{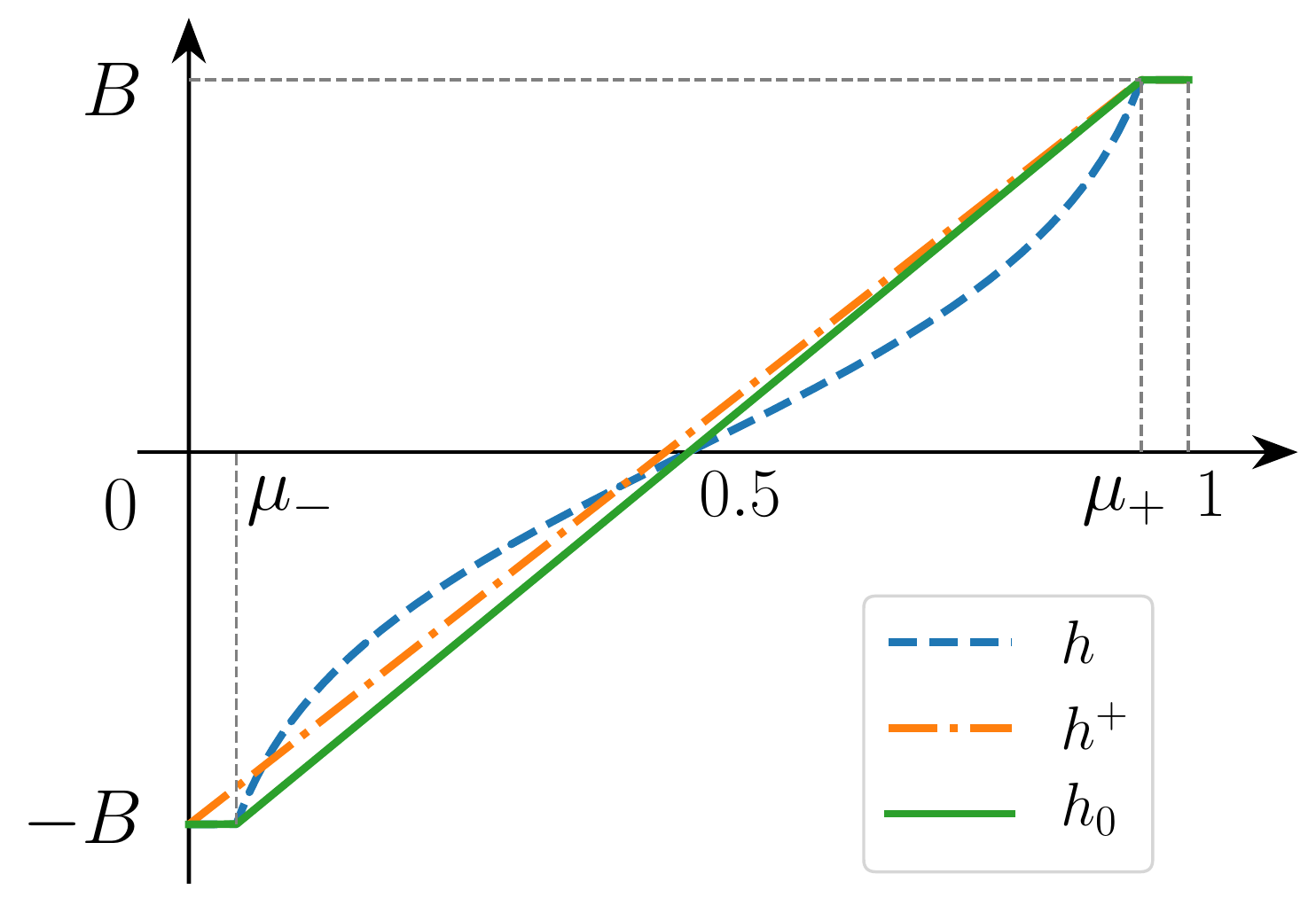}
    \caption{The functions $\funclog, \funclogplus$ and $\funcloglinear$.}
    \label{fig:func_log_surrogate}
\end{figure}

To compare the functions $\funclog$ and $\funclogplus$, we introduce an auxiliary function $\funcloglinear : [0, 1] \rightarrow [-\boundgt, \boundgt]$:
\begin{align*}
    \funcloglinear(\varmean) = \begin{cases}
        -\boundgt & \text{if } 0 \le \varmean \le \meanmin\\
        \frac{\boundgt}{\meanmax-\frac{1}{2}} (\varmean - \frac{1}{2}) & \text{if } \meanmin < \varmean < \meanmax\\
        \boundgt & \text{if } \meanmax\le \varmean \le 1.
    \end{cases}
\end{align*}
In words, the function $\funcloglinear$ is piecewise linear. On the interval $[0, \meanmin]$, its value equals the constant $-\boundgt$. On the interval $[\meanmin, \meanmax]$, it is a line passing through the points $(\meanmin, -\boundgt)$ and $(\meanmax, \boundgt)$. On the interval $[\meanmax, 1]$, its value equals the constant $\boundgt$. See Fig.~\ref{fig:func_log_surrogate} for a comparison of the three functions $\funclog, \funclogplus$ and $\funcloglinear$.

It can be verified that $\funclogplus(\varmean) \ge \funcloglinear(\varmean)$ for any $\varmean\in [0, 1]$. Hence, 
\begin{align}
    \Expect[\funclogplus(\mean)] \ge \Expect[\funcloglinear(\mean)].\label{eq:toy_lb_compare_two_surrogates}
\end{align}
Recall that our goal is to prove~\eqref{eq:toy_lb_surrogate_less_bias_raw}:
\begin{align*}
    \Expect[\funclog(\mean)] \le \Expect[\funclogplus(\mean)].    
\end{align*}
Given~\eqref{eq:toy_lb_compare_two_surrogates}, it suffices to prove that
\begin{align}
    \Expect[\funclog(\mean)] \le \Expect[ \funcloglinear(\mean)].\label{eq:toy_lb_compare_suffice}
\end{align}
The rest of the proof is devoted to proving~\eqref{eq:toy_lb_compare_suffice}.~~\\
    
It can be verified that $\funclog$ and $\funcloglinear$ are anti-symmetric around $\frac{1}{2}$. That is, for any $\varmean\in [0, 1]$, we have
\begin{subequations}\label{eq:eq:toy_lb_anti_symmetric}
\begin{align}
    & \funclog(\varmean) =-\funclog(1-\varmean)\\
    & \funcloglinear(\varmean) = -\funcloglinear(1-\varmean).
\end{align}
\end{subequations}
In particular, we have
\begin{align}\label{eq:toy_lb_anti_symmetric_at_half}
    \funclog\left(\frac{1}{2}\right) = \funcloglinear\left(\frac{1}{2}\right) = 0.
\end{align}
It can also be verified that 
\begin{align}\label{eq:toy_lb_compare_func_log_func_linear}
    \funclog(\varmean)\ge \funcloglinear(\varmean), \qquad \text{for all } \varmean\in \left[0, \frac{1}{2}\right].
\end{align}
    
Recall the notation of $\numwins  = \numcomparisons \meanemp$ representing the number of times that item $1$ beats item $2$ among the $\numcomparisons$ comparisons between them. We have $\numwins \sample \binomial(\numcomparisons, \meanmax)$. Therefore,
\begin{align}
    \Expect[\funclog(\mean)] - \Expect\left[ \funcloglinear(\mean)\right]  & =  \Expect_\numwins \left[\funclog\left(\frac{\numwins}{\numcomparisons}\right)\right] - \Expect_\numwins \left[ \funcloglinear\left(\frac{\numwins}{\numcomparisons}\right)\right] \nonumber\\
    & = \sum_{\numwinsobs=0}^{\numcomparisons} \left[\funclog\left(\frac{\numwinsobs}{\numcomparisons}\right) - \funcloglinear\left(\frac{\numwinsobs}{\numcomparisons}\right)\right]\cdot \Prob(\numwins = \numwinsobs) \nonumber\\
    & \stackrel{\stepone}{=} \left(\sum_{\numwinsobs=0}^{\floor*{\frac{\numcomparisons}{2}}} + \sum_{\numwinsobs=\ceil*{\frac{\numcomparisons}{2}}}^{\numcomparisons}\right) \left[(\funclog-\funcloglinear)\left(\frac{\numwinsobs}{\numcomparisons}\right)\right]\cdot \Prob(\numwins = \numwinsobs) \nonumber\\
    & \stackrel{\steptwo}{=} \sum_{\numwinsobs=0}^{\floor*{\frac{\numcomparisons}{2}}} \left[(\funclog-\funcloglinear)\left(\frac{\numwinsobs}{\numcomparisons}\right)\cdot \Prob(\numwins = \numwinsobs)  + (\funclog-\funcloglinear)\left(1 - \frac{\numwinsobs}{\numcomparisons}\right)\cdot \Prob(\numwins = \numcomparisons-\numwinsobs)\right] \nonumber\\
    & \stackrel{\stepthree}{=} \sum_{\numwinsobs=0}^{\floor*{\frac{\numcomparisons}{2}}} (\funclog-\funcloglinear)\left(\frac{\numwinsobs}{\numcomparisons}\right) \cdot [\Prob(\numwins = \numwinsobs) -\Prob(\numwins = \numcomparisons-\numwinsobs)],\label{eq:toy_lb_compare_func_log_func_linear_expectation}
\end{align}
where \stepone is true by~\eqref{eq:toy_lb_anti_symmetric_at_half}. Specifically, when $\numcomparisons$ is even, we double-count the term of $\numwinsobs=\frac{\numcomparisons}{2}$. This term equals $(\funclog-\funcloglinear)(\frac{1}{2}) = 0$, so double-counting this term does not affect the equality. Moreover, step~\steptwo is true by a change of variable $\numwinsobs\leftarrow \numcomparisons - \numwinsobs$ in the second summation, and step~\stepthree is true by the anti-symmetry~\eqref{eq:eq:toy_lb_anti_symmetric} of the functions $\funclog$ and $\funclogplus$.

Now consider the terms in the summation~\eqref{eq:toy_lb_compare_func_log_func_linear_expectation}. By~\eqref{eq:toy_lb_compare_func_log_func_linear}, we have
\begin{align}
    (\funclog-\funcloglinear)\left(\frac{\numwinsobs}{\numcomparisons}\right) \ge 0,\qquad \text{ for all  } 0\le \numwinsobs\le \floor*{\frac{\numcomparisons}{2}}.\label{eq:toy_lb_func_log_func_linear_diff}
\end{align}
Using the binomial probabilities of $\numwins\sample \binomial(\numcomparisons, \meanmax)$, we also have
\begin{align}
    \Prob(\numwins = \numwinsobs) - \Prob(\numwins = \numcomparisons -\numwinsobs) &= \binom{\numcomparisons}{\numwinsobs} [(\meanmax)^\numwinsobs (1-\meanmax)^{\numcomparisons-\numwinsobs} - (\meanmax)^{\numcomparisons-\numwinsobs} (1-\meanmax)^{\numwinsobs}]\nonumber \\
    & = \binom{\numcomparisons}{\numwinsobs} (\meanmax)^\numwinsobs(1-\meanmax)^{\numwinsobs}\cdot [(1-\meanmax)^{\numcomparisons-2\numwinsobs} - (\meanmax)^{\numcomparisons-2\numwinsobs}]\nonumber \\
    & \stackrel{\stepone}{\le} 0, \qquad \text{for all } 0\le \numwinsobs\le \floor*{\frac{\numcomparisons}{2}}, \label{eq:toy_lb_compare_binomial}
\end{align}
where \stepone is true because $\meanmax =\frac{1}{1 + e^{-2\boundgt}}> \frac{1}{2}$, combined with the fact that $\numcomparisons-2\numwinsobs \ge 0$, for all $0\le \numwinsobs\le \floor*{\frac{\numcomparisons}{2}}$. Plugging~\eqref{eq:toy_lb_func_log_func_linear_diff} and~\eqref{eq:toy_lb_compare_binomial} back into~\eqref{eq:toy_lb_compare_func_log_func_linear_expectation}, we have
\begin{align*}
\Expect[\funclog(\mean)] - \Expect[\funcloglinear(\mean)] & \ge 0,
\end{align*}
completing the proof of~\eqref{eq:toy_lb_compare_suffice}.

%%%%%%%%%%%%
\subsubsection{Proof of Lemma~\ref{lem:toy_lb_surrogate_desired_bias}}\label{sec:proof_lem_toy_lb_surrogate_desired_bias}
We have
\begin{align}
    \Expect[\funclogplus(\mean)] - \paramgt_1 & = \Expect_\numwins \left[\funclogplus\left(\frac{\numwins}{\numcomparisons}\right)\right] - \boundgt\nonumber\\
    & =\sum_{\numwinsobs=0} ^\numobservations\funclogplus\left(\frac{\numwinsobs}{\numobservations}\right)\cdot \Prob(\numwins=\numwinsobs) - \boundgt\nonumber\\
    & \stackrel{\stepone}{=} \sum_{\numwinsobs=0}^{\floor{\numobservations\meanmax}} \frac{2\boundgt}{\meanmax} \left(\frac{\numwinsobs}{\numobservations} - \meanmax\right)\cdot \Prob(\numwins=\numwinsobs)\nonumber\\
    &= \const\left(\underbrace{%
        \sum_{\numwinsobs=0}^{\floor{\numobservations\meanmax}} \frac{\numwinsobs}{\numobservations} \cdot \Prob(\numwins = \numwinsobs)
        }_{\term_1} - \meanmax\underbrace{%
            \sum_{\numwinsobs=0}^{\floor{\numobservations\meanmax}}\Prob(\numwins = \numwinsobs)
        }_{\term_2} \right),\label{eq:bias_at_one}
\end{align}
where \stepone is true by plugging in the definition of the function $\funclogplus$ from~\eqref{eq:toy_lb_def_func_log_plus}.

Now we consider the two terms $\term_1$ and $\term_2$ separately. For any integer $\numcomb \ge 1$, any integer $\selectcomb$ such that $0 \le \selectcomb \le \numcomb$, and any real number $\paramcomb\in [0, 1]$, we define $\probcomble(\numcomb,\paramcomb,\selectcomb)$ (resp. $\probcombeq(\numcomb,\paramcomb,\selectcomb)$) as the probability that the value of the random variable $\binomial(\numcomb,\paramcomb)$ is at most (resp. equal to) $\selectcomb$. That is,
\begin{align*}
    \probcomble(\numcomb,\paramcomb, \selectcomb) &= \Prob[ \binomial(\numcomb, \paramcomb) \le \selectcomb ],\\ \probcombeq(\numcomb,\paramcomb, \selectcomb) &= \Prob[ \binomial(\numcomb, \paramcomb) = \selectcomb ].
\end{align*}
Then the term $\term_2$ can be written as
\begin{align}
    \term_2 = \probcomble(\numcomparisons, \meanmax, \floor{\numcomparisons\meanmax}).\label{eq:bias_at_one_summation_part_two}
\end{align}
For the term $\term_1$, we have
\begin{align}
    \term_1 = \sum_{\numwinsobs=0}^{\floor{\numobservations\meanmax}} \frac{\numwinsobs}{\numobservations} \cdot \Prob(\numwins=\numwinsobs) & = \sum_{\numwinsobs=0}^{\floor{\numobservations\meanmax}} \frac{\numwinsobs}{\numobservations} \cdot \binom{\numobservations}{\numwinsobs}\meanmax^\numwinsobs (1-\meanmax)^{(\numobservations-\numwinsobs)}\nonumber\\
    & = \sum_{\numwinsobs=1}^{\floor{\numobservations\meanmax}} \frac{\numwinsobs}{\numobservations}\cdot  \frac{\numobservations!}{\numwinsobs!(\numobservations-\numwinsobs)!}\meanmax^\numwinsobs (1-\meanmax)^{(\numobservations-\numwinsobs)}\nonumber\\
    & = \meanmax\sum_{\numwinsobs=1}^{\floor{\numobservations\meanmax}}\frac{(\numobservations-1)!}{(\numwinsobs-1)!(\numobservations-\numwinsobs)!}\meanmax^{\numwinsobs-1} (1-\meanmax)^{(\numobservations-\numwinsobs)}\nonumber\\
    & \stackrel{\stepone}{=} \meanmax\sum_{\numwinsobs=0}^{\floor{\numobservations\meanmax} -1}\frac{(\numobservations-1)!}{(\numwinsobs)!(\numobservations-\numwinsobs - 1)!}\meanmax^{\numwinsobs} (1-\meanmax)^{(\numobservations-1-\numwinsobs)}\nonumber\\
    & = \meanmax\sum_{\numwinsobs=0}^{\floor{\numobservations\meanmax} -1}\binom{\numobservations-1}{\numwinsobs}\meanmax^{\numwinsobs} (1-\meanmax)^{(\numobservations-1-\numwinsobs)}\nonumber\\
    & = \meanmax \cdot \probcomble(\numcomparisons-1,  \meanmax,  \floor{\numcomparisons\meanmax}-1),\label{eq:bias_at_one_summation_part_one}
\end{align}
where \stepone is true by a change of variable $\numwinsobs\leftarrow \numwinsobs-1$. Plugging~\eqref{eq:bias_at_one_summation_part_two} and~\eqref{eq:bias_at_one_summation_part_one} back into~\eqref{eq:bias_at_one}, we have
\begin{align}
    \Expect[\funclogplus(\mean)] - \paramgt_1 = \const\meanmax \cdot [\probcomble(\numcomparisons-1, \meanmax, \floor{\numcomparisons\meanmax}-1) -\probcomble(\numcomparisons, \meanmax, \floor{\numcomparisons\meanmax})].\label{eq:bias_at_one_summation_prob_diff}
\end{align}

For any integer $\numcomb\ge 1$, any integer $\selectcomb$ such that $0 \le \selectcomb \le \numcomb$, and any $\paramcomb\in [0, 1]$, we claim the combinatorial equality
\begin{align}\label{eq:toy_lb_combinatorial_general}
    \probcomble(\numcomb, \paramcomb, \selectcomb) = \probcomble(\numcomb-1, \paramcomb, \selectcomb-1) + (1-\paramcomb)\cdot \probcombeq(\numcomb-1, \paramcomb, \selectcomb).
\end{align}
To prove~\eqref{eq:toy_lb_combinatorial_general}, we use a standard combinatorial argument. Consider $\numcomb$ balls, and we select each ball independently with probability $\paramcomb$. Then the LHS of~\eqref{eq:toy_lb_combinatorial_general} equals the probability that at most $\selectcomb$ balls are selected. This event can be decomposed into two cases. Either there are at most $(\selectcomb -1)$ balls selected from the first $(\numcomb-1)$ balls; or there are exactly $\selectcomb$ balls selected from the first $(\numcomb-1)$ balls, and the last ball is not selected. These two cases correspond to the two terms on the RHS of~\eqref{eq:toy_lb_combinatorial_general}.

Now setting $\numcomb = \numcomparisons, \paramcomb=\meanmax$, and $\selectcomb = \floor{\numcomparisons\meanmax}$  in~\eqref{eq:toy_lb_combinatorial_general}, we have 
\begin{align}\label{eq:toy_lb_combinatorial}
    \probcomble(\numcomparisons, \meanmax, \floor{\numcomparisons\meanmax}) =\probcomble(\numcomparisons-1, \meanmax, \floor{\numobservations\meanmax} - 1) + (1-\meanmax)\cdot \probcombeq(\numobservations-1, \meanmax, \floor{\numobservations\meanmax}]).
\end{align}
Combining~\eqref{eq:bias_at_one_summation_prob_diff} and~\eqref{eq:toy_lb_combinatorial}, we have
\begin{align}
     \Expect[\funclogplus(\mean)] - \paramgt_1 & =-\const (1-\meanmax)\cdot \probcombeq(\numcomparisons-1, \meanmax, \floor{\numcomparisons\meanmax}).\label{eq:toy_lb_minus_combinatorial}
\end{align}

It remains to bound the term $\probcombeq(\numcomparisons-1, \meanmax, \floor{\numcomparisons\meanmax})$ on the RHS of~\eqref{eq:toy_lb_minus_combinatorial}. Writing out the binomial probability, we have
\begin{align}
    \probcombeq(\numcomparisons-1, \meanmax, \floor{\numcomparisons\meanmax}) 
    & = \binom{\numcomparisons-1}{\floor{\numcomparisons\meanmax}} \meanmax^{\floor{\numobservations\meanmax}} (1-\meanmax)^{\numobservations-1-\floor{\numobservations\meanmax}}.\label{eq:lb_prob_eq}
\end{align}
By the Stirling's approximation, we have
\begin{align*}
\sqrt{2\pi} \cdot k^{k+\frac{1}{2}}e^{-k} \le k! \le e\cdot k^{k + \frac{1}{2}}e^{-k}, \qquad \text{for any integer } k \ge 0.
\end{align*}
Then for any integer $n \ge 1$, and any integer $k$ such that $0 \le k \le n$, we have
\begin{align}
\binom{n}{k} =\frac{n!}{k! (n-k)!} \ge \const \frac{n^{n+\frac{1}{2}}}{k^{k + \frac{1}{2}} (n-k)^{n - k + \frac{1}{2}}}.\label{eq:lb_stirling_combinatorial}
\end{align}
Plugging~\eqref{eq:lb_stirling_combinatorial} into~\eqref{eq:lb_prob_eq}, we have
\begin{align}
    \probcombeq(\numcomparisons-1, \meanmax, \floor{\numcomparisons\meanmax}) & \ge \const \frac{(\numobservations-1)^{\numobservations-\frac{1}{2}}}{(\numobservations-1-\floor{\numobservations\meanmax})^{\numobservations-\frac{1}{2} - \floor{\numobservations\meanmax}}\cdot (\floor{\numobservations\meanmax})^{\floor{\numobservations\meanmax}+\frac{1}{2}}}\cdot %
    \meanmax^{\floor{\numobservations\meanmax}} (1-\meanmax)^{\numobservations-1-\floor{\numobservations\meanmax}}\nonumber\\
    & \ge \const \frac{(\numobservations-1)^{\numobservations-\frac{1}{2}}}{(\numobservations-\numobservations\meanmax)^{\numobservations-\frac{1}{2} - \floor{\numobservations\meanmax}}\cdot (\numobservations\meanmax)^{\floor{\numobservations\meanmax}+\frac{1}{2}}} \cdot %
    \meanmax^{\floor{\numobservations\meanmax}} (1-\meanmax)^{\numobservations-1-\floor{\numobservations\meanmax}}\nonumber\\
    & \ge \const \frac{(\numobservations-1)^{\numobservations-\frac{1}{2}}}{\numcomparisons^\numcomparisons\cdot (1-\meanmax)^{\numobservations-\frac{1}{2} - \floor{\numobservations\meanmax}}\cdot  (\meanmax)^{\floor{\numobservations\meanmax}+\frac{1}{2}}} \cdot %
    \meanmax^{\floor{\numobservations\meanmax}} (1-\meanmax)^{\numobservations-1-\floor{\numobservations\meanmax}}\nonumber\\
    & = \const \frac{(\numobservations-1)^{\numobservations-\frac{1}{2}}}{\numcomparisons^\numcomparisons} \cdot %
    \meanmax^{-\frac{1}{2}} (1-\meanmax)^{-\frac{1}{2}}\nonumber\\
    & \stackrel{\stepone}{=} \const \frac{(\numobservations-1)^{\numobservations-\frac{1}{2}}}{\numobservations^{\numobservations}} \ge \const\frac{1}{\sqrt{\numobservations-1}} (1 - \frac{1}{\numobservations})^\numobservations  \greaterorder \frac{1}{\sqrt{\numobservations}},\label{eq:bias_at_boundary_prob_diff}
\end{align}
where \stepone is true because $ \meanmax =\frac{1}{1 + e^{-2\boundgt}}$ is bounded away from $0$ and $1$ by a constant.

Combining~\eqref{eq:toy_lb_minus_combinatorial} and~\eqref{eq:bias_at_boundary_prob_diff}, we have
\begin{align*}
   \Expect[\funclogplus(\mean)]-\paramgt_1 \le -\frac{\const}{\sqrt{\numobservations}},\qquad \text{for some constant } \const > 0.
\end{align*}

%%%%%%%%%%%%%%%%%%%%%%%%
\subsubsection{Proof of Lemma~\ref{lem:symmetric_oracle}}\label{sec:proof_lem_symmetric_oracle}

First consider the unconstrained oracle $\paramoracleunconstrain$. We prove that for any $\param\not \in \rangeboxoracle$, there exists some $\param'\in \rangeboxoracle$ such that $\negloglikelihood(\param') < \negloglikelihood(\param)$, where both $\param$ and $\param'$ satisfy the centering constraint.
    
Consider any $\param\not \in \rangeboxoracle$. By the definition of $\rangeboxoracle$, there exist some integers $\idxitem$ and $\idxitemalt$ where $2 \le \idxitem <\idxitemalt \le \numitems$, such that $\param_\idxitem \ne \param_\idxitemalt$. By the symmetry of the manipulated observations $\{\meanoracle_{\idxitem\idxitemalt}\}$ defined in~\eqref{eq:oracle_probabilities} with respect to item $2$ through item $\numitems$, we have that for any $\param\in \reals^\numitems$,
\begin{align}
    \negloglikelihood(\{\meanoracle_{\idxitem, \idxitemalt}; \param\}) = \negloglikelihood(\{\meanoracle_{\idxitem, \idxitemalt}; \param_\perm\}),\label{eq:lb_oracle_loss_symmetric}
\end{align}
where $\perm: \{2, \ldots, \numitems\}\rightarrow \{2, \ldots, \numitems\}$ is any permutation of item $2$ through item $\numitems$, and $\param_\perm = [\param_1, \param_{\perm(2)}, \ldots, \param_{\perm(\numitems)}]$. For every $\permshift\in \{0, 1, \ldots, \numitems-2\}$, define $\perm_\permshift$ as the permutation where item $2$ through item $\numitems$ are shifted $\permshift$ positions to the left in a circle. That is, for every $\idxitem\in \{2, \ldots, \numitems\}$, we have
\begin{align*}
    \perm_\permshift(\idxitem) = 2 + [(\idxitem-2+ \permshift) \mod (\numitems-1)].
\end{align*}
    
Now define $\param'= \frac{1}{\numitems-1}\sum_{\permshift=0}^{\numitems-2} \param_{\perm_\permshift}$. It can be verified that
\begin{align}
    \param' = \left[\param_1, \frac{1}{\numitems-1}\sum_{\idxitem=2}^\numitems \param_\idxitem, \ldots, \frac{1}{\numitems-1}\sum_{\idxitem=2}^\numitems \param_\idxitem\right]  \in \rangeboxoracle.\label{eq:lb_construction_oracle_contradiction}
\end{align} 
Moreover, we have
\begin{align*}
    \negloglikelihood(\param') = \negloglikelihood\left(\frac{1}{\numitems-1}\sum_{\permshift=0}^{\numitems - 2} \param_{\perm_\permshift}\right) & \stackrel{\stepone}{<} \frac{1}{\numitems-1}\sum_{\permshift=0}^{\numitems - 2} \negloglikelihood(\param_{\perm_\permshift}) \stackrel{\steptwo}{=} \negloglikelihood(\param),
\end{align*}
where \stepone is due to the strict convexity of the negative log-likelihood function $\negloglikelihood$ in Lemma~\ref{lem:log_likelihood_convex}, and \steptwo is due to~\eqref{eq:lb_oracle_loss_symmetric}.

Now we argue the equivalence of the unconstrained oracle $\paramoracleunconstrain$ defined in~\eqref{eq:defn_unconstrained_oracle} and~\eqref{eq:defn_unconstrained_oracle_equiv}. If a solution $\paramoracleunconstrain$ to~\eqref{eq:defn_unconstrained_oracle} exists, then we have $\paramoracleunconstrain\in \rangeboxoracle$ and it is trivially also the solution to~\eqref{eq:defn_unconstrained_oracle_equiv}. On the other hand, if a solution $\paramoracleunconstrain$ to~\eqref{eq:defn_unconstrained_oracle_equiv} exists, assume for contradiction that $\paramoracleunconstrain$ is not a solution to~\eqref{eq:defn_unconstrained_oracle}. Then either there exists no solution to~\eqref{eq:defn_unconstrained_oracle}, or the solution to~\eqref{eq:defn_unconstrained_oracle} is not $\paramoracleunconstrain$. In either case, there exists some $\param$ such that $\negloglikelihood(\param) < \negloglikelihood(\paramestunconstrain)$. By~\eqref{eq:lb_construction_oracle_contradiction}, we construct some $\param' \in \rangeboxoracle$ such that $\negloglikelihood(\param') < \negloglikelihood(\param) < \negloglikelihood(\paramestunconstrain)$. This contradicts the assumption that $\paramestunconstrain$ is the optimal solution to~\eqref{eq:defn_unconstrained_oracle_equiv}. Hence, Eq.~\eqref{eq:defn_unconstrained_oracle} and~\eqref{eq:defn_unconstrained_oracle_equiv} are equivalent definitions of the unconstrained oracle $\paramestunconstrain$.~~\\
    
The same argument can be extended to the constrained oracle $\paramestboundgt$, by noting that if $\param\in \rangeboxgt$, then in the construction~\eqref{eq:lb_construction_oracle_contradiction} we have $\param' \in \rangeboxgt$.

%%%%%%
\subsubsection{Proof of Lemma~\ref{lem:finite_solution_conditioning}}\label{sec:proof_lem_finite_solution_conditioning}

Note that the lemma statement is conditioned on the event $\eventmean$. That is, we observe $\mean_1 =\meancondition$ for some $\frac{1}{2} \le \meancondition \le \meanmax < 1$. In particular, we have $0 < \mean_1 < 1$. Then there exists at least one directed edge from node $1$ to nodes $\{2, \ldots, \numitems\}$, and at least one directed edge from nodes $\{2, \ldots, \numitems\}$ to node $1$. Then it suffices to prove that the subgraph consisting of nodes $\{2, \ldots, \numitems\}$ is strongly-connected \whpone.

Note that the observations $\{\meanemp_{\idxitem\idxitemalt}\}$ for any $2 \le \idxitem < \idxitemalt \le \numitems$ are all independent of $\mean_1$, and therefore independent of the event $\eventmean$. Using the arguments in Lemma~\ref{lem:finite_solution}, we have that the subgraph consisting of nodes $\{2, \ldots, \numitems\}$ is strongly-connected \whpone.

%%%%%%
\subsubsection{Proof of Lemma~\ref{lem:concentration_conditioning}} \label{sec:proof_lem_concentration_conditioning}

Note that the lemma statement is conditioned on the event $\eventmean$. That is, we observe $\mean_1 =\meancondition$ for some $\frac{1}{2} \le \meancondition \le \meanmax < 1$.

When $\idxitemthree=1$, the desired inequality~\eqref{eq:lb_mcdiardmid_desired} holds trivially, because conditioned on $\eventmean$, we have
\begin{align*}
    \sum_{\idxitem\ne 1} \meanemp_{1 \idxitem} - \sum_{\idxitem\ne 1} \meanoracleevent_{1 \idxitem} & =  (\numitems-1)\meancondition - (\numitems-1) \meancondition = 0.
\end{align*}

Now consider every $\idxitemthree\in \{2 \ldots, \numitems\}$. Consider the (unconditional) McDiarmid's inequality of~\eqref{eq:mcdiarmid} in the proof of Lemma~\ref{lem:concentration}. Replacing the summation sign $\sum_{\idxitem\ne \idxitemthree}$ on the LHS of~\eqref{eq:mcdiarmid} by the summation sign $\sum_{\substack{ \idxitem\ge 2\\ \idxitem\ne \idxitemthree}}$ (that is, we further exclude $\idxitem=1$ from the summation) yields the unconditional inequality:
\begin{align}\label{eq:mcdiarmid_conditioning_raw}
    \Prob\left[\abs*{\sum_{\substack{2 \le \idxitem\le \numitems\\ \idxitem\ne \idxitemthree}}\meanemp_{\idxitemthree\idxitem} -\sum_{\substack{2 \le \idxitem\le \numitems\\ \idxitem\ne \idxitemthree}}\meangt_{\idxitemthree\idxitem}} \le \const\sqrt{\frac{\numitems(\log\numitems + \log\numcomparisons)}{\numcomparisons}}\right] \ge 1- \frac{\const'}{\numitems^2\numcomparisons},
\end{align}
where $\const,\const' > 0$ are constants. Now we condition~\eqref{eq:mcdiarmid_conditioning_raw} on the event $\eventmean$. Note that for all $\idxitem, \idxitemthree\in \{2, \ldots, \numitems\}$ with $\idxitem \ne \idxitemthree$, the terms $\{\mean_{\idxitemthree\idxitem}\}$  are independent of $\eventmean$. Moreover, by the expression of $\meanoracleevent_{\idxitemthree\idxitem}$ in~\eqref{eq:defn_oracle_probabilities_condition}, we have $\meangt_{\idxitemthree\idxitem} = \frac{1}{2} = \meanoracleevent_{\idxitemthree\idxitem}$ conditioned on $\eventmean$. Hence, we have
\begin{align}\label{eq:mcdiarmid_conditioning}
    \Prob\left[\left.\abs*{\sum_{\substack{2 \le \idxitem\le \numitems\\ \idxitem\ne \idxitemthree}}\meanemp_{\idxitemthree\idxitem} -\sum_{\substack{2 \le \idxitem\le \numitems\\ \idxitem\ne \idxitemthree}}\meanoracleevent_{\idxitemthree\idxitem}} \le \const\sqrt{\frac{\numitems(\log\numitems + \log\numcomparisons)}{\numcomparisons}}\quad \right|\; \eventmean \,\right] \ge 1- \frac{\const'}{\numitems^2\numcomparisons}.
\end{align}
    
Now we bound the quantity $\abs*{\meanemp_{\idxitemthree 1} - \meanoracleevent_{\idxitemthree 1}}$ conditioned on $\eventmean$. By the definition of $\meanemp_1$, we have that among the $(\numitems-1)\numcomparisons$ comparisons $\{\resultcomparison_{1\idxitemalt}^{(\idxcomparison)}\}_{\idxitemalt\in\{2, \ldots, \numitems\}, \idxcomparison\in [\numcomparisons]}$ in which item $1$ is involved, there are $(\numitems-1)\numcomparisons \mean_1$ terms that have value $1$, and the rest have value $0$. Hence, each $\mean_{1\idxitemalt}$ can be thought of as the mean of $\numcomparisons$ comparisons sampled without replacement from the $(\numitems-1)\numcomparisons$ comparisons $\{\resultcomparison_{1\idxitemalt}^{(\idxcomparison)}\}_{\idxitemalt\in\{2, \ldots, \numitems\}, \idxcomparison\in [\numcomparisons]}$. By Hoeffding's inequality (sampling without replacement), we have that for every $\idxitemalt\in \{2, \ldots, \numitems\}$,
\begin{align*}
    \Prob\left[\left.\abs*{\meanemp_{ 1\idxitemalt} - \meanoracleevent_{ 1\idxitemalt}} \le \const\sqrt{\frac{\log\numitems+\log\numcomparisons}{\numcomparisons}}\; \right | \; \eventmean \right] & \ge 1 - 2 \exp\left(- \const' (\log\numitems+\log\numcomparisons)\right)\\
    & \ge 1 - \frac{\const''}{\numitems^2\numcomparisons},
\end{align*}
where $\const, \const', \const'' > 0$ are constants. Equivalently, by a change of variables, we have that for every $\idxitemalt\in \{2, \ldots, \numitems\}$,
\begin{align}\label{eq:hoeffding}
    \Prob\left[\left.\abs*{\meanemp_{\idxitemthree 1} - \meanoracleevent_{\idxitemthree 1}} \le \const\sqrt{\frac{\log\numitems+\log\numcomparisons}{\numcomparisons}}\; \right| \; \eventmean \right] \ge 1 - \frac{\const''}{\numitems^2\numcomparisons}.
\end{align}
        
Combining~\eqref{eq:mcdiarmid_conditioning} and~\eqref{eq:hoeffding} by the triangle inequality, and taking a union bound over $\idxitemthree \in \{2, \ldots, \numitems\}$ completes the proof.

%%%%%%
\section{Proof of Theorem~\ref{thm:l2_ub_lb}}\label{sec:proof_l2_ub_lb}

In this \app, we present the proof of Theorem~\ref{thm:l2_ub_lb}. Both Theorem~\ref{thm:l2_ub_lb}\ref{part:l2_lb} and Theorem~\ref{thm:l2_ub_lb}\ref{part:l2_ub} are closely related to Theorem 2 from~\cite{shah2016topology}. Under our setting, the quantity $\sigma$ defined in~\cite{shah2016topology} is a universal constant, and the quantities $\zeta$ and $\gamma$ defined in~\cite{shah2016topology} are constants that depend only on the constant $\boundgt$.

%%%%%%
\subsection{Proof of Theorem~\ref{thm:l2_ub_lb}\ref{part:l2_lb}}

Theorem~\ref{thm:l2_ub_lb}\ref{part:l2_lb} is a direct consequence of Theorem 2(a) from~\cite{shah2016topology}. We now provide some details on how to apply Theorem 2(a) from~\cite{shah2016topology}. Under our setting, each pair of items is compared $\numcomparisons$ times. Therefore, the sample size $\numsamples$ is 
\begin{align}\label{eq:l2_sample_size}
    \numsamples = \binom{\numitems}{2} \numcomparisons =\bigTheta(\numitems^2\numcomparisons).
\end{align} 
Moreover, under our setting the underlying topology is a complete graph. Let $\laplacian$ denote the scaled Laplacian as defined in Eq.~(4) from~\cite{shah2016topology}, and let $\laplacianinv$ denote the Moore-Penrose pseudoinverse of $\laplacian$. From~\cite{shah2016topology}, the spectrum of $\laplacian$ for a complete graph is $0, \frac{2}{\numitems-1}, \ldots, \frac{2}{\numitems-1}$. Therefore, we have
\begin{subequations}\label{eq:l2_laplacian_quantities}
\begin{align}
    \eigenvalue_2(\laplacian) & = \frac{2}{\numitems-1},\\
    \trace(\laplacianinv) & = (\numitems-1)\cdot \frac{\numitems-1}{2} = \frac{(\numitems-1)^2}{2}.
\end{align}
\end{subequations}

Plugging~\eqref{eq:l2_sample_size} and~\eqref{eq:l2_laplacian_quantities} into Theorem 2(a) from~\cite{shah2016topology} shows that the Theorem~\ref{thm:l2_ub_lb}\ref{part:l2_lb} holds for all $\numcomparisons\ge \constcomparisons$, where $\constcomparisons$ is a constant.

%%%%%%
\subsection{Proof of Theorem~\ref{thm:l2_ub_lb}\ref{part:l2_ub}}
The proof of Theorem~\ref{thm:l2_ub_lb}\ref{part:l2_ub} closely mimics the proof of Theorem 2(b) from~\cite{shah2016topology} (which is in turn based on Theorem 1(b) from~\cite{shah2016topology}). In what follows, we state a minor modification to be made in order to extend the proof from~\cite{shah2016topology} to Theorem~\ref{thm:l2_ub_lb}\ref{part:ub}.

In the proof from~\cite{shah2016topology}, the box constraint for the \mle $\paramestbound{\boundgt}$ is only used to obtain the following bound (see \App A.2 from~\cite{shah2016topology}):
\begin{align}
    v^T \deltwo \negloglikelihood(w) v \ge \frac{\gamma}{\numsamples\sigma^2} \norm{Xv}_2^2,\qquad \text{for all } v, w\in \rangeboxgt.\label{eq:l2_box_constraint_used}
\end{align}
Now we fix any constant $\boundused$ such that $\boundused>\boundgt$. It can be verified that~\eqref{eq:l2_box_constraint_used} still holds when replacing $\rangeboxgt$ by $\rangeboxused$, where we now allow $\gamma$ to depend on both $\boundused$ and $\boundgt$. Since $\boundused$ is assumed to be a constant, we have that $\gamma$ is still a constant. Then the rest of the arguments from~\cite{shah2016topology} carry to the proof of Theorem~\ref{thm:l2_ub_lb}\ref{part:l2_ub}.

\end{document}